\pdfoutput=1

\documentclass{article}

\usepackage[final, nonatbib]{neurips_data_2021}

\usepackage[T1]{fontenc}    
\usepackage{url}            
\usepackage{booktabs}       
\usepackage{amsfonts}       
\usepackage{nicefrac}       
\usepackage{microtype}      
\usepackage{xcolor}         

\definecolor{green}{RGB}{48,128,20}
\usepackage{amsmath}
\usepackage{titlesec}
\usepackage{textcomp,booktabs}

\usepackage{colortbl}
\definecolor{pink}{rgb}{.99,.91,.95}
\usepackage{graphics}
\usepackage{multirow}
\usepackage{pdfpages}
\usepackage{amsmath, amssymb, amsthm, xspace, color}

\usepackage{graphicx} 
\usepackage{float} 
\usepackage{bm}
\usepackage{xcolor}
\usepackage{subfig}
\usepackage{enumitem}

\usepackage{compact}

\usepackage[colorlinks,citecolor=blue]{hyperref}
\usepackage[maxbibnames=99,autolang=other]{biblatex}
\newcommand{\etc}{\emph{etc.}\xspace} 
\newcommand{\ie}{\emph{i.e.}\xspace} 
\newcommand{\eg}{\emph{e.g.\xspace}} 

\newcommand{\nop}[1]{}

\newcommand{\ra}[1]{\renewcommand{\arraystretch}{#1}}
\newcommand{\benchmark}{\textsc{Wrench}\xspace}

\newcommand{\red}[1]{{\textcolor{red}{#1}}}

\newcommand{\blue}[1]{{\textcolor{blue}{#1}}}

\title{\benchmark: A Comprehensive Benchmark \\ for Weak Supervision}

\author{%
Jieyu Zhang$^{1,2}$, Yue Yu$^3$, Yinghao Li$^3$, Yujing Wang$^1$, Yaming Yang$^1$,  Mao Yang$^1$,\\\textbf{Alexander Ratner$^2$}\\
$^1$Microsoft Research Asia\quad $^2$University of Washington\quad $^3$Georgia Institute of Technology\\
\texttt{\small \{jieyuz2, ajratner\}@cs.washington.edu} \\
\texttt{\small \{yujwang, yayaming, maoyang\}@microsoft.com} \\
\texttt{\small \{yueyu, yinghaoli\}@gatech.edu} \\
}

\begin{document}

\maketitle

\begin{abstract}

Recent \emph{Weak Supervision (WS)} approaches have had widespread success in easing the bottleneck of labeling training data for machine learning by synthesizing labels from multiple potentially noisy supervision sources. However, proper measurement and analysis of these approaches remain a challenge.
First, datasets used in existing works are often private and/or custom, limiting standardization.
Second, WS datasets with the same name and base data often vary in terms of the labels and weak supervision sources used, a significant "hidden" source of evaluation variance.
Finally, WS studies often diverge in terms of the evaluation protocol and ablations used.
To address these problems, we introduce a benchmark platform, \benchmark, for thorough and standardized evaluation of WS approaches. It consists of 22 varied real-world datasets for classification and sequence tagging;
a range of real, synthetic, and procedurally-generated weak supervision sources;
and a modular, extensible framework for WS evaluation, including implementations for popular WS methods.
We use \benchmark to conduct extensive comparisons over more than 120 method variants to demonstrate its efficacy as a benchmark platform. The code is available at \url{https://github.com/JieyuZ2/wrench}.

\end{abstract}

\section{Introduction}

One of the major bottlenecks for deploying modern machine learning models in real-world applications is the need for substantial amounts of manually-labeled training data.
Unfortunately, obtaining such manual annotations is typically time-consuming and labor-intensive, prone to human errors and biases, and difficult to keep updated in response to changing operating conditions. 
To reduce the efforts of annotation, recent weak supervision (WS) frameworks have been proposed which focus on enabling users to leverage a diversity of weaker, often programmatic supervision sources~\cite{ratner2017snorkel, Ratner16, Ratner19} to label and manage training data in an efficient way. 
Recently, WS has been widely applied to various machine learning tasks in a diversity of domains: scene graph prediction~\cite{chen2019scene}, video analysis~\cite{fu2019rekall, Varma2019multi}, image classification~\cite{das2020goggles}, image segmentation~\cite{hooper2020cut}, autonomous driving~\cite{Weng2019UtilizingWS}, relation extraction~\cite{Jia2021HeterogeneousGN,zhou2020nero,liu2017heterogeneous}, named entity recognition~\cite{safranchik2020weakly,lison2020named,li2021bertifying, lan2020connet, DBLP:conf/naacl/GoelORVR21}, text classification~\cite{ren2020denoising, yu-etal-2021-fine,shu2020learning,shu2020leveraging}, dialogue system~\cite{DBLP:conf/aaai/MallinarSUGGHLZ19}, biomedical~\cite{Kuleshov2019AMD,fries2017swellshark,Mallory2020ExtractingCR}, healthcare~\cite{Fries2021OntologydrivenWS,DBLP:journals/patterns/DunnmonRSKMSGLL20,Fries2019WeaklySC,DBLP:conf/miccai/SaabDGRSRR19,Wang2019ACT,Saab2020WeakSA}, software engineering~\cite{rao2021search}, sensors data~\cite{furst2020transport,khattar2019multi}, E-commerce~\cite{mathewdefraudnet,zhang2021queaco}, and multi-agent systems~\cite{DBLP:conf/iclr/ZhanZYSL19}.

\begin{figure*}[t]
  \centering
  \includegraphics[width=0.8\columnwidth]{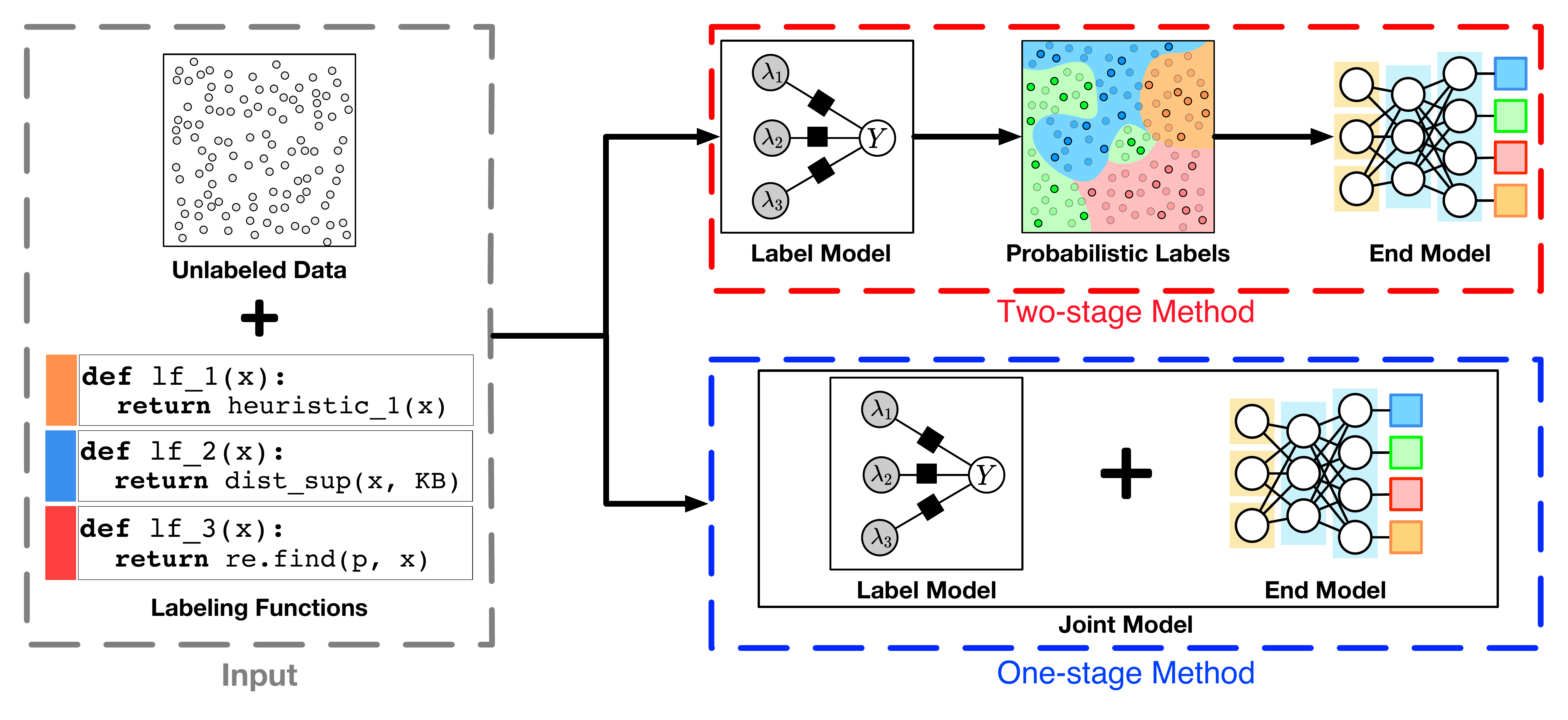} 
  \caption{An overview of WS pipeline.}
  \label{fig:overview}
\end{figure*}

In a WS approach, users leverage \emph{weak supervision sources}, \eg, heuristics, knowledge bases, and pre-trained models, instead of manually-labeled training data. In this paper, we use the \textit{data programming} formalism~\cite{Ratner16} which abstracts these weak supervision sources as \emph{labeling functions}, which are user-defined programs that each provides labels for some subset of the data, collectively generating a large but potentially overlapping set of votes on training labels.
The labeling functions may have varying error rates and may generate conflicting labels on certain data points. 
To address these issues, researchers have developed modeling techniques which aggregate the noisy votes of labeling functions to produce training labels (often referred to as a \textit{label model})~\cite{Ratner16, Ratner19, fu2020fast, Varma2019multi}, which often build on prior work in modeling noisy crowd-worker labels, e.g.~\cite{DawidSkene}.
Then, these training labels (often confidence-weighted or probabilistic) are in turn used to train an \emph{end model} which can generalize beyond the labels for downstream tasks.
These two-stage methods mainly focus on the efficiency and effectiveness of the label model, while maintaining the maximal flexibility of the end model.
Recent approaches have also focused on integrating semi- or self-supervised approaches~\cite{yu-etal-2021-fine}; we view these as modified end models in our benchmarking framework.
In addition to these two-stage methods, researchers have also explored the possibility of coupling the label model and the end model in an end-to-end manner~\cite{ren2020denoising, lan2020connet, karamanolakis2021self}. 
We refer to these one-stage methods as \emph{joint models}.
An overview of WS pipeline can be found in Fig.\ref{fig:overview}.

Despite the increasing adoption of WS approaches, a common benchmark platform is still missing, leading to an evaluation space that is currently rife with custom and/or private datasets, weak supervision sources that are highly varied and in often hidden and uncontrolled ways, and basic evaluation protocols that are highly variable.
Several thematic issues are widespread in the space:
\begin{itemize}[leftmargin=0.5cm]
    \item \textbf{Private and/or custom datasets:} Due to the lack of standardized benchmark datasets, researchers often construct their own datasets for comparison. In particular, WS approaches are often practically motivated by real-world use cases where labeled data is difficult to generate, resulting in datasets are often based on real production need and therefore are not publicly avaiable.
    \item \textbf{Hidden weak supervision source variance:} Unlike traditional supervised learning problems, WS datasets vary not just in the unlabeled data $X$, but also crucially in the labels $Y$ and weak supervision sources they derive from (see Fig.~\ref{fig:dataset_stats_box}).
    This latter degree of variance has a major effect on the performance of WS approaches; however it is often poorly documented and controlled for.
    For example, it is not uncommon to have two datasets with \textit{completely different weak supervision sources} bear the exact same name (usually deriving from the source of the unlabeled data $X$) in experimental results, despite being entirely different datasets from a WS perspective.
    \item \textbf{End-to-end evaluation protocol:} WS approaches involve more complex (e.g. two-stage) pipelines, requiring greater (yet often absent) care to normalize and control evaluations. For example, it is not uncommon to see significant variance in which stage of a two-stage pipeline performance numbers are reported for, what type of training labels are produced, etc~\cite{Ratner16, yu-etal-2021-fine}.
\end{itemize}

To address these issues and contribute a resource to the growing WS community, we developed \textbf{W}eak Supe\textbf{r}vision B\textbf{ench}mark (\benchmark), a benchmark platform for WS with $22$ diverse datasets from the literature, a range of standardized real, synthetic, and procedurally generated weak supervision sources, and a modular, extendable framework for execution and evaluation of WS approaches, along with initial implementations of recent popular WS methods.
\benchmark includes:
\begin{itemize}[leftmargin=0.5cm]
    \item A diverse (and easily extensible) set of 22 real-world datasets for two canonical, annotation-intensive machine learning problems, classification and sequence tagging, including datasets used in existing WS studies and new ones we contribute.
    \item A range of real (user-generated) weak supervision sources, and new synthetic and procedural weak supervision source generators, enabling systematic study on the effect of different supervision source types on the performances of WS methods, e.g. with respect to accuracy, variance, sparsity, conflict and overlap, correlation, and more. 
    \item A modular, extensible Python codebase for standardization of implementation, evaluation, and ablation of WS methods, including standardized evaluation scripts for prescribed metrics, unified interfaces for publicly available methods, and re-implementations of some other popular ones.
\end{itemize}

To demonstrate the utility of \benchmark, we analyze the effect of a range of weak supervision attributes using \benchmark's procedural weak supervision generation suite, illustrating the effect of various salient factors on WS method efficacy (Sec.~\ref{sec:generator}).
We also conduct extensive experiments to render a comprehensive evaluation of popular WS methods (Sec.~\ref{sec:exp}), exploring more than 120 compared methods and their variants (83 for classification and 46 for sequence tagging).
We plan to \emph{continuously update} \benchmark with more datasets and methods as the field advances.
We welcome contributions and expect the scope and breadth of \benchmark to increase over time.

\section{Related Work}

\paragraph{Weak Supervision.} Weak supervision builds on many previous approaches in machine learning, such as distant supervision \cite{mintz2009distant,Hoffmann2011KnowledgeBasedWS,Takamatsu2012ReducingWL}, crowdsourcing \cite{Gao2011HarnessingTC,Krishna2016VisualGC}, co-training methods \cite{Blum1998CombiningLA}, pattern-based supervision \cite{Gupta2014ImprovedPL}, and feature annotation \cite{Mann2010GeneralizedEC,Zaidan2008ModelingAA}.  
Specifically, weak supervision methods take multiple noisy supervision sources and an unlabeled dataset as input, aiming to generate training labels to train an end model (two-stage method) or directly produce the end model for the downstream task (one-stage method) without any manual annotation. 
Weak supervision has been widely applied on both classification~\cite{Ratner16, Ratner19, fu2020fast, yu-etal-2021-fine, ren2020denoising} and sequence tagging~\cite{lison2020named,nguyen2017aggregating, safranchik2020weakly, li2021bertifying, lan2020connet} to help reduce human annotation efforts.

\paragraph{Weak Supervision Sources Generation.} 
To further reduce the efforts of designing supervision sources, many works propose to generate supervision sources automatically.  
Snuba~\cite{varma2018snuba} generates heuristics based on a small set of labeled datasets. 
IWS \cite{boecking2021interactive} and Darwin \cite{darwin} interactively generate labeling functions based on user feedback.
TALLOR~\cite{TALLOR} and GLaRA~\cite{glara} automatically augment an initial set of labeling functions with new ones.
Different from existing works that optimize the task performance, the procedural labeling function generators in \benchmark facilitate the study of the impact of different weak supervision sources.
Therefore, we assume access to a fully-labeled dataset and generate diverse types of weak supervision sources.

\paragraph{The Scope of this Benchmark.}
We are aware that there are numerous works on learning with \emph{noisy} or \emph{distantly labeled} data for various tasks, including relation extraction~\cite{luo2017learning,mintz2009distant,Takamatsu2012ReducingWL}, sequence tagging
~\cite{liang2020bond,liu2021noisy,peng2019distantly,autoner}, image classification~\cite{co_teaching,li2021mopro,mirzasoleiman2020coresets} and visual relation detection~\cite{yao2021visual,Zhang_2017_ICCV}. There are also several benchmarks targeting on this topic~\cite{hedderich2021analysing,pmlrjiang20c,riedel2010modeling,Xiao_2015_CVPR,chu2021natcat} with different noise levels and patterns. However, these studies mainly concentrate on learning with \emph{single-source} noisy labels and cannot leverage complementary information from multiple annotation sources in weak supervision. 
Separately, there are several works~\cite{Awasthi2020Learning,karamanolakis2021self,maheshwari2021semi, mazzetto:aistats21, mazzetto:icml21} leveraging additional clean, labeled data for denoising multiple weak supervision sources, while our focus is on benchmarking weak supervision methods that do not require any labeled data.
So we currently do not include these methods in \benchmark, that being said, we plan to gradually incorporate them in the future. 



\section{Background: Weak Supervision}
\label{sec:background}
We first give some background on weak supervision (WS) at a high level.
In the WS paradigm, multiple weak supervision sources are provided which assign labels to data, which may be inaccurate, correlated, or otherwise noisy.
The goal of a WS approach is the same as in supervised learning: to train an \textit{end} model based on the data and weak supervision source labels.
This can be broken up into a \textit{two-stage} approach–separating the integration and modeling of WS from the training of the end model–or tackled jointly as a \textit{one-stage} approach.

\subsection{Problem Setup}
\label{sec:overview_problem}
We more formally define the setting of WS here.
We are given a dataset containing $n$ data points $\bm{X}=[X_1, X_2, \ldots, X_n]$ with $i$-th data point denoted by $X_i \in \mathcal{X}$.
Let $m$ be the number of WS sources $\{S_j\}_{j=1}^m$, each assigning a label $\lambda_j \in \mathcal{Y}$ to $X_i$ to vote on its respective $Y_i$ or abstaining ($\lambda_j=-1$).
We define the \emph{propensity} of one source $S_j$ as $p(\lambda_j \neq -1)$.
For concreteness, we follow the general convention of WS~\cite{Ratner16} and refer to these sources as \emph{labeling functions} (LFs) throughout the paper.
In \benchmark, we focus on two major machine learning tasks:

\paragraph{Classification:} for each $X_i$, there is an unobserved true label denoted by $Y_i \in \mathcal{Y}$.
A label matrix $L\in\mathbb{R}^{n \times m}$ is obtained via applying $m$ LFs to the  dataset $\bm{X}=[X_1, X_2, \ldots, X_n]$. We seek to build an end model $f_w: \mathcal{X} \rightarrow \mathcal{Y}$ to infer the labels $\hat{Y}$ for each $X \in \bm{X}$.

\paragraph{Sequence tagging:} each $X_i\in \bm{X}$ is a sequence of tokens $[x_{i,1}, x_{i,2}, \ldots, x_{i,t}]$, where $t$ is the length of $X_i$, with an unobserved true label list denoted by $Y_i = [y_{i,1}, y_{i,2}, \ldots, y_{i,t}]$ where $y_{i,j} \in \mathcal{Y}$.
For each sequence $X_i$ with its associated label matrix $L_i\in\mathbb{R}^{n \times t}$, we aim to produce an sequence tagger model $f_w: \mathcal{X} \rightarrow \mathcal{Y}$ which infers labels $\hat{Y} = [\hat{y}_1, \hat{y}_1, \ldots, \hat{y}_t]$ for each sequence. 

It is worth noting that, different from the \textit{semi-supervised} setting, and some recent WS work, where some ground-truth labeled data is available~\cite{Awasthi2020Learning,maheshwari2021semi,karamanolakis2021self,mazzetto:icml21, mazzetto:aistats21}, we consider the setting where we train the end model \emph{without observing any ground truth training labels}.
However, we note that \benchmark can be extended in future work to accommodate these settings as well.

\subsection{Two-stage Method}

Two-stage methods usually decouple the process of training label models and end models.
In the first stage, a \textit{label model} is used to combine the label matrix $L$ with either probabilistic \textit{soft labels} or one-hot \textit{hard labels}, which are in turn used to train the desired \emph{end model} in the second stage.
Most studies focus on developing label models while leaving the end model flexible to the downstream tasks.
Existing label models include Majority Voting (MV), Probabilistic Graphical Models (PGM) \cite{DawidSkene, Ratner16,Ratner19, fu2020fast,lison2020named,safranchik2020weakly,li2021bertifying}, \etc.
Note that prior crowd-worker modeling work can be included and subsumed by this set of approaches, e.g.~\cite{DawidSkene}.

\subsection{One-stage Method}
One-stage methods attempt to effectively train a label model and end model simultaneously~\cite{ren2020denoising,lan2020connet}.
Specifically, they usually design a neural network for aggregating the prediction of labeling functions while utilizing another neural network for final prediction.
We refer to the model designed for one-stage methods as a \emph{joint model}.

\section{Wrench Benchmark Platform}

\begin{table}[t]
    \begin{minipage}[b]{0.56\linewidth}
    \centering
    \caption{Statistics of all the tasks, domains and datasets included in \benchmark.}
    \scalebox{0.5}{
    \begin{tabular}{ l l l c c c c c }
        \toprule
         \multicolumn{5}{c}{} &
         \multicolumn{1}{c}{\textbf{Train}}    &
         \multicolumn{1}{c}{\textbf{Dev}}    &
         \multicolumn{1}{c}{\textbf{Test}}   \\
         \cmidrule(lr){6-6}
         \cmidrule(lr){7-7}
         \cmidrule(lr){8-8}
           \textbf{Task ($\downarrow$)} &\textbf{Domain ($\downarrow$)} & \textbf{Dataset ($\downarrow$)} & \textbf{\#Label} & \textbf{\#LF} &\textbf{\#Data} & \textbf{\#Data} & \textbf{\#Data}   \\
         \midrule
     \multirow{1}{*}{Income Class.}    & Tabular Data & Census~\cite{kohavi1996scaling, Awasthi2020Learning}  & 2 & 83  & 10,083  & 5,561 &  16,281 \\\midrule[0.05pt] \midrule[0.05pt]
     
    \multirow{2}{*}{Sentiment Class.}   

     & Movie & IMDb~\cite{IMDB,ren2020denoising}  & 2 & 5 & 20,000 & 2,500 & 2,500 \\      
     
     & Review & Yelp~\cite{AGNews,ren2020denoising}  & 2 & 8  & 30,400  & 3,800 & 3,800      \\\midrule[0.05pt] \midrule[0.05pt]
     
    \multirow{2}{*}{Spam Class. }    & Review & Youtube~\cite{youtube}  & 2 & 10  & 1,586 & 120 & 250 \\      
    
        & Text Message & SMS~\cite{sms, Awasthi2020Learning}  & 2 & 73 & 4,571 & 500 & 500  \\\midrule[0.05pt] \midrule[0.05pt]
        
    \multirow{1}{*}{Topic Class.}    & News & AGNews~\cite{AGNews,ren2020denoising}  & 4 & 9  & 96,000 & 12,000 & 12,000  \\\midrule[0.05pt] \midrule[0.05pt]
    
    \multirow{1}{*}{Question Class.}    & Web Query & TREC~\cite{trec, Awasthi2020Learning}  & 6 & 68 & 4,965 & 500 & 500  \\\midrule[0.05pt] \midrule[0.05pt]
    
     \multirow{4}{*}{Relation Class.}    & News & Spouse~\cite{spouse, ratner2017snorkel}  & 2 & 9 & 22,254 & 2,811 & 2,701       \\
     
     & Biomedical & CDR~\cite{davis2017comparative, ratner2017snorkel}  & 2 & 33  & 8,430 & 920 & 4,673  \\     
     & Web Text & SemEval~\cite{hendrickx2010semeval, zhou2020nero}  & 9 & 164  & 1,749 & 200 & 692   \\
     
     & Chemical & ChemProt~\cite{chemprot,yu-etal-2021-fine}  & 10 & 26  & 12,861 & 1,607 & 1,607 \\\midrule[0.05pt] \midrule[0.05pt]
     
     \multirow{3}{*}{Image Class.}  & \multirow{3}{*}{Video} & Commercial~\cite{fu2020fast}  &  2 & 4 & 64,130  & 9,479 & 7,496  \\     
     
     &  & Tennis Rally~\cite{fu2020fast}  &  2 & 6 & 6,959    &   746 & 1,098     \\
     
     &  & Basketball~\cite{fu2020fast}  &  2 & 4  & 17,970    & 1,064 & 1,222    \\\midrule[0.05pt] \midrule[0.05pt]
     
     \multirow{8}{*}{Sequence Tagging} 
    &News & CoNLL-03~\cite{conll03,lison2020named}             & 4 & 16  & 14,041 & 3250 & 3453    \\\cmidrule(lr){2-8} 
    & \multirow{2}{*}{Web Text}  & WikiGold~\cite{wikigold,lison2020named}       & 4 & 16  &  1,355 & 169 & 170     \\
    &  & OntoNotes 5.0~\cite{weischedel2011ontonotes}  & 18 & 17 & 115,812 & 5,000 & 22,897\\   \cmidrule(lr){2-8}   
    & \multirow{2}{*}{Biomedical}  & BC5CDR~\cite{cdr,li2021bertifying}            & 2 & 9   & 500 & 500 & 500    \\
    &  & NCBI-Disease~\cite{dougan2014ncbi,li2021bertifying} & 1 & 5 &  592 & 99 & 99   \\\cmidrule(lr){2-8} 
    & \multirow{2}{*}{Review} & Laptop-Review~\cite{laptop,li2021bertifying}      & 1 & 3  & 2,436 & 609 & 800     \\
    &  & MIT-Restaurant~\cite{mitr}       & 8 & 16 & 7,159 &  500   &   1,521     \\\cmidrule(lr){2-8} 
    & Movie & MIT-Movies~\cite{mitmovie}        & 12 & 7 & 9,241 &  500   &      2,441  \\
    \bottomrule
    \end{tabular}
    }
    \label{tab:dataset_stats}
\end{minipage}\hfill
\begin{minipage}[b]{0.4\linewidth}
\centering
  \includegraphics[width=1.0\textwidth]{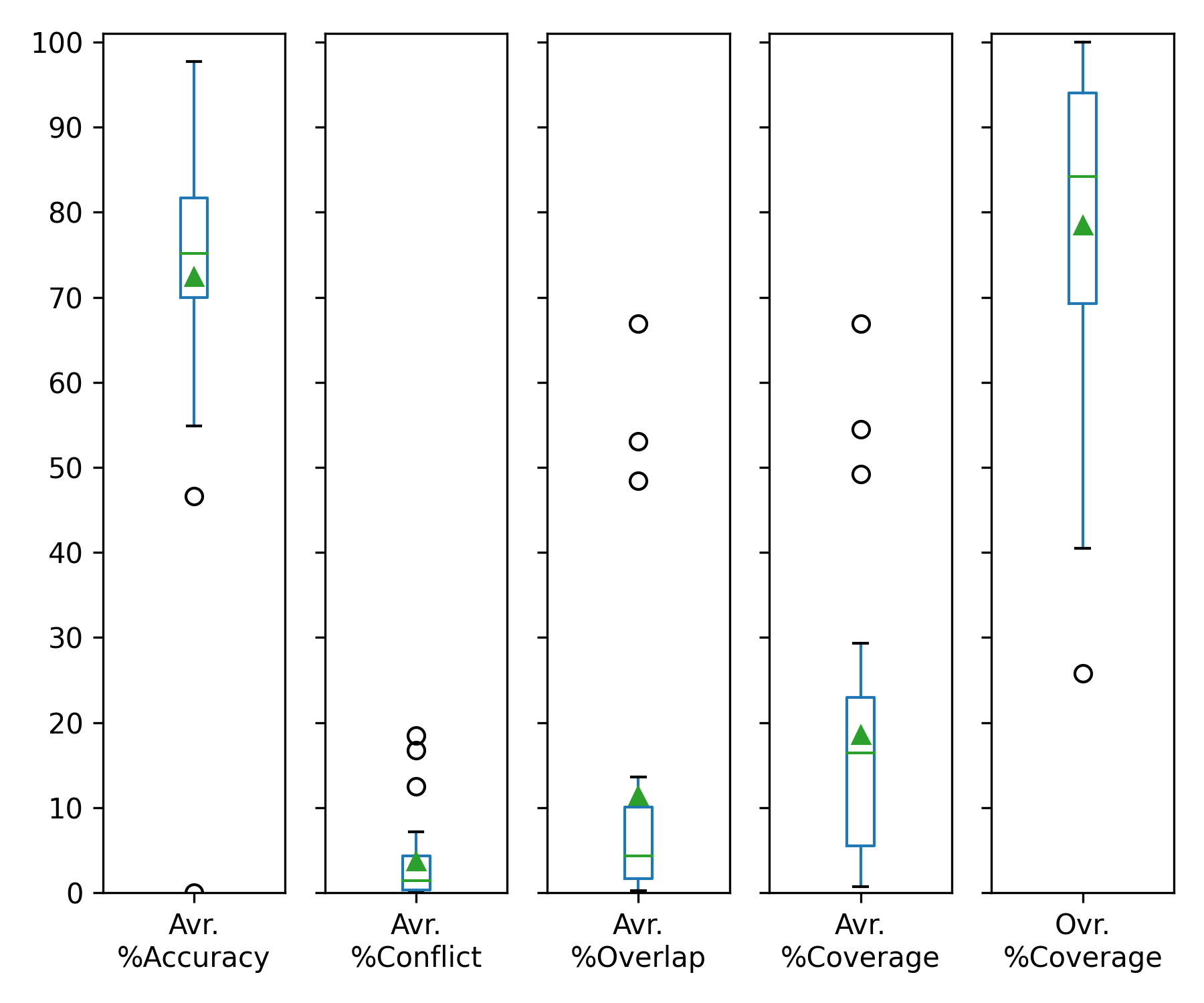}
  \captionof{figure}{\textbf{Box plots}: The coverage, overlap, conflict and accuracy of LFs in collected datasets. We can see the LFs have diverse properties across datasets.}
  \label{fig:dataset_stats_box}
\end{minipage}

\end{table}

We propose the first benchmark platform, \benchmark, for weak supervision (WS). 
Specifically, {\benchmark} includes the following components:

\paragraph{A collection of 22 real-world datasets.}
We collect 22 publicly available real-world datasets and the corresponding user-provided LFs from the literature.
The statistics of the datasets is in Table~\ref{tab:dataset_stats}.
The datasets cover a wide range of topics, including both generic domains such as web text, news, videos and specialized ones including biomedical and chemical publications.
The corresponding LFs have various forms, such as key words~\cite{autoner}, regular expressions~\cite{Awasthi2020Learning}, knowledge bases~\cite{liang2020bond} and human-provided rules~\cite{fu2020fast}.
Some relevant statistics of the LFs is in Fig.~\ref{fig:dataset_stats_box};
the box plots demonstrate that the LFs have diverse properties across datasets, enabling more thorough comparisons among WS approaches.
The description of each dataset and detailed statistics are in App.~\ref{sec:dataset}. 
\vspace{-2mm}
\paragraph{A range of procedural labeling function generators.}
In addition to the manually-created LFs coupled with each dataset, \benchmark provides a range of procedural labeling function generators for the first time, giving users fine-grain control over the space of weak supervision sources.
It facilitates researchers to evaluate and diagnose WS methods on (1) synthetic datasets or (2) real-world datasets with procedurally generated LFs.
Based on the generators, users could study the relationship between different weak supervision sources and WS method performances.
The details of the generators and the provided studies are in Sec.~\ref{sec:generator}.
Notably, the unified interface of \benchmark allows users to add more generators covering new types of LFs easily.
\vspace{-2mm}

\paragraph{Abundant baseline methods and extensive comparisons.}
\benchmark provides unified interfaces for a range of publicly available and popular methods.
A summary of models currently included in \benchmark is in Table~\ref{tab:methods}.
With careful modularization, users could pick \emph{any} label model and end model to form a two-stage WS method, while also choosing to use soft or hard labels for training the end model, leading to more than 100 method variants.
We conduct extensive experiments to offer a systematic comparison over all the models and possible variants on the collected 22 datasets (Sec. ~\ref{sec:exp}).
Another benefit of this modularity is that other approaches can be easily contributed, and we plan to add more models in the future.

\begin{table*}[t]
    \centering
    \caption{The initial set of methods included in \benchmark. A brief introduction of each method can be found in App.~\ref{sec:methods}. We plan to add more methods in near future.}
    \scalebox{0.6}{
    \setlength{\tabcolsep}{2em}
    \begin{tabular}{ l l l l }
        \toprule
           \textbf{Task} & \textbf{Module} & \textbf{Method} &\textbf{Abbr.}  \\
         \midrule
         
    \multirow{14}{*}{Classification} &  \multirow{6}{*}{Label Model}
    & Majority Voting &  {MV}  \\
    & & Weighted Majority Voting & {WMV} \\
    & & Dawid-Skene~\cite{DawidSkene} & {DS} \\
    & & Data Progamming~\cite{Ratner16} & {DP} \\
    & & MeTaL~\cite{Ratner19} & {MeTaL} \\
    & & FlyingSquid~\cite{fu2020fast} & {FS} \\
    \cmidrule(lr){2-4} 
     &  \multirow{5}{*}{End Model} & Logistic Regression & {LR} \\
     & & Multi-Layer Perceptron Neural Network & {MLP} \\
     & & {BERT~\cite{devlin2019bert}} & B  \\
    & & {RoBERTa~\cite{liu2019roberta}} & R  \\
    & & {COSINE-BERT~\cite{yu-etal-2021-fine}} & BC  \\
    & & {COSINE-RoBERTa~\cite{yu-etal-2021-fine}} & RC  \\
     \cmidrule(lr){2-4} 
     &  \multirow{1}{*}{Joint Model} & {Denoise~\cite{ren2020denoising}} & {Denoise}  \\\midrule[0.05pt] \midrule[0.05pt]
     
     \multirow{6}{*}{Sequence Tagging}    &  \multirow{2}{*}{Label Model} & Hidden Markov Model~\cite{lison2020named} & {HMM} \\
     & & Conditional Hidden Markov Model~\cite{li2021bertifying} & {CHMM} \\     \cmidrule(lr){2-4} 
     &  \multirow{2}{*}{End Model} & LSTM-CNNs-CRF~\cite{ma2016end} & {LSTM-CNNs-CRF} \\
     & & BERT~\cite{devlin2019bert} & {BERT} \\     \cmidrule(lr){2-4} 
    &  \multirow{1}{*}{Joint Model} & Consensus Network~\cite{lan2020connet} & {ConNet}  \\
    \bottomrule
    \end{tabular}
    }
    \label{tab:methods}
\end{table*}

\section{Labeling Function Generators}
\label{sec:generator}

In addition to user-generated labeling functions collected as part of the 22 benchmark datasets in \benchmark, we provide two types of weak supervision source generators in \benchmark in order to enable fine-grain exploration of WS method efficacy across different types of weak supervision:
(1) \textit{synthetic} labeling function generators, which directly generate labels from simple generative label models;
(2) \textit{procedural} labeling function generators, which automatically generate different varieties of real labeling functions given an input labeled dataset.
In this section, we introduce the generators in detail and provide some sample studies to demonstrate the efficacy of these generators in enabling finer-grain exploration of the relationship between weak supervision source qualities and WS method performances. 
For simplicity, in this section we constrain our study to binary classification tasks and the details on implementation and parameter can be found in App.~\ref{sec:para_study}.

\subsection{Synthetic Labeling Function Generator}

The synthetic labeling function generators are independent of the input data $X$; instead, they directly generate the labeling function output labels.
We provide one initial synthetic generator to start in \benchmark's first release, which generates labels according to a classic model where the LF outputs are conditionally independent given the unseen true label $Y$~\cite{DawidSkene, Ratner16}.
For this model, \benchmark provides users with control over two important dimensions: accuracy and propensity.
In addition, users can control the \emph{variance} of the LF accuracies and propensities via the respective \emph{radius} of accuracy and propensity parameters.
For example, the accuracy of each LF could be chosen to be uniformly sampled from $[a-b, a+b]$, where $a$ is the mean accuracy and $b$ is the radius of accuracy, resulting in a variance of $\frac{b^2}{12}$.
We construct the synthetic label generators to be extensible, for example, to include more controllable parameters and more complex models.

\begin{figure}[t]
\begin{minipage}{.45\linewidth}
\centering
\subfloat[]{\label{main:a}\includegraphics[scale=.2]{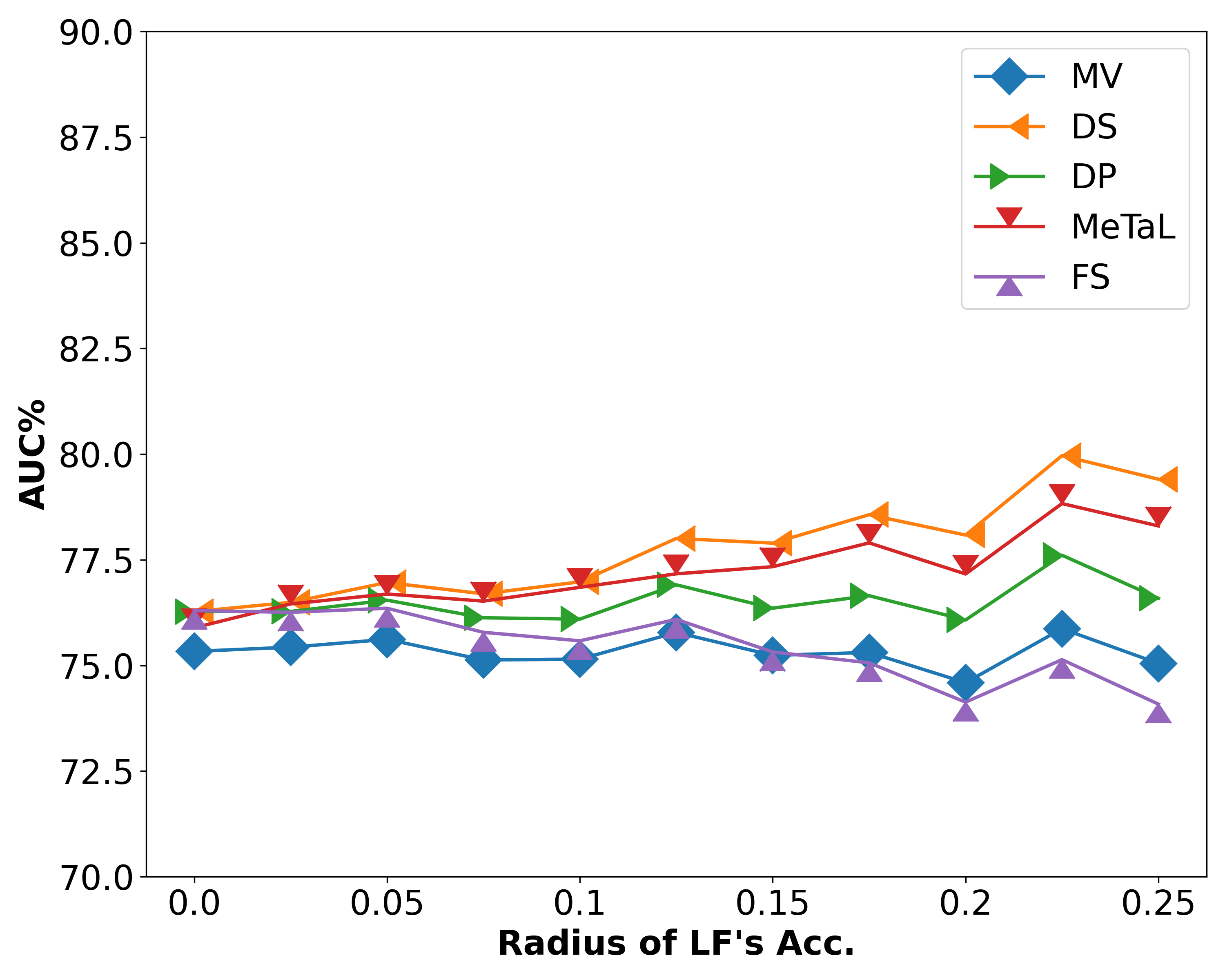}}
\end{minipage}%
\begin{minipage}{.45\linewidth}
\centering
\subfloat[]{\label{main:b}\includegraphics[scale=.2]{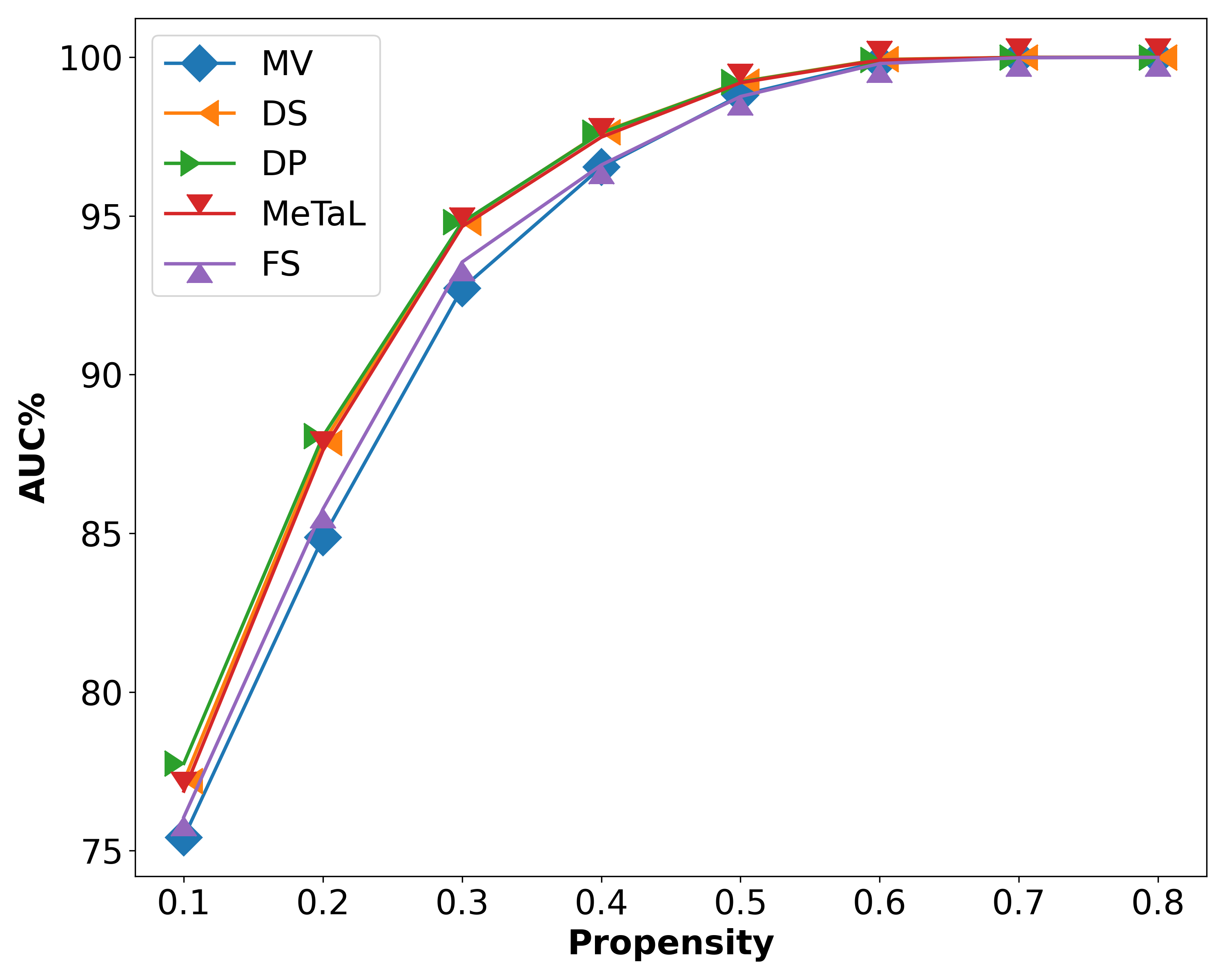}}
\end{minipage}
\caption{Label models performance (AUC) on synthetic LFs with varying (a) radius of LF's accuracy and (b) propensity. We can see that when the radius of LF's accuracy is large or the propensity of LFs is small, the label model performance are more divergent.}
\label{fig:syn}
\end{figure}

Based on this generator, we study different dimensions of LFs and found that the comparative performance of label models are largely dependent on the variance of accuracy and propensity of LFs.
First, we fix other dimensions and vary the radius of LF's accuracy, and generate $\bm{Y}$ and LFs for binary classification.
As shown in Fig.~\ref{fig:syn}(a), we can see that the performance of label models diverge when we increase the variance of LFs' accuracy by increasing the radius of accuracy.
Secondly, we vary the propensity of LFs. 
From the curves in Fig.~\ref{fig:syn}(b), we can see that if we increase the propensity of LFs, the label models' performance keep increasing and converge eventually, while when the propensity is lower, the label models perform differently.
These observations indicate the importance of the dimensions of LFs,  which could lead to the distinct comparative performance of label models.

\subsection{Procedural Labeling Function Generator}

The procedural labeling function generator class in \benchmark requires the input of a labeled dataset $(X, Y)$, i.e. with data features and ground truth training labels.
The procedural generators create a pool of \emph{candidate LFs} based on a given \textit{feature lexicon}.
Each candidate LF $S$ consists of a single or composite feature from the provided lexicon and a label.
The final set of generated LFs is those candidate LFs whose individual parameters (e.g. accuracies and propensities) and group parameters (e.g. correlation and data-dependency) meet user-provided thresholds.
These procedurally-generated LFs mimic the form of user-provided ones, but enable fine-grain control over the LF attributes.

\begin{figure*}[t]
\begin{minipage}{.33\linewidth}
\centering
\subfloat[]{\label{main:a}\includegraphics[scale=.18]{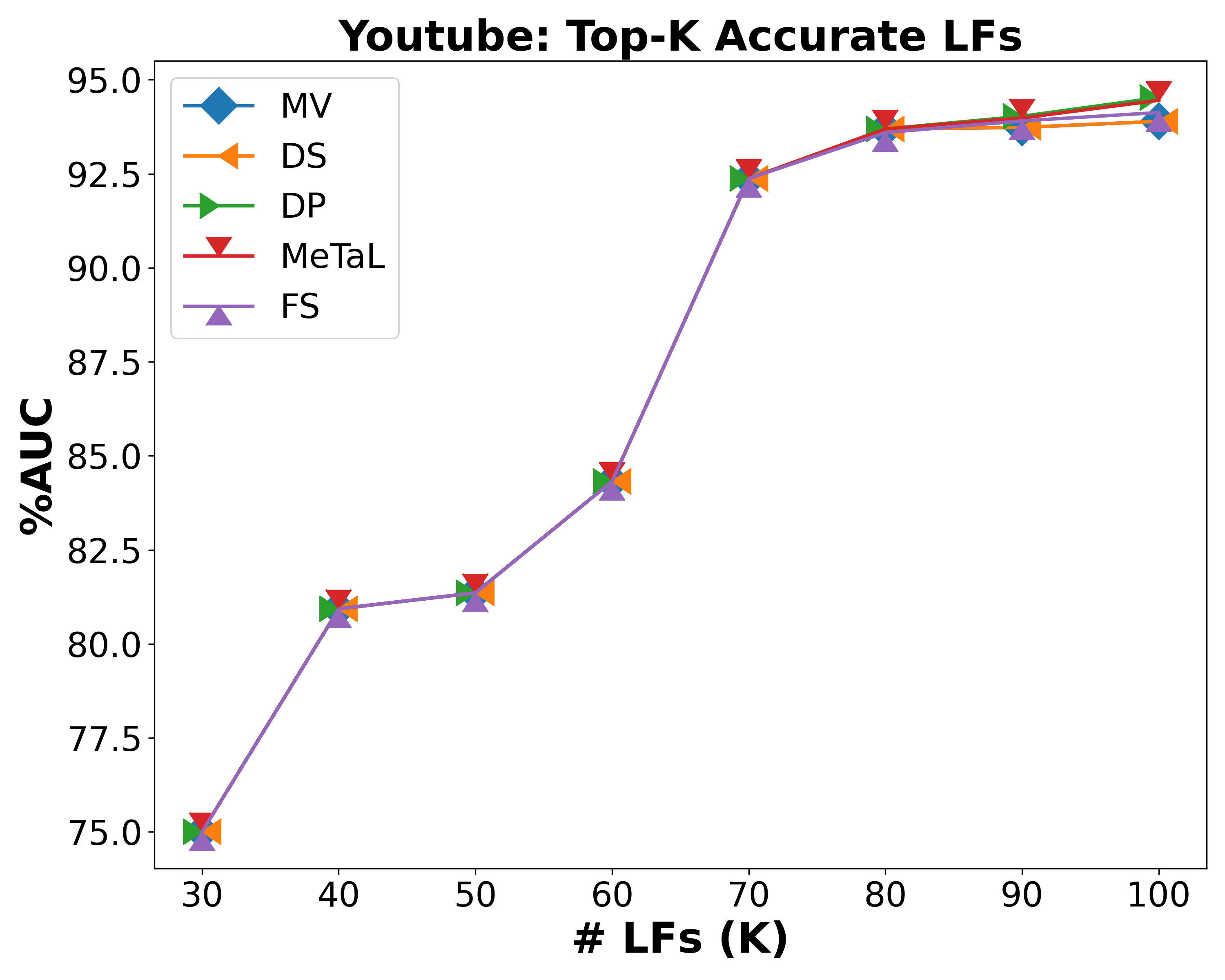}}
\end{minipage}%
\begin{minipage}{.33\linewidth}
\centering
\subfloat[]{\label{main:b}\includegraphics[scale=.18]{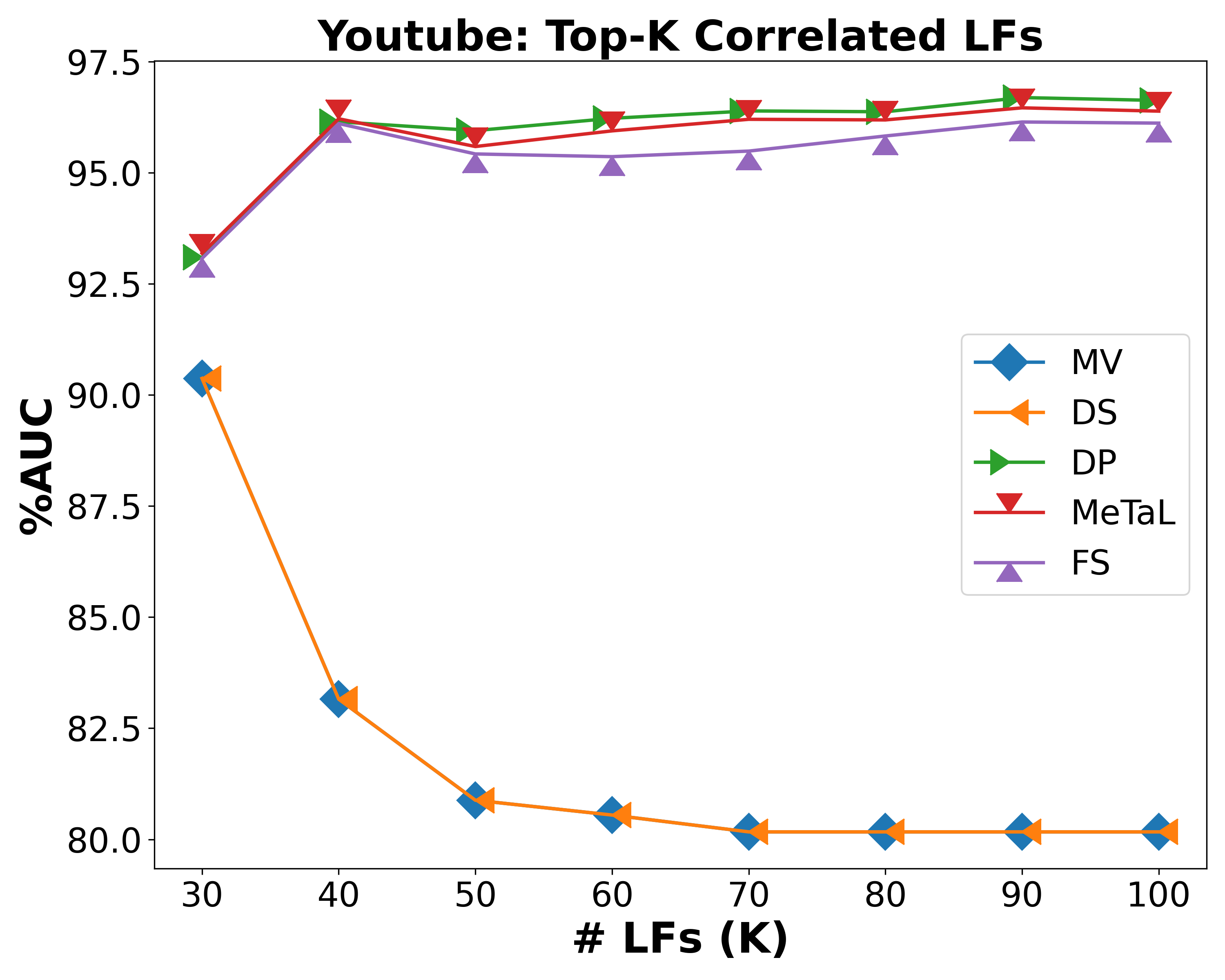}}
\end{minipage}%
\begin{minipage}{.33\linewidth}
\centering
\subfloat[]{\label{main:c}\includegraphics[scale=.18]{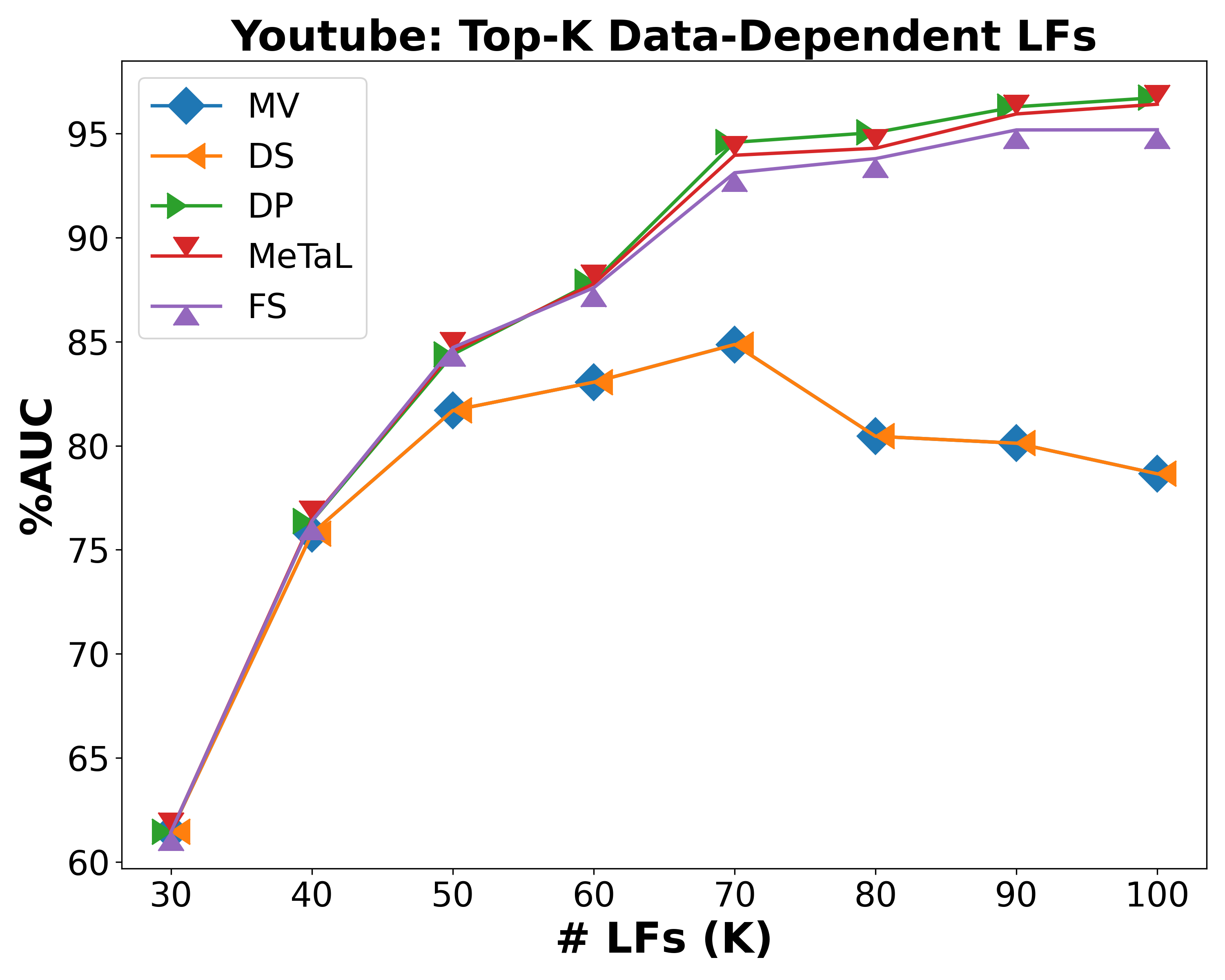}}
\end{minipage}
\begin{minipage}{.33\linewidth}
\centering
\subfloat[]{\label{main:d}\includegraphics[scale=.18]{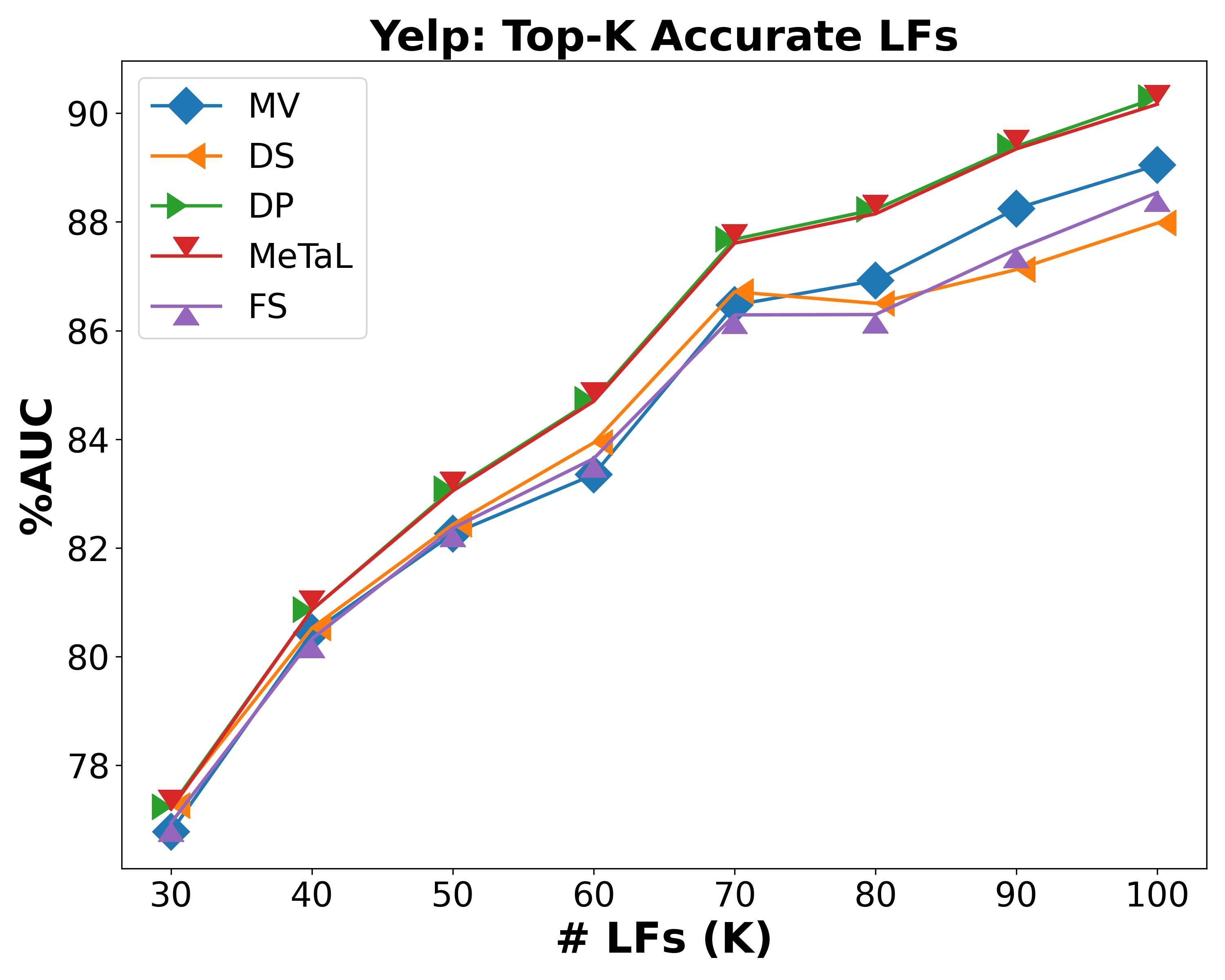}}
\end{minipage}%
\begin{minipage}{.33\linewidth}
\centering
\subfloat[]{\label{main:e}\includegraphics[scale=.18]{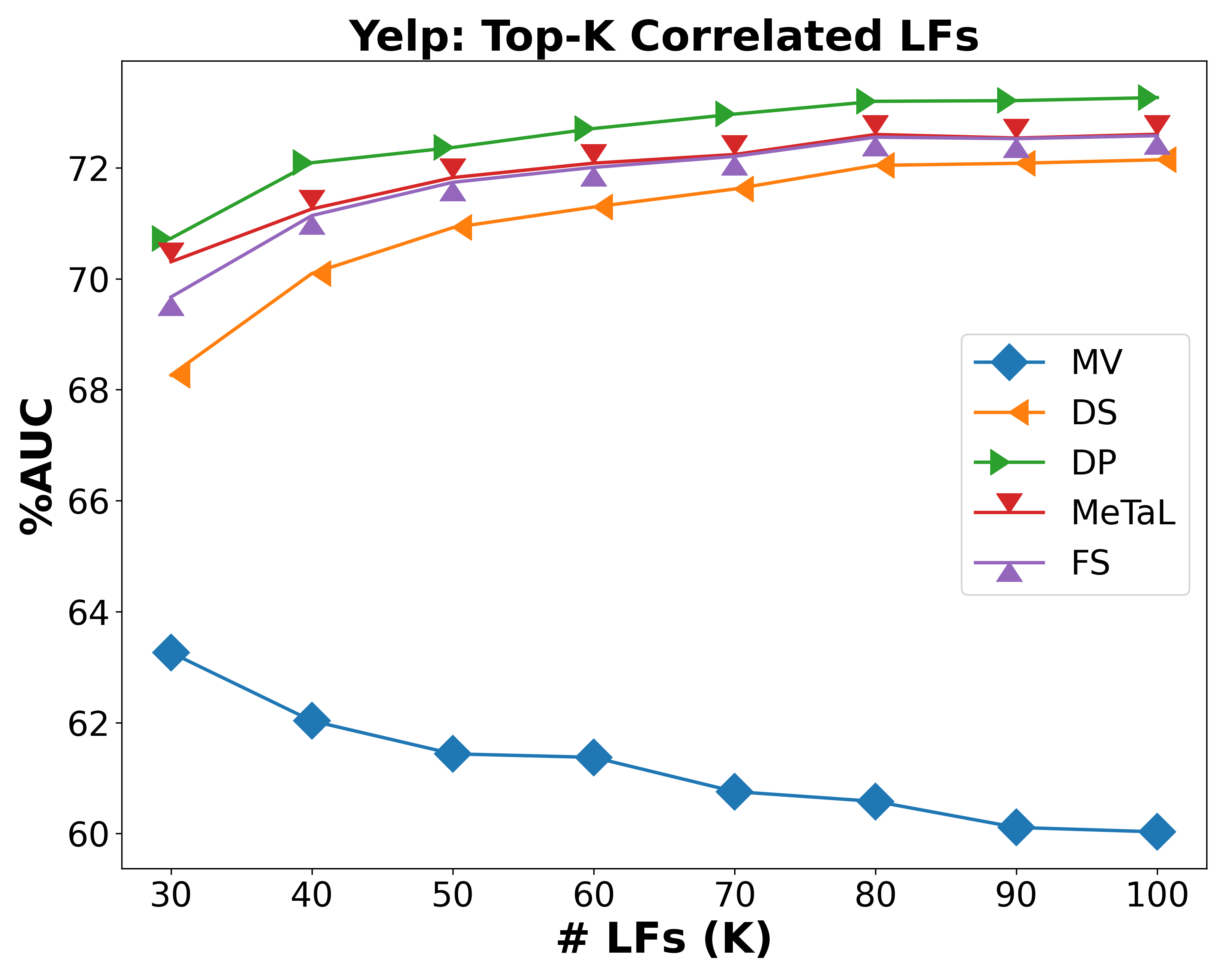}}
\end{minipage}%
\begin{minipage}{.33\linewidth}
\centering
\subfloat[]{\label{main:f}\includegraphics[scale=.18]{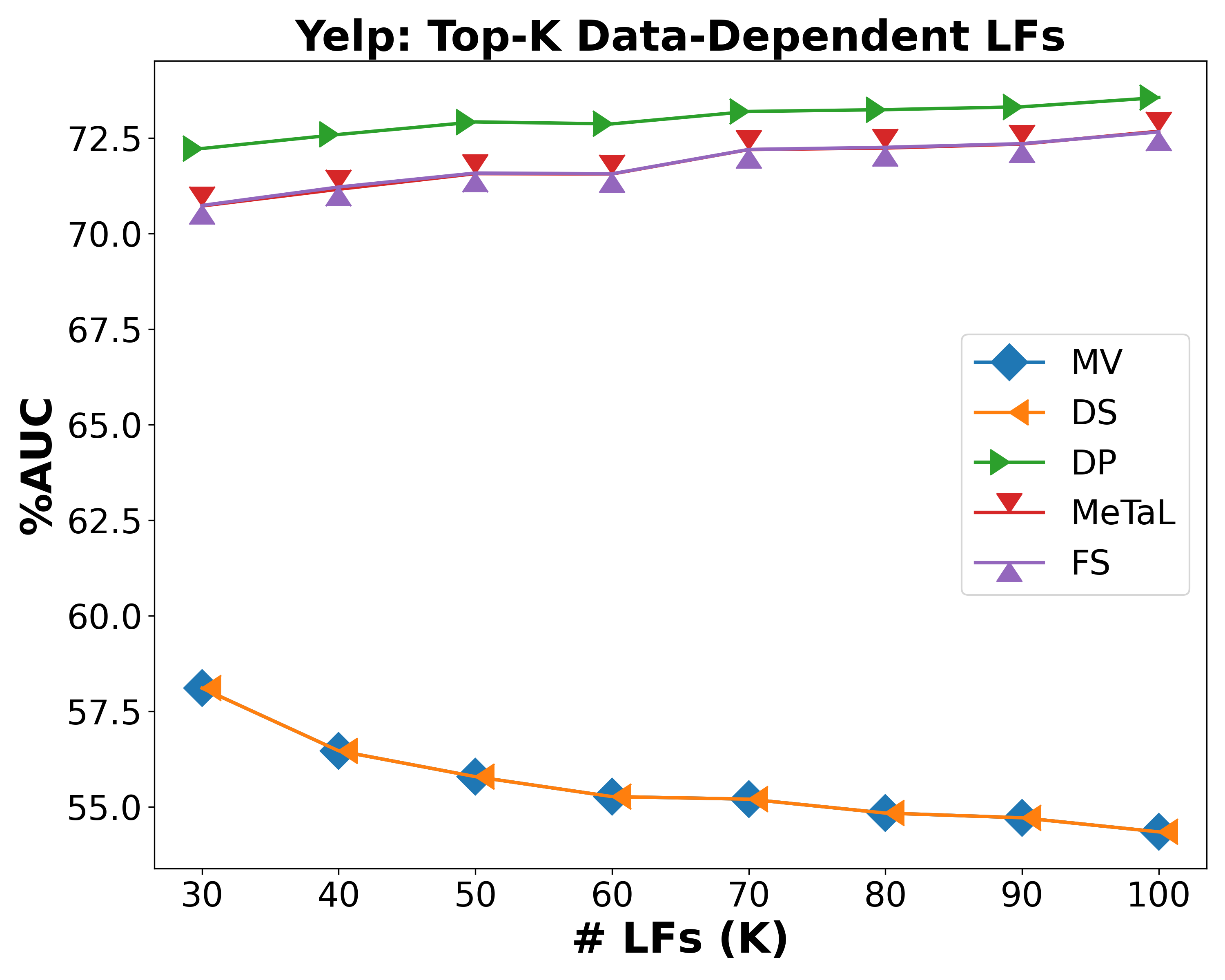}}
\end{minipage}
\caption{Label models performance (AUC) on Youtube and Yelp with varying types of procedural LFs, namely, top-k accurate, correlated, or data-dependent LFs. We can see that the dependency properties of LF (correlated and data-dependent) have a major effect on the comparative performance of label models.}
\label{fig:semi}
\end{figure*}

In this section, we provide an example study of how label models perform with different types of LFs on two real-world text classification datasets, Yelp and Youtube, to demonstrate the utility of these procedural generators.
For simplicity, we adopt ($n$, $m$)-gram features, where $n$ and $m$ are the minimum and maximum length of gram respectively and are input by users.
Specifically, a candidate LF $S$ consists of one label value $y$ and an ($n$, $m$)-gram feature $f$; for each data point, if the feature $f$ exists, then $S$ assigns label $y$, otherwise returns abstain ($\lambda=-1$).
We generate and examine three sets of LFs, namely, the LFs with highest (1) accuracies, (2) pairwise correlations and (3) data-dependencies (Fig.~\ref{fig:semi}).
For (2), the correlations of LFs are measured by the \emph{conditional mutual information} of a pair of candidate LFs given the ground truth labels $\bm{Y}$. 
We are interested in (2) because existing works often assume the LFs are independent conditional on $\bm{Y}$~\cite{Ratner16, Bach2017LearningTS}, however, users can hardly generate perfectly conditionally independent LFs; therefore, it is of great importance to study how label models perform when LFs are not conditionally independent.
The reason for (3) is that previous studies typically assume the LFs are \emph{uniformly} accurate across the dataset~\cite{Ratner16, Bach2017LearningTS, Ratner19, fu2020fast}, however, in practice, this is another often violated assumption–e.g. specific LFs are often more accurate on some subset of data than the other. Thus, we measure the data-dependency of LFs by the variance of accuracies of LF over clusters of data and pick LFs with the highest data dependency.

The results are in Fig.~\ref{fig:semi}.
First, in the case of top-k accurate LFs (Fig.~\ref{fig:semi}(a)\&(d)), the label models perform similarly, however, for the other two types of LFs, there are large gaps between label model performance and the superiority of recently-proposed methods, \ie, DP, MeTaL, FS, can be clearly seen.
Secondly, even within the same type of LFs, one label model can result in varying performance on different datasets; for example, when correlated LFs are generated (Fig.~\ref{fig:semi}(b)\&(e)), the DS model performs much better on Yelp than Youtube compared to the MV model.
These observations further confirm that the LFs have a major effect on the efficacy of different WS approaches, and it is critical to provide a benchmark suite for WS with varied datasets and varying types of LFs.

\section{Benchmark Experiments}
\label{sec:exp}

To demonstrate the utility of \benchmark in providing fair and rigorous comparisons among WS methods, we conduct extensive experiments on the collected real-world datasets with real labeling functions.
Here, we consider all the possible ways to compose a two-stage method using the initial models that we implement in \benchmark (Table~\ref{tab:methods}), ablating over the choice of soft and hard labels, as well as considering the one-stage methods listed.

\subsection{Classification}

\subsubsection{Evaluation Protocol}
We evaluate the performance of (1) the label model directly applied on test data; (2) the end model trained with labels provided by label model for two-stage methods; (3) the end model trained within a joint model for one-stage methods; and (4) the "gold" method, namely training an end model with ground truth labels, with different end models.
We include all the possible two-stage methods as well as the variants using soft or hard labels in our comparison, leading to 83 methods in total.

For each dataset, we adopt the evaluation metric used in previous work.
For LR and MLP applied on textual datasets, we use a pre-trained BERT model to extract the features.
Note that for the Spouse dataset, we do not have the ground truth training labels, so we do not include gold methods for it.
In addition, due to privacy issues, for the video frame classification datasets (\ie, Commerical, Tennis Rally and Basketball), we only have access to the features extracted by pre-trained image classifier instead of raw images, thus, we choose LR and MLP as end models.

\subsubsection{Evaluation Results}

Due to the space limit, we defer the complete results as well as the standard deviations to the App.\ref{sec:results}, while only presenting the top 3 best WS methods and the gold method with the best end model for each dataset in Table~\ref{tab:classification_main}.
From the table, we could observe a diversity of the best WS methods on different datasets.
In other words, there is no such method that could consistently outperform others.
This observation demonstrates that it remains challenging to design a generic method that works for diverse tasks.
For textual datasets, it is safe to conclude that fine-tuning a large pretrain language model is the best option of the end model, and COSINE could successfully improve the performance of fine-tuned language models.
Moreover, fine-tuning a pre-trained language model is, not surprisingly, much better than directly applying label model on test data in most cases, because it is well-known that large pre-trained language models like BERT can easily adapt to new tasks with good generalization performance.

\begin{table*}[t]
    \centering
    \caption{\textbf{Classification.} The performance of the best gold method and top 3 best weak supervision methods for each dataset. EM and LM stand for the end model and label model respectively. \underline{Underline} indicates using the soft label for training end model. Datasets with * are non-textual data on which BERT/RoBERTa are not applicable. Each metric value is averaged over 5 runs. The detailed results and average performance can be found in App.~\ref{sec:results}.}
    \scalebox{0.75}{
    \begin{tabular}{ l l c c c c c c c c c c c }
        \toprule 
         \multicolumn{1}{c}{} & \multicolumn{1}{c}{} & \multicolumn{2}{c}{\textbf{Best Gold}} &
         \multicolumn{3}{c}{\textbf{Top 1}} & \multicolumn{3}{c}{\textbf{Top 2}} & \multicolumn{3}{c}{\textbf{Top 3}}  \\

         \cmidrule(lr){3-4} \cmidrule(lr){5-7} \cmidrule(lr){8-10} \cmidrule(lr){11-13}
  \textbf{Dataset} &  \textbf{Metric} &\textbf{EM}  & \textbf{Value} &\textbf{EM} & \textbf{LM}  & \textbf{Value} & \textbf{EM}  & \textbf{LM} &\textbf{Value} & \textbf{EM}  & \textbf{LM} & \textbf{Value} \\
         \midrule

    IMDb   &Acc.     & R & 93.25          &RC& \underline{MeTaL}&88.86 &RC&\underline{FS}&88.48 &RC&\underline{MV}&88.48   \\\midrule
    
    Yelp&Acc. & R & 97.13 &RC&FS&95.45&RC&\underline{FS}&95.33&RC&\underline{DS}&95.01 \\\midrule
    
    Youtube&Acc. & B & 97.52   &BC&MV&98.00 &RC&MV&97.60 &RC&\underline{MV}&97.60  \\\midrule

    SMS&F1  & B & 96.96 &RC&WMV&98.02 &RC&MeTaL&97.71 &RC&\underline{WMV}&97.27  \\\midrule
    
    AGNews&Acc. & R & 91.39 &RC&DS&88.20   &RC&MV&88.15 &RC&\underline{WMV}&88.11  \\\midrule 
    TREC &Acc.  & R & 96.68 &RC&DP&82.36&RC&\underline{MeTaL}&79.84&BC&DP&78.72   \\\midrule
    
    Spouse &F1 & -- & -- & BC &FS&56.52 &--&MeTaL&46.62 &RC&\underline{MV}&46.28   \\\midrule 
    
    CDR &  F1 & R & 65.86 &--&MeTaL&69.61 &--&DP&63.51 &RC&DP&61.40   \\\midrule 
    
    SemEval & Acc. & B & 95.43 &BC&\underline{DP}&88.77  &BC&MV&86.80 &RC&\underline{DP}&86.73   \\\midrule
    
    ChemProt & Acc. & B & 89.76 &BC&\underline{DP}&61.56 &RC&MV&59.43 &RC&\underline{MV}&59.32  \\\midrule 
    
    Commerical* &F1 & MLP & 91.69 &\multicolumn{2}{c}{Denoise}&91.34  &LR&MV&90.62  &MLP&\underline{MV}&90.55  \\\midrule
    
    Tennis Rally* &F1 & LR & 82.73 &MLP&\underline{FS}&83.77 &MLP&\underline{MeTaL}&83.70 &LR&\underline{FS} &83.68  \\\midrule
    
    Basketball* &F1 & MLP & 64.97 &MLP&\underline{FS}&43.18 &MLP&\underline{WMV}&40.73 &MLP&\underline{DP}&40.70  \\\midrule
    
     Census*&F1 & MLP & 67.13   &LR&\underline{MeTaL}&58.16 &MLP&\underline{MeTaL}&57.84&MLP&MeTaL&57.66 \\
    \bottomrule
    \end{tabular}
    }

    \label{tab:classification_main}
\end{table*}

\subsection{Sequence Tagging}

\subsubsection{Evaluation Protocol}
Same as the evaluation scheme on classification tasks, we evaluate the performance of (1) the label models; (2) the end models trained by predictions from the label models; (3) the joint models; and (4) the end models trained by gold labels on the training set.  
Note that following previous works~\cite{lan2020connet,safranchik2020weakly,autoner}, we adopt \emph{hard} labels in order to fit end models which contain CRF layers. 
To adapt label models designed for classification tasks to sequence tagging tasks, we split each sequence by tokens and reformulate it as a token-level classification task. We discuss the detailed procedure on adapting label model for sequence tagging tasks in App.~\ref{sec:adapt}.
However, these models neglect the internal dependency between labels within the sequence. 
In contrast, HMM and CHMM take the whole sequence as input and predict the label for tokens in the whole sequence. 
For the end model with LSTM/BERT, we run experiments with two settings: (1) stacking a CRF layer on the top of the model, (2) using a classification head for token-level classification; and the best performance is reported. 

Following the standard protocols, we use \emph{entity-level} F1-score as the metric~\cite{lison2020named,ma2016end} and  use \textbf{BIO} schema~\cite{tjong2002introduction,li2021bertifying,lison2020named}, which labels the beginning token of an entity as \texttt{B-X} and the other tokens inside that entity as \texttt{I-X}, while non-entity tokens are marked as \texttt{O}. For methods that predict token-level labels (\eg MV), we transform token-level predictions to entity-level predictions when calculating the F1 score. 
Since BERT tokenizer may separate a word into multiple subwords, for each word, we use the result of its first token as its prediction.

\subsubsection{Evaluation Results}
Table~\ref{tab:seq_main2} demonstrates the main result of different methods on sequence tagging tasks. 
For label models, we conclude that considering dependency relationships among token-level labels during learning generally leads to better performance, as HMM-based models achieve best performance on 7 of 8 datasets. 
One exception is the MIT-Restaurants dataset, where weak labels have very small coverage. In this case, the simple majority voting-based methods achieve superior performance compared with other complex probabilistic models.
For end models, surprisingly, directly training a neural model with weak labels \emph{does not guarantee} the performance gain, especially for LSTM-based model which is trained from scratch. Such a phenomenon arrives when the quality of LFs is poor (\eg~MIT-Restaurants, LaptopReview). 
Under this circumstance, the weak labels generated through LFs are often noisy and incomplete~\cite{liang2020bond,liu2021noisy}, and the end model can easily overfit to them. As a result, there is still a significant performance gap between the results trained by gold labels and weak labels, which motivates the future research on designing methods robust against the induced noise.

\definecolor{blizzardblue}{rgb}{0.67, 0.9, 0.93}
\definecolor{columbiablue}{rgb}{0.61, 0.87, 1.0}
\definecolor{LightCyan}{rgb}{0.88,1,1}

\begin{table*}[t]
    \centering
    \caption{\textbf{Sequence Tagging.} Comparisons among different methods. The number stands for the F1 score. Each metric value is averaged over 5 runs. {\textcolor{red}{red}} and {\color{blue}{blue}} indicate the best and second best result for each end model respectively, and \colorbox{lightgray!60}{gray} is the best weak supervision method. The first 8 rows with end model as -- indicates directly apply label models on test data. The detailed results are in App.~\ref{sec:seq_app}.}
    \scalebox{0.54}{
    \begin{tabular}{ l c c c c c c c c c >{\columncolor{LightCyan}}c}
        \toprule
           \textbf{End Model ($\downarrow$)} &\textbf{Label Model ($\downarrow$)} & \textbf{CoNLL-03} & \textbf{WikiGold} & \textbf{BC5CDR} & \textbf{NCBI-Disease}& \textbf{Laptop-Review} & \textbf{MIT-Restaurant} & \textbf{MIT-Movies} & \textbf{Ontonotes 5.0} & \textbf{\underline{Average}}   \\
         \midrule
    
  \multirow{8}{*}{--}    
     & {MV}     & 60.36 & 52.24 & 83.49 & \blue{78.44} & \blue{73.27} & \colorbox{lightgray!60}{\red{48.71}} & 59.68& 58.85 & 64.38     \\
     & {WMV}    & 60.26 & 52.87 & 83.49 & \blue{78.44} & \blue{73.27} & \blue{48.19} & 60.37 & 57.58     &   64.31   \\ 
     & {DS}     & 46.76 & 42.17 & 83.49 & \blue{78.44} & \blue{73.27} & 46.81 & 54.06 & 37.70  &   57.84  \\
     & {DP}      & 62.43 & 54.81 & \blue{83.50} & \blue{78.44} & \blue{73.27} & 47.92 & 59.92 & \blue{61.85}     &  \blue{65.27}    \\
     & {MeTaL}   & 60.32 & 52.09 & 83.50 & \blue{78.44} & 64.36 & 47.66 & 56.60 & 58.27  &62.66       \\
     &{FS}     & \blue{62.49} & \blue{58.29} & 56.71 & 40.67 & 28.74 & 13.86 & 43.04 & 5.31  &  38.64     \\
     & {HMM}   & 62.18 & 56.36 & 71.57 & 66.80 & \colorbox{lightgray!60}{\red{73.63}} & 42.65 & \blue{60.56} & 55.67& 61.88 \\
     & {CHMM}    & \red{63.22} & \red{58.89} & \colorbox{lightgray!60}{\red{83.66}} &  \colorbox{lightgray!60}{\red{78.74}} & 73.26 & 47.34 & \red{61.38} & \red{64.06} & \red{66.32} \\
     \midrule[0.05pt] \midrule[0.05pt]
         
    \multirow{10}{*}{LSTM-CNN} & {Gold} & 87.46& 80.45 & 78.59 & 79.39 & 71.25 & 79.18 & 87.07 & 79.52 & 79.83\\
     \cmidrule(lr){2-2} 
      & {MV}     & 66.33 & 58.27 & \blue{74.75} & 72.44 & 63.52 & \blue{41.70} & \blue{62.41} & 61.92 & 62.47 \\
     & {WMV}    & 64.60 & 55.39 & 74.31 & 72.21 & 63.02 & 41.27 & 61.79 & 59.22 & 61.37  \\
     & {DS}     & 50.60 & 40.61 & \red{75.37} & \red{72.86} & \red{63.96} & 41.21 & 55.99 & 44.92 & 55.58 \\
     & {DP}     & \red{67.15} & 57.89  & \blue{74.79} & \blue{72.50} & 62.59 & \blue{41.62} & 62.29 & \red{63.82}  & \blue{62.83} \\
     & {MeTaL}  & 65.05 & 56.31 & 74.66 & 72.42 & \blue{63.87} & 41.48 & 62.10 & 60.43 & 61.85  \\
     & {FS}     & 66.49 & \blue{60.49} & 54.49 & 44.90 & 28.35 & 13.09 & 45.77 & 43.25 & 44.51 \\
     & {HMM}    & 66.18 &  \red{62.51} & 64.07 & 59.12 & 62.57 & 37.90 & 61.94 & 59.43 & 59.17 \\
     & {CHMM}   & 66.67 & 61.34 & 74.54 & 72.15 & 62.28 & 41.59 & \red{62.97} & \blue{63.71} & \red{62.97}\\
      \midrule[0.05pt] \midrule[0.05pt]
    \multicolumn{2}{c}{LSTM-ConNet} & 66.02 & 58.04 & 72.04 & 63.04 & 50.36 & 39.26 & 60.46 & 60.58 & 58.73
     \\\midrule[0.05pt] \midrule[0.05pt]
     
     \multirow{10}{*}{BERT} & {Gold} & 89.41 & 87.21 & 82.49 & 84.05 & 81.22 & 78.85 & 87.56 & 84.11 & 84.36 \\
     \cmidrule(lr){2-2} 
     & {MV}     & 67.08 & 63.17 & \blue{77.93} & 77.93 &71.12 & \red{42.95} & 63.71 & 63.97 & 65.62 \\
     & {WMV}    & 65.96& 61.28 & 77.76 & 78.53 & \blue{71.60} & {42.62} & 63.44 & 61.63 & 63.92  \\
     & {DS}     & 54.04 & 49.09 & 77.57 & \blue{78.69} & 71.41 & 42.26 & 58.89 & 48.55 & 58.87 \\
     & {DP}     & 67.66 & 62.91 & 77.67 & 78.18 & 71.46 & 42.27 & {63.92} & \blue{65.16} & \blue{65.97}  \\
     & {MeTaL}  & 66.34 & 61.74 & 77.80 & \red{79.02} & \red{71.80} & 42.26 & \blue{64.19} & 63.08  & 65.61\\
     & {FS}     & 67.54 & \colorbox{lightgray!60}{\red{66.58}} & 62.89 & 46.50 & 38.57 & 13.80 & 49.79 & 49.63  & 49.11 \\
     & {HMM}    & \colorbox{lightgray!60}{\red{68.48}} & 64.25 & 68.70 & 65.52 & 71.51 & 39.51 & 63.38 & 61.29 & 62.59 \\
     & {CHMM}   & \blue{68.30} & \blue{65.16} & \red{77.98} & 78.20 & 71.17 & \blue{42.79} & \colorbox{lightgray!60}{\red{64.58}} & \colorbox{lightgray!60}{\red{66.03}} & \colorbox{lightgray!60}{\red{66.50}} \\
      \midrule[0.05pt] \midrule[0.05pt]
     \multicolumn{2}{c}{BERT-ConNet} & 67.83 & 64.18 & 72.87 & 71.40 & 67.32 & 42.37 & 64.12 & 60.36 & 63.81 \\
    \bottomrule
        \end{tabular}
    }
    \label{tab:seq_main2}
\end{table*}

\section{Discussion and Recommendation}
\begin{itemize}[leftmargin=0.5cm]
    
    
    \item \textbf{Correctly categorization of method and comparing it to right baselines are critical.}
    As stated in Sec.~\ref{sec:background}, weak supervision methods could be categorized into label model, end model and joint model.
    However, we observed that in previous work, researchers, more or less, did not clearly categorize their method and compare it to inappropriate baselines.
    For example, COSINE is an end model but in the original paper, the authors coupled COSINE with MV (a label model) and compared it with another label model, MeTaL\footnote{The Snorkel baseline in their paper.}, without coupling MeTaL with an end model.
    This comparison is hardly fair and effective.
    
    \item \textbf{Strong weakly supervised models rely on high-quality supervision sources.}
    From the result shown in table~\ref{tab:classification_main} and \ref{tab:seq_main2}, we observe that both label model and end model perform well only when the quality of the overall labeling functions is reasonably good. For those datasets which have very noisy labeling functions (\eg Basketball) or very limited coverage (\eg MIT-Restaurants), there is still a large gap between the performance of fully-supervised model and weakly-supervised model. Such phenomenon illustrates that it is still necessary to check the quality of initial labeling functions before applying weak supervision models for new tasks. Otherwise, directly adopting these models may not lead to satisfactory results, and may even hurt the performance. 
    
    \item \textbf{When the end models become deeper, using soft label may be a good idea.} Based on the average performance of models across tasks, we observe that using soft labels to train the end model is better than hard labels in most cases, especially when the end model become deeper (from logistic regression to pretrained language model).
    We think this is relevant to the idea of "label smoothing"~\cite{NEURIPS2019_f1748d6b}, which prevents the deep models from overfitting to (noisy) training data.
    
    \item \textbf{Uncovered data should be used when training end models.} A common practice of weak supervision is to train an end model using only \emph{covered} data\footnote{\url{https://www.snorkel.org/use-cases/01-spam-tutorial}}, \ie, the subset of data which receive at least one weak signal.
    However, the superiority of COSINE suggests that those uncovered data should also be used in training an end model; this inspires future direction of exploring new end model training strategy combined with semi-supervised learning techniques.
    
    \item \textbf{For sequence tagging tasks, selecting appropriate tagging scheme is important.} As studied in App.~\ref{sec:schema}, choosing different tagging schema can cause up to 10\% performance in terms of F1 score. 
    This is mainly because when adopting more complex tagging schema (\eg, \texttt{BIO}), the label model could predict \emph{incorrect} label sequences, which may hurt final performance especially for the case where the number of entity types is small. 
    Under this circumstance, it is recommended to use \texttt{IO} schema during model training. 
    For other datasets including more types of entities, there is no clear winners for different schemes.
    
    \item \textbf{For classification tasks, MeTaL and MV are the most worth-a-try label models and for end model, deeper is better.}
    According to the model performance averaged over datasets, we find MeTaL and MV are the best label models when using different end models or directly applying label models on test set.
    For the choices of end model, not surprisingly, deeper model is better.
    
    \item \textbf{For sequence tagging tasks, CHMM gains an advantage over other baselines in terms of label model.} 
    CHMM generally outperforms other label models and achieves highest average score. 
    We remark that CHMM is the only label model that combines the outputs of labeling function with data feature (\ie BERT embeddings). 
    The superiority of CHMM indicates that developing \emph{data-dependent label model} will be a promising direction for the future research. 
    For the end model, pre-trained language models are more suitable end models, as it can capture general semantics and syntactic information~\cite{rogers2020primer} which will benefit the downstream tasks.

\end{itemize}

\section{Conclusion and Future Work}

We introduce \benchmark, a comprehensive benchmark for weak supervision. 
It includes 22 datasets for classification and sequence tagging with a wide range of domains, modalities, and sources of supervision. 
Through extensive comparisons, we conclude that designing general-purpose weak supervision methods still remains challenging. 
We believe that \benchmark provides an increasingly needed foundation for addressing this challenge.
In addition, \benchmark provides procedural labeling function generators for systematic study of various types of weak supervision sources.
Based on the generators, we study a range of aspects of weak supervision, in order to help understand the weak supervision problem and motivate future research directions.

For future work, we plan to include more weak supervision methods and novel tasks, covering more aspects of weak supervision. 
Specifically, we plan to incorporate following aspects:

\noindent \textbf{(1) Learning the dependency structure of supervision sources.}
The dependency structure among supervision sources is frequently ignored in applications of weak supervision. 
However, as reported in \cite{MisspecificationInDP}, this unawareness and consequent dependency structure misspecification could result in a serious performance drop.
To address this, several approaches have been proposed~\cite{Bach2017LearningTS,Varma2017InferringGM,Varma2019LearningDS}, but a benchmark for this purpose is missing.
To complement this, we plan to add more datasets with varying dependency structure for benchmarking the dependency structure learning in weak supervision.

\noindent \textbf{(2) Active generation and repurposing of supervision sources.}
To further reduce human annotation efforts, very recently, researchers turn to active generation~\cite{varma2018snuba,TALLOR,glara,boecking2021interactive,darwin} and repurposing~\cite{DBLP:conf/naacl/GoelORVR21} of supervision sources.
In the future, we plan to incorporate these new tasks and methods into \benchmark to extend its scope.

\noindent \textbf{(3) More applications of weak supervision.}
\benchmark focus on two applications of weak supervision: classification and sequence tagging. 
To unleash the potential of weak supervision and push the community to move forward, we plan to add more applications into \benchmark in the future.

Finally, the \textbf{model performance prediction} based on properties of dataset and supervision sources is of great importance yet challenging and open.
We believe the labeling function generator in \benchmark and the new proposed measurements of supervision sources, \ie, correlation and data-dependency, would contribute to this goal.

\printbibliography


\appendix

\clearpage

\section{Key Information}

\subsection{Dataset Documentations}

The dataset is provided in \emph{json} format; there are three json files corresponding to the train, validation and test split.

Each data point contains the following fields:
\begin{itemize}
    \item \texttt{id}: unique identifier for the example;
    \item \texttt{label}: the label of the example;
    \item \texttt{weal\_labels}: the output of the labeling functions;
    \item \texttt{data}: a dictionary contains the raw data;
\end{itemize}

Details of each dataset can be found in App.~\ref{sec:dataset}.

\subsection{Intended Uses}
\benchmark is intended for researchers in machine learning and related fields to innovate novel methods for the weak supervision problem and data scientists to apply machine learning algorithms which require manual annotations.

\subsection{Hosting and Maintenance Plan}
\benchmark codebase is hosted and version-tracked via GitHub.
It will be permanently available under the link \url{https://github.com/JieyuZ2/wrench}.
The download link of all the datasets can be found in the Github repository.

\benchmark is a community-driven and open-source initiative. 
We are committed and has resources to maintain and actively develop \benchmark for at minimum the next five years. 
We plan to grow \benchmark by including new learning tasks and datasets. 
We welcome external contributors.

\subsection{Licensing}
We license our work using Apache 2.0\footnote{https://www.apache.org/licenses/LICENSE-2.0}.
All the datasets are publicly released by previous work.

\subsection{Author Statement}
 We the authors will bear all responsibility in case of violation of rights.

\subsection{Limitations}
Weak supervision is an increasing field, and there are important tasks and datasets yet to be included in \benchmark. 
However, \benchmark is an ongoing effort and we plan to continuously include more datasets and tasks in the future.

\subsection{ Potential Negative Societal Impacts}
\benchmark does not involve human subjects research and does not contain any personally identifiable information.
Possible misuse may lead to negative outcomes, such as direct usage of the model predictions to detect spam message without prior rigorous validation of the model performance.

\section{Real-world Datasets}
\label{sec:dataset}

\subsection{Detailed Statistics and Visualization}

\begin{figure*}[h]
  \centering
  \includegraphics[width=1.0\textwidth]{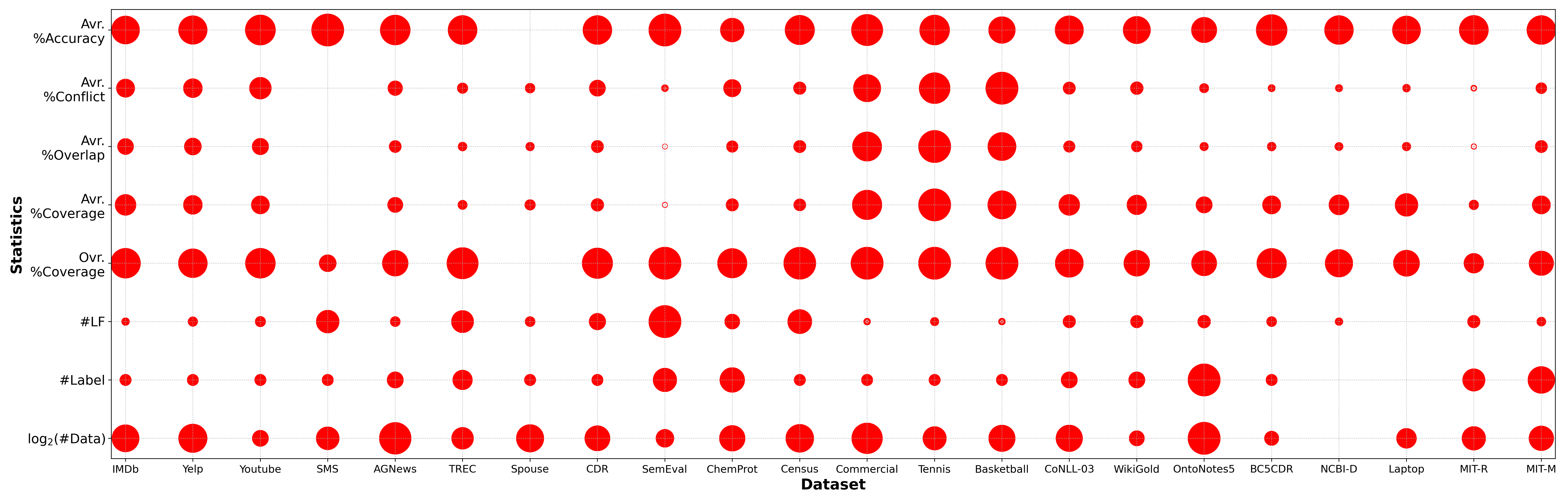}
  \caption{Visualization of all statistics of datasets.}
  \label{fig:dataset_stats_bubble}
\end{figure*}

We provide the detailed statistics of real-world datasets in Table~\ref{tab:dataset_stats_class}-\ref{tab:dataset_stats_seq}.
We also visualize the dataset statistics in Fig.~\ref{fig:dataset_stats_bubble}, where each value is normalized to [0, 1] range across datasets.

\begin{table*}[h]
    \centering
    \caption{Detailed statistics of classification datasets included in \benchmark.}
    \scalebox{0.5}{
    \begin{tabular}{ l l l c c c c c c c c c c}
        \toprule
         &
         \multicolumn{5}{c}{} &
         \multicolumn{4}{c}{\textbf{Avr. over LFs}}    &
         \multicolumn{1}{c}{\textbf{Train}}    &
         \multicolumn{1}{c}{\textbf{Dev}}    &
         \multicolumn{1}{c}{\textbf{Test}}   \\
         \cmidrule(lr){7-10}
         \cmidrule(lr){11-11}
         \cmidrule(lr){12-12}
         \cmidrule(lr){13-13}
           \textbf{Task ($\downarrow$)} &\textbf{Domain ($\downarrow$)} & \textbf{Dataset ($\downarrow$)} & \textbf{\#Label} & \textbf{\#LF} & \textbf{Ovr. \%Coverage}& \textbf{\%Coverage} & \textbf{\%Overlap} & \textbf{\%Conflict} & \textbf{\%Accuracy} & \textbf{\#Data} & \textbf{\#Data} & \textbf{\#Data}   \\
         \midrule
     \multirow{1}{*}{Income Classification}    & Tabular Data & Census~\cite{kohavi1996scaling, Awasthi2020Learning}  & 2 & 83 & 99.13 & 5.41 & 5.34 & 1.50 & 78.74 & 10,083  & 5,561 &  16,281 \\\midrule[0.05pt] \midrule[0.05pt]
     
    \multirow{2}{*}{Sentiment Classification}    

     & Movie & IMDb~\cite{IMDB,ren2020denoising}  & 2 & 5 & 87.58 & 23.60 & 11.60 & 4.50 & 69.88 & 20,000 & 2,500 & 2,500 \\      
     
     & Review & Yelp~\cite{AGNews,ren2020denoising}  & 2 & 8 & 82.78 & 18.34 & 13.58 & 4.94 & 73.05 & 30,400  & 3,800 & 3,800      \\\midrule[0.05pt] \midrule[0.05pt]
     
    \multirow{2}{*}{Spam Classification }    & Review & Youtube~\cite{youtube}  & 2 & 10 & 87.70 & 16.34 & 12.49 & 7.14 & 83.16 & 1,586 & 120 & 250 \\      
    
        & Text Message & SMS~\cite{sms, Awasthi2020Learning}  & 2 & 73 & 40.52 & 0.72 & 0.29 & 0.01 & 97.26 & 4,571 & 500 & 2719  \\\midrule[0.05pt] \midrule[0.05pt]
        
    \multirow{1}{*}{Topic Classification}    & News & AGNews~\cite{AGNews,ren2020denoising}  & 4 & 9 & 69.08 & 10.34 & 5.05 & 2.43 & 81.66 & 96,000 & 12,000 & 12,000  \\\midrule[0.05pt] \midrule[0.05pt]
    
    \multirow{1}{*}{Question Classification}    & Web Query & TREC~\cite{trec, Awasthi2020Learning}  & 6 & 68 & 95.13 & 2.55 & 1.82 & 0.84 & 75.92 & 4,965 & 500 & 500  \\\midrule[0.05pt] \midrule[0.05pt]
    
     \multirow{4}{*}{Relation Classification}    & News & Spouse~\cite{spouse, ratner2017snorkel}  & 2 & 9 & 25.77 & 3.75 & 1.66 & 0.65 & -- & 22,254 & 2,811 & 2,701       \\
     
     & Biomedical & CDR~\cite{davis2017comparative, ratner2017snorkel}  & 2 & 33 & 90.72 & 6.27 & 5.36 & 3.21 & 75.27 & 8,430 & 920 & 4,673  \\     
     & Web Text & SemEval~\cite{hendrickx2010semeval, zhou2020nero}  & 9 & 164 & 100.00 & 0.77 & 0.32 & 0.14 & 97.69 & 1,749 & 200 & 692   \\
     
     & Chemical & ChemProt~\cite{chemprot,yu-etal-2021-fine}  & 10 & 26 & 85.62 & 5.93 & 4.40 & 3.95 & 46.65 & 12,861 & 1,607 & 1,607 \\\midrule[0.05pt] \midrule[0.05pt]
     
     \multirow{3}{*}{Image Classification}  & \multirow{3}{*}{Video} & Commercial~\cite{fu2020fast}  &  2 & 4 & 100.00 & 54.51 & 53.09 & 12.51 & 91.33 & 64,130  & 9,479 & 7,496  \\     
     
     &  & Tennis Rally~\cite{fu2020fast}  &  2 & 6 & 100.00 & 66.86 & 66.86 & 16.76 & 81.70 & 6,959    &   746 & 1,098     \\
     
     &  & Basketball~\cite{fu2020fast}  &  2 & 4 & 100.00 & 49.24 & 48.46 & 18.50 & 62.04 & 17,970    & 1,064 & 1,222    \\
    \bottomrule
    \end{tabular}
    }

    \label{tab:dataset_stats_class}
\end{table*}

\begin{table*}[h]
    \centering
    \caption{Detailed statistics of sequence tagging datasets included in \benchmark.}
    \scalebox{0.5}{
    \begin{tabular}{ l l c c c c c c c c c c}
        \toprule
         \multicolumn{5}{c}{} &
         \multicolumn{4}{c}{\textbf{Avr. over LFs}}    &
         \multicolumn{1}{c}{\textbf{Train}}    &
         \multicolumn{1}{c}{\textbf{Dev}}    &
         \multicolumn{1}{c}{\textbf{Test}}   \\
         \cmidrule(lr){6-9}
         \cmidrule(lr){10-10}
         \cmidrule(lr){11-11}
         \cmidrule(lr){12-12}
         \textbf{Domain ($\downarrow$)} & \textbf{Dataset ($\downarrow$)} & \textbf{\#Label} & \textbf{\#LF} & \textbf{Ovr. \%Coverage}& \textbf{\%Coverage} & \textbf{\%Overlap} & \textbf{\%Conflict} & \textbf{\%Precision} & \textbf{\#Data} & \textbf{\#Data} & \textbf{\#Data}   \\
         \midrule
         
    News & CoNLL-03~\cite{conll03,lison2021skweak}             & 4 & 16 & 79.51  & 23.71  & 4.30  & 1.44 & 72.19 & 14,041 & 3250 & 3453    \\\midrule[0.05pt] \midrule[0.05pt]
    
      \multirow{2}{*}{Web Text} & WikiGold~\cite{wikigold,lison2021skweak}       & 4 & 16 & 69.68 & 20.30  & 3.65 & 1.61 & 65.87 &  1,355 & 169 & 170     \\
     & OntoNotes 5.0~\cite{weischedel2011ontonotes}  & 18 & 17 & 66.79  & 12.45 & 1.55  & 0.54  & 54.84 & 115,812 & 5,000 & 22,897\\  \midrule[0.05pt] \midrule[0.05pt]  
     
     \multirow{2}{*}{Biomedical}  & BC5CDR~\cite{cdr,li2021bertifying}            & 2 & 9  & 86.62 & 16.75 & 1.77 & 0.17 & 88.23 & 500 & 500 & 500    \\
     & NCBI-Disease~\cite{dougan2014ncbi,li2021bertifying} & 1 & 5 & 77.15 & 21.16 & 1.40 & 0.18 & 74.88 &  592 & 99 & 99   \\\midrule[0.05pt] \midrule[0.05pt]
     
     \multirow{2}{*}{Review} & Laptop-Review~\cite{laptop,li2021bertifying}      & 1 & 3  & 70.62 & 29.37 & 1.65 & 0.25 & 70.30 & 2,436 & 609 & 800     \\
     & MIT-Restaurant~\cite{mitr,Awasthi2020Learning}       & 8 & 16 & 47.84 & 2.87 & 0.37 & 0.06 & 76.65 & 7,159 &  500   &   1,521     \\\midrule[0.05pt] \midrule[0.05pt]
     
     Movie Query & MIT-Movies~\cite{mitmovie}        & 12 & 7 & 64.14 & 16.60 & 5.29 & 0.97 & 75.10 & 9,241 &  500   &      2,441  \\
    \bottomrule
    \end{tabular}
    }

    \label{tab:dataset_stats_seq}
\end{table*}

\subsection{Classification Datasets}
\textbf{Census~\cite{kohavi1996scaling}}.
This UCI dataset is extracted from the 1994 U.S. census. It
lists a total of 13 features of an individual such as age, education level, marital status, country of origin etc. The primary task on it is binary classification - whether a person earns more than 50K or not. The train data consists of 32,563 records. The labeling functions are generated synthetically by \cite{Awasthi2020Learning} as follows:
We hold out disjoint 16k random points from the training dataset as a proxy for human knowledge and extract a PART decision list~\cite{frank1998generating} from it as labeling functions.

\textbf{SMS~\cite{sms}}. 
This dataset contains 4,571 text messages labeled as spam/not-spam, out of which 500 were held out for validation and 2719 for testing. The labeling functions are generated manually by \cite{Awasthi2020Learning}, including 16 keyword-based and 57 regular expression-based rules.

\textbf{AGNews~\cite{AGNews}}.
This dataset is a collection of more than one million news articles. It is constructed by \cite{ren2020denoising} choosing the 4 largest topic classes from the original corpus. The total number of training samples is 96K and both validation and testing are 12K. The labeling functions are also generated by \cite{ren2020denoising}, including 9 keyword-based rules.

\textbf{Yelp~\cite{AGNews}}.
This dataset is a subset of Yelp's businesses, reviews, and user data for binary sentiment classification. It is constructed by \cite{ren2020denoising}, including 30.4K training samples, 3.8K validation samples and 3.8K testing samples. The labeling functions are also generated by \cite{ren2020denoising}, including 7 heuristic rules on keywords and 1 third-party model on polarity of sentiment.

\textbf{Youtube~\cite{youtube}}.
This dataset is a public set of comments collected for spam detection. It has five datasets composed of 1,586 real messages extracted from five videos. The number of validation samples is 120 and that of testing samples is 250. The labeling functions are generated manually by \cite{ratner2017snorkel}, including 5 keyword-based, 1 regular expression-based, 1 heuristic, 1 complex preprocessors, and 2 third-party model rules.

\textbf{IMDb~\cite{IMDB}}.
This is a dataset for binary sentiment classification containing a set of 20,000 highly polar movie reviews for training, 2,500 for validation and 2,500 for testing. It is constructed by \cite{ren2020denoising}. The labeling functions are also generated by \cite{ren2020denoising}, including 4 heuristic rules on keywords and 1 heuristic rules on expressions.

\textbf{TREC~\cite{trec}}.
This dataset contains 4,965 labeled questions in the training set, 500 for validation set and another 500 for the testing set. It has 6 classes. The labeling functions are generated by \cite{Awasthi2020Learning}, including 68 keyword-based rules.

\textbf{Spouse~\cite{spouse}}.
This dataset is constructed by \cite{ratner2017snorkel} and to identify mentions of spouse relationships in a set of news articles from the Signal Media~\cite{spouse}.
It contains 22,254 training samples 2,811 validation samples and 2,701 testing samples. The labeling functions are generated by Snorkel\footnote{\url{https://github.com/snorkel-team/snorkel-tutorials/tree/master/spouse}}. Note that the gold labels for the training set is not available. Therefore, we are unable to calculate the accuracy for labeling functions on the training set.

\textbf{CDR~\cite{cdr}}.
This dataset is constructed by \cite{ratner2017snorkel},  where the
task is to identify mentions of causal links between chemicals and diseases in PubMed abstracts. It has 8,430 training samples 920 validation samples and 4,673 testing samples. The labeling functions can be found in Snorkel tutorial\footnote{\url{https://github.com/snorkel-team/snorkel-extraction/tree/master/tutorials/cdr}}.

\textbf{SemEval~\cite{hendrickx2010semeval}}.
This relation classification dataset is constructed by \cite{zhou2020nero} with 9 relation types. The size of the training, validation and test set are 1,749, 200 and 692 respectively. The labeling functions are generated by \cite{zhou2020nero}, including 164 heuristic rules.

\textbf{ChemProt~\cite{chemprot}}.
This is a 10-way relation classification dataset constructed by \cite{yu-etal-2021-fine}, containing 12,861 training samples, 1,607 validation samples and 1,607 testing samples. The labeling functions are generated by \cite{yu-etal-2021-fine}, including 26 keyword-based rules.

\textbf{Basketball, Commercial, Tennis Rally~\cite{fu2020fast}}.
These datasets are video frame classification datasets collected by~\cite{fu2020fast}. All the labeling functions are the same as previous work~\cite{fu2020fast}.
Due to privacy issues, we only have access to the features extracted by a ResNet-101 model pre-trained on ImageNet.

\subsection{Sequence Tagging Datasets}
\textbf{CoNLL-03~\cite{conll03}}. This is a well-known open-domain NER dataset from
the CoNLL 2003 Shared Task. It consists of 1393 English news
articles and is annotated with 4 entity types: \emph{person}, \emph{location}, \emph{organization}, and \emph{miscellaneous}\footnote{In the original dataset, it has \texttt{-DOCSTART-} lines to separate documents, but these lines are removed here.}. Note that different papers~\cite{peng2019distantly,li2021bertifying,lison2020named} use different weak supervision sources with varying quality. In our study, we use the labeling function generated by~\cite{lison2020named} for fair comparison.
(We use \texttt{BTC}, \texttt{core\_web\_md}, \texttt{crunchbase\_cased}, \texttt{crunchbase\_uncased}, \texttt{full\_name\_detector}, \texttt{geo\_cased}, \texttt{geo\_uncased}, \texttt{misc\_detector},  \texttt{wiki\_cased},  \texttt{wiki\_uncased}, \texttt{multitoken\_crunchbase\_cased}, \texttt{multitoken\_crunchbase\_uncased}, \texttt{multitoken\_geo\_cased}, \texttt{multitoken\_geo\_uncased}, \texttt{multitoken\_wiki\_cased}, \texttt{multitoken\_wiki\_uncased} as weak supervision sources)

\textbf{BC5CDR~\cite{cdr}}. This dataset accompanies the BioCreative V CDR challenge and consists of 1,500 PubMed articles and is annotated with \emph{chemical} and \emph{disease} mentions. The labeling functions  are selected from~\cite{li2021bertifying}. (We use \texttt{DictCore-Chemical}, \texttt{DictCore-Chemical-Exact}, \texttt{DictCore-Disease}, \texttt{DictCore-Disease-Exact}, \texttt{ Element, Ion, or Isotope}, \texttt{Organic Chemical}, \texttt{Antibiotic}, \texttt{Disease or Syndrome}, \texttt{PostHyphen}, \texttt{ExtractedPhrase} as weak supervision sources.)

\textbf{NCBI-Disease~\cite{dougan2014ncbi}}. This dataset includes 793 PubMed abstracts annotated with \emph{disease} mentions only.  The labeling functions are the same as~\cite{li2021bertifying}.

\textbf{Laptop-Review~\cite{laptop}}. This dataset is from the SemEval 2014 Challenge, Task 4 Subtask 1 and consists of 3,845 sentences with \emph{laptop}-related entity mentions. The labeling functions are selected from~\cite{li2021bertifying}. (We use \texttt{CoreDictionary}, \texttt{ExtractedPhrase}, \texttt{ConsecutiveCapitals} as weak supervision sources.)

\textbf{Wikigold~\cite{wikigold}}. This dataset contains a set of Wikipedia articles (40k tokens) randomly selected from a 2008 English dump and manually
annotated with the four CoNLL-03 entity types. Since the label type of Wikigold is the same as CoNLL-03, we also use the labeling function provided in~\cite{lison2020named}. (We use \texttt{BTC}, \texttt{core\_web\_md}, \texttt{crunchbase\_cased}, \texttt{crunchbase\_uncased}, \texttt{full\_name\_detector}, \texttt{geo\_cased}, \texttt{geo\_uncased}, \texttt{misc\_detector},  \texttt{wiki\_cased},  \texttt{wiki\_uncased}, \texttt{multitoken\_crunchbase\_cased}, \texttt{multitoken\_crunchbase\_uncased}, \texttt{multitoken\_geo\_cased}, \texttt{multitoken\_geo\_uncased}, \texttt{multitoken\_wiki\_cased}, \texttt{multitoken\_wiki\_uncased} as weak supervision sources).

\textbf{MIT-Restaurant~\cite{mitr}}. This is a slot-filling dataset includeing sentences about restaurant search queries. It contains 8 entity types with 9180 examples. We follow the data split by~\cite{Awasthi2020Learning} and use regular expression in their paper as weak supervision. Besides, we also extract \texttt{restaurant names} and \texttt{cuisines} from yelp database\footnote{\url{https://www.yelp.com/dataset}} to augment the labeling function.\footnote{In~\cite{Awasthi2020Learning,karamanolakis2021self,yu-etal-2021-fine}, they treat this dataset as token-level classification problem and use \emph{token-level} F1 score for evaluation, which is in conflict with the original evaluation protocol of the dataset~\cite{mitr}.}

\textbf{MIT-Movies~\cite{mitmovie}}. This dataset includes sentences on movie search queries with 12 entity types. For this dataset, we curate the weak supervision via several class-related keywords, semantic patterns based on regular expressions (listed in table~\ref{tab:mitmovies}) and knowledge-base matching. Specifically, we collect the movie-related information on JsonMC\footnote{\url{https://github.com/jsonmc/jsonmc}}, Movies-dataset\footnote{\url{https://github.com/randfun/movies-dataset}}, IMDB\footnote{\url{https://www.imdb.com/}} and Wikidata\footnote{\url{https://www.wikidata.org/}}. There are 7 weak supervision sources in total.

\textbf{Ontonotes 5.0~\cite{weischedel2011ontonotes}}. This is a fine-grained NER dataset with text documents from multiple domains, including broadcast conversation, P2.5 data and Web data. It consists of 113 thousands of training data and is annotated with 18 entity types. We adopt a set of the weak supervision sources presented in Skweak\footnote{\url{https://github.com/NorskRegnesentral/skweak}}~\cite{lison2021skweak} including \texttt{money\_detector}, \texttt{date\_detector}, \texttt{number\_detector}, \texttt{company\_type\_detector}, \texttt{full\_name\_detector}, \texttt{crunchbase\_cased}, \texttt{crunchbase\_uncased}, \texttt{geo\_cased}, \texttt{geo\_uncased}, \texttt{misc\_detector},  \texttt{wiki\_cased},  \texttt{wiki\_uncased} (12 in total). 
Since some of the weak supervision sources in Skweak  only have coarse-level annotation (e.g. they can only label for entity types listed in CoNLL-03: \emph{person}, \emph{location}, \emph{organization}, and \emph{miscellaneous}), we follow the method used in~\cite{liang2020bond} to use SPARQL to query the categories of an entity in the knowledge ontology in wikipedia. 
Apart from it, we also extracted multi-token phrases from wikidata knowledge base and match with the corpus. Finally, we include several class-related keywords, and regular expressions (listed in table \ref{tab:ontonotes}) as the labeling functions. 
As a result, there are 17 weak supervision sources. To control the size for the validation set, we only use the first 5000 sentences as the validation set and put others into the test set.

\begin{table*}[htb!]\small
    \centering
    \caption{Examples of labeling functions on MIT-Movies.}
    \vskip -0.1in
    \begin{tabular}
    {p{400pt} }
    \toprule
        \textbf{Labeling Functions} \\
        \midrule
        \textbf{LF \#1: Key words:} \\
        \midrule
    \texttt{[`tom hanks', `jennifer lawrence', `tom cruises', `tom cruse', `clint eastwood', `whitney houston', `leonardo dicaprio', `jennifer aniston', `kristen stewart'] $\rightarrow$ ACTOR}
     \\ \hline
    \texttt{[`'(18|19|20)$\backslash$d{2}', `'(18|19|20)$\backslash$d{2}s', `last two years', `last three years'] $\rightarrow$ YEAR} \\ \hline
    \texttt{[`batman', `spiderman', `rocky', `twilight', `titanic', `harry potter', `ice age', `speed', `transformers', `lion king', `pirates of the caribbean']$\rightarrow$ TITLE}
     \\ \hline
    \texttt{ [`i will always love you', `endless love', `take my breath away', `hes a tramp', `beatles', `my heart will go on', `theremin', `song', `music'] $\rightarrow$ SONG}  \\ \hline
    \texttt{[`comedy',`comedies', `musical', `romantic', `sci fi', `horror', `cartoon', `thriller', `action', `documentaries', `historical', `crime', `sports', `classical']$\rightarrow$ GENRE}  \\ \hline
     \texttt{[`x', `g', `r', `pg', `rated (x|g|r)', `pg rated', `pg13', `nc17'] $\rightarrow$ RATING} \\ \hline
     \texttt{[`(five|5) stars', `(highest|lowest) rated', `(good|low) rating', `mediocre'] $\rightarrow$ RATINGS\_AVERAGE} \\ \hline
     \texttt{[`classic', `recommended', `top', `review', `awful', `reviews', `opinions', `say about', `saying about', `think about'] $\rightarrow$ REVIEW}\\ \hline
     \texttt{[`trailers', `trailer', `scene', `scenes', `preview', `highlights'] $\rightarrow$ TRAILER}\\ \hline
    \texttt{[`007', `james bond', `ron weasley', `simba', `peter pan', `santa clause', `mr potato head', `buzz lightyear', `darth bader', `yoda', `dr evil'] $\rightarrow$ CHARACTER}\\  \midrule
     \textbf{LF \#2: Context Patterns (The NOUN matched in * will be recognized as the entity with the target type):} \\ \midrule
      \texttt{[`movie named *', `film named *', `movie called', `film called *',  `movie titled *', `film titled *'] $\rightarrow$ CHARACTER}\\ \hline 
      \texttt{[`(starring|starred|stars) *',  `featured *'] $\rightarrow$ ACTOR}
      \texttt{[`(about|about a|about the) *',`set in *'] $\rightarrow$ PLOT}\\ \hline 
      \texttt{[`directed by *',`produced by *'] $\rightarrow$ DIRECTOR}\\ \hline 
      \texttt{[`a rating of *'] $\rightarrow$ RATINGS\_AVERAGE}\\ \hline 
      \texttt{[`during the *',`in the *'] $\rightarrow$ YEAR}\\ \hline 
      \texttt{[`the song *',`a song *'] $\rightarrow$ SONG}\\ \hline 
      \texttt{[`pg *',`nc *'] $\rightarrow$ RATING}\\ \hline 
      
     \textbf{LF \#3: Regex Patterns (The entity matched in \underline{underlined} * will be recognized as the entity with the target type):} \\ \midrule
      \texttt{['(is|was|has) \underline{[$\backslash$w$\backslash$s]*} (ever been in|in|ever in) (a|an) [$\backslash$w]* (movie|film)'] $\rightarrow$ ACTOR}\\ \hline 
      \texttt{['(show me|find) (a|the) \underline{[ˆ$\backslash$w]*} (from|for)'] $\rightarrow$ TRAILER} \\ \hline 
      \texttt{[`(a|an) \underline{[$\backslash$w]*} (movie|film)'] $\rightarrow$ GENRE}\\ \hline 
      \texttt{[`(did|does) \underline{[ˆ$\backslash$w|$\backslash$s]*} direct'] $\rightarrow$ DIRECTOR}\\ \hline 
      \texttt{[`(past|last) \underline{([0-9]*|[ˆ$\backslash$w]) years}'] $\rightarrow$ YEAR}\\ \hline 
      \texttt{[ˆ\underline{$\backslash$d\{1\} stars"}] $\rightarrow$ RATINGS\_AVERAGE}\\ \hline 
    \bottomrule
    \end{tabular}
    \label{tab:mitmovies}
\end{table*}

\begin{table*}[htb!]\small
    \centering
    \caption{Examples of labeling functions on Ontonotes.}
    \vskip -0.1in
    \begin{tabular}
    {p{400pt} }
    \toprule
        \textbf{Labeling Functions} \\
        \midrule
        \textbf{LF \#1: Key words:} \\
        \midrule
    \texttt{[`Taiwan', `Hong Kong', `China', `Japan', `America', `Germany', `US', `Singapore"] $\rightarrow$ GPE}
     \\ \hline
    \texttt{[`World War II', `World War Two','nine eleven', `the Cold War', `World War I', `Cold War', `New Year', `Chinese New Year', `Christmas"] $\rightarrow$ EVENT} \\ \hline
    \texttt{[`the White House', `Great Wall', `Tiananmen Square', `Broadway"]$\rightarrow$ FAC}
     \\ \hline
    \texttt{ [ `Chinese', `Asian', `American', `European', `Japanese', `British', `French', `Republican"] $\rightarrow$ NORP}  \\ \hline
    \texttt{[`English', `Mandarin', `Cantonese"]$\rightarrow$ LANG}  \\ \hline
     \texttt{[`Europe', `Asia', `North America', `Africa`, `South America'] $\rightarrow$ LOC} \\ \hline
     \texttt{[`WTO', `Starbucks', `mcdonald', `google', `baidu', `IBM', `Sony', `Nikon'] $\rightarrow$ ORG} \\ \hline
     \texttt{['toyota', `discovery', `Columbia', `Cocacola', `Delta', `Mercedes', `bmw'] $\rightarrow$ PRODUCT}\\ \hline
     \texttt{[`this year', `last year', `last two years', `last three years', `recent years', `today', `tomorrow', `(18|19|20)$\backslash$\{d\}{2}'] $\rightarrow$ DATE}\\ \hline
    \texttt{[`(this|the|last) (morning|evening|afternoon|night)'] $\rightarrow$ TIME}\\ \hline
    \texttt{[·first', `second’, `third‘, `fourth‘, `fifth', `sixth', `seventh‘, `ninth’, `firstly', `secondly', `1 st', `2 nd', `(3|...|9) th'] $\rightarrow$ ORDINAL} \\ \midrule
    
     \textbf{LF \#2: Numerical Patterns (The numerical value end with the terms in the list below will be recognized as the entity with the target type):} \\ \midrule
      \texttt{`dollars', `dollar', `yuan', `RMB', `US dollar', `Japanese yen', 'HK dollar', `Canadian dollar', `Australian dollar', `lire', `francs'] $\rightarrow$ MONEY}\\ \hline 
      \texttt{[`the (18|19|20)$\backslash$\{d\}{2}s', `the (past|last|previous) (one|two|three|four|five|six|seven|eight|nine|ten) (decade|day|month|year)', `the [0-9$\backslash$w]+ century'] $\rightarrow$ DATE}\\ \hline 
      \texttt{[`tons', `tonnes', `barrels', `m', `km', `mile', `miles', `kph', `mph', `kg', `°C', `dB', `ft', `gal', `gallons', `g', `kW', `s', `oz',
        `m2', `km2', `yards', `W', `kW', `kWh', `kWh/yr', `Gb', `MW', `kilometers',"square meters', `square kilometers', `meters', `liters', `litres', `g', `grams', `tons/yr',
        'pounds', 'cubits', 'degrees', 'ton', 'kilograms', 'inches', 'inch', 'megawatts', 'metres', 'feet', 'ounces', 'watts', 'megabytes',
        'gigabytes', 'terabytes', 'hectares', 'centimeters', 'millimeters'] $\rightarrow$ QUANTITY}\\ \hline 
      \texttt{[`year', `years', `ago', `month', `months', `day', `days', `week', `weeks', `decade'] $\rightarrow$ DATE}\\ \hline 
      \texttt{[`$\%$', 'percent'] $\rightarrow$ PERCENT} \\ \hline 
      \texttt{[`AM', `PM', `p.m', `a.m', `hours', `minutes', `EDT', `PDT', `EST', `PST"] $\rightarrow$ TIME}\\ \midrule
      
     \textbf{LF \#3: Regex Patterns (The term matched in * will be recognized as the entity with the target type):} \\ \midrule
      \texttt{`(NT|US)$\$$ [$\backslash$w,]* (billion|million|trillion)', `[0-9.]+ (billion yuan|yuan|million yuan)',"[0-9.]+ (billion yen|yen|million yen)', `[0-9.]+ (billion US dollar|US dollar|million US dollar)',"[0-9.]+ (billion HK dollar|HK dollar|million HK dollar)',\
     `[0-9.]+ (billion francs|francs|million francs)', `[0-9.]+ (billion cent|cent|million cent)', `[0-9.]+ (billion marks|marks|million marks)', `[0-9.]+ (billion Swiss francs|Swiss francs|million Swiss francs)', `$\$$ [0-9.,]+ (billion|million|trillion)', `about $\$$ [0-9.,]+"] $\rightarrow$ MONEY}\\ \hline 
      \texttt{[ `the (18|19|20)$\backslash$d{2}s', `the (past|last|previous) (one|two|three|four|five|six|seven|eight|nine|ten) (decade|day|month|year)', `the [0-9$\backslash$w]+ century"] $\rightarrow$ DATE}\\ \hline 
      \texttt{[`(t|T)he [$\backslash$w ]+ Standard', `(t|T)he [$\backslash$w ]+ Law', `(t|T)he [$\backslash$w ]+ Act', `(t|T)he [$\backslash$w ]+ Constitution',`set in *'] $\rightarrow$ LAW}\\ \hline 
      \texttt{[`[0-9.]+ ($\%$|percent)'] $\rightarrow$ PERCENT}\\ \hline 
      \texttt{[`$\backslash$d\{1\}:$\backslash$d\{2\} (am|a.m.|pm|p.m.)', `$\backslash$d\{2\}:$\backslash$d\{2\} (am|a.m.|pm|p.m.)'] $\rightarrow$ TIME} \\
    \bottomrule
    \end{tabular}
    \label{tab:ontonotes}
\end{table*}
\section{Compared Methods}
\label{sec:methods}

\subsection{Classification}

\subsubsection{Label Model}

\textbf{MV / WMV:} We adopt the classic majority voting (MV) algorithm as one label model, as well as its extension weighted majority voting (WMV) where we reweight the final votes by the label prior. Notably, the abstaining LF, \ie, $\lambda_j = -1$ won't contribute to the final votes.

\textbf{DS~\cite{DawidSkene}:} Dawid-Skene (DS) model estimates the accuracy of each LF with expectation maximization (EM) algorithm by assuming a naive Bayes distribution over the LFs’ votes and the latent ground truth.

\textbf{DP~\cite{Ratner16}:} Data programming (DP) models the distribution $p(L, Y)$ as a factor graph. It is able to describe the distribution in terms of pre-defined factor functions, which reflects the dependency of any subset of random variables. The log-likelihood is optimized by SGD where the gradient is estimated by Gibbs sampling, similarly to contrastive divergence~\cite{HintonCD}.

\textbf{MeTaL~\cite{Ratner19}:} MeTal models the distribution via a Markov Network and recover the parameters via a matrix completion-style approach.
Notably, it requires label prior as input.

\textbf{FS~\cite{fu2020fast}:} FlyingSquid (FS) models the distribution as a binary Ising model, where each LF is represented by two random variables.
A Triplet Method is used to recover the parameters and therefore no learning is needed, which makes it much faster than data programming and MeTal.
Notably, FlyingSquid is designed for binary classification and the author suggested applying a one-versus-all reduction repeatedly to apply the core algorithm.
The label prior is also required.

\subsubsection{End Model}

\textbf{LR:} We choose Logistic Regression (LR) as an example of linear model. 

\textbf{MLP:} For non-linear model, we take Multi-Layer Perceptron Neural Networks (MLP) as an example.

\textbf{BERT~\cite{devlin2019bert} / RoBERTa~\cite{liu2019roberta}:} It is also interested how recent large scale pretrained language models perform as end models for textual datasets, so we include both BERT\cite{devlin2019bert} and RoBERTa~\cite{liu2019roberta}, shortened as \textbf{B} and \textbf{R} respectively.
Notably, these pretrained language models can only work for textual datasets, and for text relation classification task, we adopt the R-BERT~\cite{wu2019enriching} architecture.

\textbf{COSINE~\cite{yu-etal-2021-fine}}: COSINE uses self-training and contrastive learning to bootstrap over unlabeled data for improving a pretrained language model-based end model.
We denote the BERT-based COSINE and the RoBERTa-based COSINE by \textbf{BC} and \textbf{RC} respectively.

\subsubsection{Joint Model}

\textbf{Denoise~\cite{ren2020denoising}:} Denoise adopts an attention network to aggregate over weak labels, and use a neural classifier to leverage the data feature. These two components are jointly trained in an end-to-end manner.

\subsection{Sequence Tagging}

\subsubsection{Label Models}

\textbf{HMM~\cite{lison2020named}}:
Hidden Markov models 
\cite{lison2020named} represent true labels as latent variables and inferring them from the independently observed noisy labels through unsupervised learning with expectation-maximization algorithm \cite{welch2003hidden}.

\textbf{CHMM \cite{li2021bertifying}}:
Conditional hidden Markov model (CHMM) \cite{li2021bertifying} substitutes the constant transition and emission matrices by token-wise counterpart predicted from the BERT embeddings of input tokens.
The token-wise probabilities are representative in modeling how the true labels should evolve according to the input tokens.

\subsubsection{End Model}

\textbf{LSTM-CNNs-CRF~\cite{ma2016end}}:
LSTM-CNNs-CRF  
encodes character-level features with convolutional neural networks (CNNs), and use bi-directional long short-term memory (LSTM) network~\cite{hochreiter1997long} to model word-level features.
A conditional random field (CRF) layer~\cite{lafferty2001conditional} is stacked on top of LSTM to impose constraints over adjacent output labels.

\textbf{BERT}:
It use the pre-trained BERT~\cite{devlin2019bert} to leverage the pre-trained context knowledge stored in BERT. In our experiments, we use \texttt{BERT-base-cased} for other datasets as our encoder, and we stack a linear layer to predict token labels. 
We also run experiment on stacking a CRF layer on the top of the model, and report the best performance.

\subsubsection{Joint Model}


\textbf{ConNet~\cite{lan2020connet}}:
Consensus Network (ConNet)  adopts a two-stage training approach for learning with multiple  supervision signals. In the decoupling phase, it trains BiLSTM-CNN-CRF \cite{ma2016end} with multiple parallel CRF layers for each labeling source individually. Then, the aggregation phase aggregates the CRF transitions with attention scores and outputs a unified label sequence.

\section{Adapting Label Model for Sequence Tagging Problem}
\label{sec:adapt}
\subsection{Label Correction Technique}
One of the main difference between sequence tagging and classification is that for sequence tagging, there is a specific type `\texttt{O}' which indicates the token does not belong to any pre-defined types. Therefore, if one token cannot be matched with all labeling functions, it will be automatically labeled as type `\texttt{O}'. 
In our study,  we use a label correction technique to differentiate the type `\texttt{O}' with \texttt{Abstain} as follows:
$$
l_{i,c}= \begin{cases}l_{i,c} & \text { if } L_{i, c}\neq\text{O};  \\ \texttt{Abstain (-1)}  & \text { if } L_{i, c}=\text{ O and } \exists ~ c^{'} \in [1, n] \text{ s.t. } L_{i, c^{'}}\neq\text{O;} \\ \text{O} & \text { otherwise }\end{cases}
$$
We have also tried on another choice that regard the weak label for tokens that cannot be matched with any labeling functions as `\texttt{O}'. The comparison of results is in table~\ref{tab:seq_adapt}.

From the table, it is clear that when regarding unmatched token as type \texttt{O}, it leads to a drastic decrease in the final performance. Specifically, the recall of the model is much lower since most of the tokens will be recognized as \texttt{O} when without label modification. One exception is in DS method, as it achieve better performance on 4 out of 8 datasets without label modification. However, when without label modification is better, the performance gain between the two methods is between 0.56\% -- 5.66\% in terms of F1 score. In contrast, when using label modification is better, the gain on F1 score is much larger, i.e., between 4.76\% -- 42.34\%. Therefore, using label modification is generally a better way to adapt label models for classification to sequence tagging problems, and we use this technique in our experiments by default.

\subsection{Comparision of IO and BIO Tagging Schema}
\label{sec:schema}
We also compare the performance of label model with IO and BIO tagging scheme, as both of them have been adopted in previous studies~\cite{safranchik2020weakly,lison2020named,li2021bertifying}. The comparision result is shown in table~\ref{tab:seq_adapt}. From the result, we find that for datasets that when the number of entity types is small (\eg~ BC5CDR, NCBI-Disease), using IO schema leads to higher F1 score. For other datasets, there is no clear winner, as IO schema excels on Ontonotes and MIT-Restaurants datasets while BIO performs better on the others. To conclude, the optimal tagging scheme is highly data-dependent and  we use IO tagging schema in our experiments.

\begin{table*}[t]
    \centering
    \caption{\textbf{Sequence Tagging.} The comparison of label models with or without label modification and IO/BIO tagging scheme. The number stands for the F1 score (Precision, Recall) with standard deviation in the bracket under each value. Each metric value is averaged over 5 runs.}
    \scalebox{0.41}{
    \begin{tabular}{ l c c c c c c c c c }
        \toprule
           \textbf{End Model ($\downarrow$)} &\textbf{Label Model ($\downarrow$)} & \textbf{CoNLL-03} & \textbf{WikiGold} & \textbf{BC5CDR} & \textbf{NCBI-Disease}& \textbf{Laptop-Review} & \textbf{MIT-Restaurant} & \textbf{MIT-Movies} & \textbf{Ontonotes 5.0}   \\
         \midrule
        \multirow{12}{*}{IO Schema, with Label Modification}    
     & \multirow{2}{*}{MV}     & 60.36(59.06/61.72) & 52.24(48.95/56.00) & 83.49(91.69/76.64) & {78.44(93.04/67.79) } & {73.27(88.86/62.33) } & 48.71(74.25/36.24) & 59.68(69.92/52.05) & 58.85(54.17/64.40)        \\
     & & (0.00) & (0.00) & (0.00) & (0.00) & (0.00) & (0.00) & (0.00) & (0.00) \\
     \cmidrule(lr){2-10} 
     & \multirow{2}{*}{WMV}    & 60.26(59.03/61.54) & 52.87(50.74/55.20) & 83.49(91.66/76.66) & {78.44(93.04/67.79) } & {73.27(88.86/62.33) } & {48.19(73.73/35.80) } & 60.37(70.98/52.52) & 57.58(53.15/62.81)         \\ 
     & & (0.00) & (0.00) & (0.00) & (0.00) & (0.00) & (0.00) & (0.00) & (0.00) \\
     \cmidrule(lr){2-10} 
     & \multirow{2}{*}{DS}     & 46.76(45.29/48.32) & 42.17(40.05/44.53) & 83.49(91.66/76.66) & {78.44(93.04/67.79) } & {73.27(88.86/62.33) } & 46.81(71.71/34.75) & 54.06(63.64/46.99) & 37.70(34.33/41.82)      \\
     & & (0.00) & (0.00) & (0.00) & (0.00) & (0.00) & (0.00) & (0.00) & (0.00) \\
     \cmidrule(lr){2-10} 
     & \multirow{2}{*}{DP}      & 62.43(61.62/63.26) & 54.81(53.10/56.64) & {83.50(91.69/76.65) } & {78.44(93.04/67.79) } & {73.27(88.86/62.33) } & 47.92(73.24/35.61) & 59.92(70.65/52.01) & {61.85(57.44/66.99) }        \\
     & & (0.22) & (0.13) & (0.00) & (0.00) & (0.00) & (0.00) & (0.43) & (0.19) \\
     \cmidrule(lr){2-10} 
     & \multirow{2}{*}{MeTaL}   & 60.32(59.07/61.63) & 52.09(50.31/54.03) & 83.50(91.66/76.67) & {78.44(93.04/67.79) } & 64.36(83.21/53.63) & 47.66(73.40/35.29) & 56.60(72.28/47.70) & 58.27(54.10/63.14)         \\
     & & (0.08) & (0.23) & (0.00) & (0.00) & (17.81) & (0.00) & (7.71) & (0.48) \\
     \cmidrule(lr){2-10} 
     & \multirow{2}{*}{FS}     & {62.49(63.25/61.76) } & {58.29(62.77/54.40) } & 56.71(88.03/41.83) & 40.67(72.24/28.30) & 28.74(60.59/18.84) & 13.86(84.10/7.55) & 43.04(77.73/29.75) & 5.31(2.87/35.74)        \\
     & & (0.00) & (0.00) & (0.00) & (0.00) & (0.00) & (0.00) & (0.00) & (0.00) \\
     \midrule
     \multirow{12}{*}{BIO Schema, with Label Modification}    
     & \multirow{2}{*}{MV}             & 61.73 (59.70/63.89) & 55.30 (51.02/59.73) & {81.71 (88.22/76.09)} & 72.37 (82.74/64.30) & 67.43 (79.01/58.80) & {{47.55 (72.91/35.29)}} & 59.78 (70.38/51.95) & 57.89 (52.68/64.24)\\
     &  & (0.00) & (0.00)  & (0.00) & (0.00) & (0.00) & (0.00) & (0.00) & (0.00) \\
     \cmidrule(lr){2-10} 
     & \multirow{2}{*}{WMV}            & 60.93 (58.46/63.62) & 54.54 (50.57/59.20) & 80.45 (86.04/75.54) & 76.50 (89.61/66.84) & 62.09 (70.57/55.43) & {47.07 (72.33/34.91)} & 60.22 (70.88/52.35) & 56.81 (51.85/62.84)\\ 
     &  & (0.00) & (0.00)  & (0.00) & (0.00) & (0.00) & (0.00) & (0.00) & (0.00) \\
     \cmidrule(lr){2-10} 
     & \multirow{2}{*}{DS}             & 47.29 (45.52/49.20) & 40.82 (37.50/44.80) & 81.54 (87.32/76.48) & 77.54 (91.15/67.47) & 72.25 (86.53/62.02) & 35.60 (52.71/26.88) & 55.86 (65.76/48.54) & 39.25 (36.55/42.39) \\
     & & (0.00) & (0.00)  & (0.00) & (0.00) & (0.00) & (0.00) & (0.00) & (0.00) \\
     \cmidrule(lr){2-10} 
     & \multirow{2}{*}{DP}             & {{63.62 (61.83/65.52)}} & 55.40 (51.86/59.46) & 83.08 (90.16/77.03) & 76.69 (89.56/67.05) & 62.15 (70.70/55.44) & 46.93 (72.15/34.78) & 60.14 (70.60/52.39) & 61.66 (56.55/67.85)\\
     & & (0.10) & (0.07) & (0.09) & (0.03) & (0.00) & (0.02) & (0.05) & (0.04) \\
     \cmidrule(lr){2-10} 
     & \multirow{2}{*}{MeTaL}          & 61.69 (59.57/63.95) & 54.63 (51.07/58.73) & 80.64 (86.44/75.65)& 76.37 (89.26/66.73) & 64.79 (77.84/55.43) & 46.69 (72.22/34.50) & 60.24 (70.90/52.37) & 58.75 (53.65/62.68) \\
     & & (0.08) & (0.04)  & (0.12) & (0.03) & (0.01) & (0.03) & (0.11) & (0.38) \\
     \cmidrule(lr){2-10} 
     & \multirow{2}{*}{FS}             & 61.97 (62.34/61.59) & {57.21 (61.73/53.33)} & 56.77 (91.29/41.19) & 42.66 (88.67/28.09) & 60.89 (73.01/52.20) & 13.29 (86.31/7.20) & 42.27 (77.22/29.09) & 8.01 (4.60/31.06) \\
     & & (0.00) & (0.00)  & (0.00) & (0.00) & (0.00) & (0.00) & (0.00) & (0.00) \\
    \midrule
      \multirow{12}{*}{IO Schema, without Label Modification}    
     & \multirow{2}{*}{MV}             & 8.10 (85.41/4.25) & 8.14 (88.89/4.27) & 0.04 (100.00/0.02) & 6.68 (80.49/3.48) & 29.71 (70.29/18.84) & 0.00 (0.00/0.00) & 8.93 (81.29/4.72) & 0.00 (0.00/0.00) \\
     &  & (0.00) & (0.00)  & (0.00) & (0.00) & (0.00) & (0.00) & (0.00) & (0.00) \\
     \cmidrule(lr){2-10} 
     & \multirow{2}{*}{WMV}            & 0.00 (0.00/0.00) & 0.00 (0.00/0.00) & 0.00 (0.00/0.00) & 0.00 (0.00/0.00) & 0.46 (30.00/0.46) & 0.00 (0.00/0.00) & 0.00 (0.00/0.00) & 0.00 (0.00/0.00) \\ 
     &  & (0.00) & (0.00)  & (0.00) & (0.00) & (0.00) & (0.00) & (0.00) & (0.00) \\
     \cmidrule(lr){2-10} 
     & \multirow{2}{*}{DS}             & 49.24 (49.84/48.65) & 41.38 (42.86/40.00) & 71.84 (89.25/60.12) & 56.69 (83.23/42.98) & 29.91 (77.56/18.53) & 41.26 (77.45/28.12) & 51.10 (73.27/39.24) & 40.83 (42.68/39.14)\\
     & & (0.00) & (0.00)  & (0.00) & (0.00) & (0.00) & (0.00) & (0.00) & (0.00) \\
     \cmidrule(lr){2-10} 
     & \multirow{2}{*}{DP}             & 7.74 (84.50/4.05) & 8.65 (94.44/4.53) & 0.04 (100.00/0.02) & 6.73 (100.00/3.48) & 30.45 (79.35/18.84) & 0.00 (0.00/0.00) & 8.73 (83.10/4.61) & 0.00 (0.00/0.00)\\
     & & (0.00) & (0.00) & (0.00) & (0.00) & (0.00) & (0.00) & (0.00) & (0.00) \\
     \cmidrule(lr){2-10} 
     & \multirow{2}{*}{MeTaL}          & 6.59 (88.72/3.42) & 7.57 (92.50/3.95) & 0.01 (20.00/0.00) & 6.73 (100.00/3.48) & 30.20 (78.71/18.68) & 0.00 (0.00/0.00) & 8.17 (82.22/4.30) & 0.00 (0.00/0.00)\\
     & & (0.08) & (0.04)  & (0.00) & (0.00) & (0.00) & (0.00) & (0.00) & (0.00) \\
     \cmidrule(lr){2-10} 
     & \multirow{2}{*}{FS}             & 50.77 (81.11/36.95) & 45.44 (83.57/31.20) & 19.27 (88.91/10.80) & 42.87 (93.93/27.77) & 29.81 (78.95/18.38) & 0.00 (0.00/0.00) & 26.36 (83.18/15.66) & 27.28 (70.04/16.94) \\
     & & (0.00) & (0.00)  & (0.00) & (0.00) & (0.00) & (0.00) & (0.00) & (0.00) \\  
    \bottomrule
    \end{tabular}
    }
    \label{tab:seq_adapt}
\end{table*}

\section{Implementation Details}

\subsection{Hardware and Implementation}
Our models are implemented based on Python and PyTorch.
For gradient-based optimization, we adopt AdamW Optimizer and linear learning rate scheduler; and we early stop the training process based on the evaluation metric values on validation set.
For all the compared methods, we either re-implement them based on official released code or create an interface for calling their official implementations. 
For fine-tuning pre-trained language models, we use the dumps provided by HuggingFace\footnote{https://huggingface.co/models}.

We use a pre-trained BERT model\footnote{https://huggingface.co/bert-base-cased} to extract features for textual classification datasets. 
For text classification dataset, we use the outputting embedding of the [CLS] token as data feature; 
for relation classification, we follow the R-BERT~\cite{wu2019enriching} to use the concatenation of embeddings of [CLS] and the two entity tokens as data feature. 
Other features, \eg, TF-IDF feature, or other pre-trained language models are also supported in \benchmark.

All experiments are run on CPUs or 64 Nvidia V100 GPUs (32GB VRAM) on Microsoft Azure.

\subsection{Hyper-parameter Search Space}
For each model, we use grid search to find the best hyer-parameters on validation set.
For each trial, we repeat 3 runs with different initializations and for final evaluation, we repeat 5 runs with different initializations.
The search space is based on the suggestions in original paper and can be found in Table~\ref{tab:search}.

\begin{table*}[h]
    \centering
    \caption{The hyper-parameters and search space. Note that the ConNet shares search space of other parameters with its backbone, \ie, LSTM-CRF/BERT-CRF.}
    \ra{1.2}
    \scalebox{0.67}{
    \begin{tabular}{ l l  l  l }
        \toprule
        \textbf{Model} &
         \textbf{Hyper-parameter} &\textbf{Description} & \textbf{Range} \\
         \midrule
    \multirow{3}{*}{MeTal} &  \texttt{lr} & learning rate & 1e-5,1e-4,1e-3,1e-2,1e-1\\
    & \texttt{weight\_decay} & weight decay & 1e-5,1e-4,1e-3,1e-2,1e-1\\
    & \texttt{num\_epoch} & the number of training epochs & 5,10,50,100,200\\ 
    \midrule[0.05pt] \midrule[0.05pt]
    
    \multirow{3}{*}{DP} &  \texttt{lr} & learning rate & 1e-5,5e-5,1e-4\\
    & \texttt{weight\_decay} & weight decay & 1e-5,1e-4,1e-3,1e-2,1e-1\\
    & \texttt{num\_epoch} & the number of training epochs & 5,10,50,100,200\\ 
    \midrule[0.05pt] \midrule[0.05pt]
         
    \multirow{3}{*}{LogReg} & \texttt{batch\_size}    &  {the input batch\_size} & 32,128,512\\
    & \texttt{lr} & learning rate & 1e-5,1e-4,1e-3,1e-2,1e-1\\
    & \texttt{weight\_decay} & weight decay & 1e-5,1e-4,1e-3,1e-2,1e-1\\ 
    \midrule[0.05pt] \midrule[0.05pt]
    
    \multirow{5}{*}{MLP} & \texttt{batch\_size}    &  {the input batch\_size} & 32,128,512\\
    & \texttt{lr} & learning rate & 1e-5,1e-4,1e-3,1e-2,1e-1\\
    & \texttt{weight\_decay} & weight decay & 1e-5,1e-4,1e-3,1e-2,1e-1\\ 
    &\texttt{ffn\_num\_layer}    &  {the number of MLP layers} & 2 \\
    &\texttt{ffn\_hidden\_size}    &  {the hidden size of MLP layers} & 100\\
    \midrule[0.05pt] \midrule[0.05pt]
    
    \multirow{2}{*}{BERT} & \texttt{batch\_size}    &  {the input batch\_size} & 16,32\\
    & \texttt{lr} & learning rate & 2e-5,3e-5,5e-5\\
    \midrule[0.05pt] \midrule[0.05pt]

    \multirow{8}{*}{COSINE} & \texttt{batch\_size}    &  {the input batch\_size} & 32\\
    & \texttt{lr} & learning rate & 1e-6,1e-5\\
    & \texttt{weight\_decay} &  weight decay & 1e-4\\ 
    & $T$ & the period of updating model & 50,100,200\\ 
    & $\xi$ & the confident threshold & 0.2,0.4,0.6,0.8\\ 
    & $\lambda$ & the weight for confident regularization & 0.01,0.05,0.1\\ 
    & $\mu$ & the weight for contrastive regularization & 1\\ 
    & $\gamma$ & the margin for contrastive regularization & 1\\ 
    \midrule[0.05pt] \midrule[0.05pt]
    
    \multirow{9}{*}{Denoise} & \texttt{batch\_size}    &  {the input batch\_size} & 32,128,512\\
    & \texttt{lr} & learning rate & 1e-4,1e-3,1e-2\\
    & \texttt{weight\_decay} & weight decay & 0.0\\ 
    & \texttt{alpha} & momentum term for temporal ensembling & 0.6\\ 
    & \texttt{c1} & coefficient of denoiser loss & 0.1,0.3,0.5,0.7,0.9\\ 
    & \texttt{c2} & coefficient of classifier loss & 0.1,0.3,0.5,0.7,0.9\\ 
    & \texttt{c3} & coefficient of unsupervised self-training loss & 1-\texttt{c2}-\texttt{c1}\\ 
    &\texttt{ffn\_num\_layer}    &  {the number of MLP layers} & 2 \\
    &\texttt{ffn\_hidden\_size}    &  {the hidden size of MLP layers} & 100\\
    \midrule[0.05pt] \midrule[0.05pt]
    
    \multirow{13}{*}{LSTM-CRF} & \texttt{batch\_size}    &  {the input batch\_size} & 16,32,64\\
    & \texttt{lr} & learning rate & 1e-2,5e-3,1e-3\\
    & \texttt{weight\_decay} & weight decay & 1e-8\\
    & \texttt{dropout} & dropout ratio & 0.0,0.5\\ 
    &\texttt{word\_feature\_extractor}    &  {the word feature extractor layers} & LSTM,GRU\\
    &\texttt{word\_embed\_dimension}    &  {the embedding dimension of word} & 100\\
    &\texttt{LSTM/GRU\_hidden\_size}    &  {the hidden size of LSTM/GRU layers} & 200\\
    &\texttt{num\_hidden\_layer}    &  {the number of LSTM/GRU layers} & 1\\
    &\texttt{LSTM/GRU\_hidden\_size}    &  {the hidden size of LSTM/GRU layers} & 200\\
    &\texttt{num\_hidden\_layer}    &  {the number of LSTM/GRU layers} & 1\\
    &\texttt{char\_feature\_extractor}    &  {the character feature extractor layers} & CNN\\
    &\texttt{char\_embed\_dimension}    &  {the embedding dimension of character} & 30\\
    \midrule[0.05pt] \midrule[0.05pt]    
    
    \multirow{5}{*}{BERT-CRF} & \texttt{batch\_size}    &  {the input batch\_size} & 16,32,8\\
    & \texttt{lr} & learning rate & 2e-5,3e-5,5e-5\\
    & \texttt{lr\_crf} & learning rate for the CRF layer & 1e-3,5e-3,1e-2\\
     & \texttt{weight\_decay} & weight decay & 1e-6\\
      & \texttt{weight\_decay\_crf} & weight decayfor the CRF layer & 1e-8\\
    \midrule[0.05pt] \midrule[0.05pt]
    
    \multirow{2}{*}{HMM} & $\gamma$    &  {redundency factor} & 0,0.1,0.3,0.5,0.7,0.9\\
    & \texttt{num\_epoch} & the number of training epochs & 50\\ 
    \midrule[0.05pt] \midrule[0.05pt]
    
    \multirow{5}{*}{CHMM} & \texttt{batch\_size}    &  {the input batch\_size} & 16,64,128\\
    & \texttt{nn\_lr} & learning rate of NN & 1e-3,5e-4,1e-4\\
    & \texttt{hmm\_lr} & learning rate of HMM & 1e-2,5e-3,1e-3\\
    & \texttt{num\_pretrain\_epoch} & the number of pre-training epochs & 2,5\\
    & \texttt{num\_epoch} & the number of training epochs & 50\\ \midrule[0.05pt] \midrule[0.05pt]
    
    \multirow{1}{*}{ConNet} & \texttt{n\_steps\_phase1}    &  {the number of training steps of phase1} & 200,500,1000\\
    \bottomrule
    \end{tabular}
    }
    \label{tab:search}
\end{table*}

\subsection{Parameters for studies in Sec.~\ref{sec:generator}}
\label{sec:para_study}

\textbf{Fig.~\ref{fig:syn} (a)} We generate 10 labeling functions; 5 for positive label and 5 for negative label. The mean accuracy, mean propensity and radius of propensity is set to 0.75, 0.1, 0.0, respectively.

\textbf{Fig.~\ref{fig:syn} (b)} We generate 10 labeling functions; 5 for positive label and 5 for negative label. The mean accuracy, radius of accuracy and radius of propensity is set to 0.75, 0.1, 0.0, respectively.

\textbf{Fig.~\ref{fig:semi}} The minimum propensity of candidate LFs is 0.1. 
The minimum accuracy is set to be the label prior plus 0.1, \eg, for LFs labeling positive label, the minimum accuracy is $P(y=1)+0.1$. 
For ($n$, $m$)-gram features, $n$ is set to 1 and $m$ is 2.
\section{Additional Results}
\label{sec:results}

\subsection{Classification}
The detailed comparisons over the collected classification datasets are in Table~\ref{tab:classification}-\ref{tab:classification_continue}.

\begin{table*}[h]
    \centering
    \caption{\textbf{Classification}: detailed comparison. Each metric value is averaged over 5 runs. \underline{Underline} indicates using soft label for training end model.  {\textcolor{red}{red}} and {\color{blue}{blue}} indicate the best and second best result for each end model respectively, and \colorbox{lightgray!60}{gray} is the best weak supervision method in this table.}
    \scalebox{0.5}{
    \begin{tabular}{ c c c c c c c c c c c c c c c c c}
        \toprule
          \textbf{End} & \textbf{Label} & \textbf{IMDb} & \textbf{Yelp} & \textbf{Youtube}& \textbf{SMS} & \textbf{AGNews} & \textbf{TREC} & \textbf{Spouse} & \textbf{CDR} & \textbf{SemEval} & \textbf{ChemProt}  & \textbf{Commercial} & \textbf{Tennis Rally} & \textbf{Basketball}  & \textbf{Census} & \multirow{2}{*}{\textbf{\underline{Average}} }  \\
          \textbf{Model ($\downarrow$)} & \textbf{Model ($\downarrow$)} & (Acc.)  & (Acc.)  & (Acc.) & (F1) & (Acc.) & (Acc.)  & (F1) &(F1)& (Acc.) & (Acc.) & (F1) & (F1) & (F1) & (F1)\\
         \midrule
    \multirow{12}{*}{--}     &  \multirow{2}{*}{MV}  &\color{red}{71.04}  &\color{red}{70.21}  &\color{red}{84.00}  &\color{red}{23.97}  &63.84  &\color{blue}{60.80}  &20.81 &60.31  &\color{red}{77.33} &49.04  &85.28  &81.00  &16.33 &   32.80 & \blue{56.91} \\
     &  &(0.00) & (0.00) & (0.00) & (0.00) & (0.00) & (0.00) &(0.00) & (0.00) & (0.00) & (0.00) & (0.00) & (0.00) & (0.00) & (0.00)\\\cmidrule(lr){2-17}
     
     & \multirow{2}{*}{WMV}  &\color{red}{71.04}  &68.50  &78.00  &\color{red}{23.97} &\color{red}{64.00}  &57.20 &20.53 &52.12  &\color{blue}{71.00}  &\color{red}{52.08}  &83.80  &\color{red}{82.61}  &13.13 &9.99  &53.43 \\
     &  & (0.00) & (0.00) & (0.00) & (0.00) & (0.00) & (0.00)& (0.00)& (0.00) & (0.00) & (0.00) & (0.00) & (0.00) & (0.00)& (0.00)\\\cmidrule(lr){2-17} 
     
     & \multirow{2}{*}{DS}  & 70.60 & \color{blue}{71.45} & \color{blue}{83.20} & 4.94 & 62.76 & 50.00 & 15.53 & 50.43 & \color{blue}{71.00} & 37.59 & \color{red}{88.24} & 80.65 & 13.79 & \color{red}{47.16}  &53.38 \\
     &  & (0.00) & (0.00) & (0.00) & (0.00) & (0.00) & (0.00)& (0.00)& (0.00) & (0.00) & (0.00) & (0.00) & (0.00) & (0.00)& (0.00)\\\cmidrule(lr){2-17} 
     
     & \multirow{2}{*}{DP}   &\color{blue}{70.96}  &69.37  &82.00  &\color{blue}{23.78}  &\color{blue}{63.90}  &\color{red}{64.20} &21.12  &\color{blue}{63.51}   &\color{blue}{71.00}  &47.42  &77.29  &\color{blue}{82.55}  &\color{red}{17.39} &  22.66&55.51 \\
     &  & (0.00) & (0.03) & (2.02) & (0.89) & (0.08) &(0.51)& (0.08) & (0.07)  & (0.00) & (0.29) & (0.00) & (0.00) & (0.00)&  (0.02)\\\cmidrule(lr){2-17} 
     & \multirow{2}{*}{MeTaL}  &70.96  &68.30  &\color{red}{84.00}  &7.06  &62.27  &57.60  &\colorbox{lightgray!60}{\red{46.62}}  &\colorbox{lightgray!60}{\red{69.61}}   &\color{blue}{71.00}  &\color{blue}{51.96}  &\color{blue}{88.20}  &82.52  &13.13  &\color{blue}{44.48}& \red{58.40} \\
     &  & (0.59) &(0.43) & (0.00) & (0.00) & (0.27) & (0.00) &(0.00) & (0.01)  & (0.00) & (0.00) &(0.00) & (0.04) &(0.00)&(2.34) \\\cmidrule(lr){2-17} 
     & \multirow{2}{*}{FS}  &70.36  &68.68  &76.80  &0.00  &60.98  &31.40  &\color{blue}{34.30}  &20.18  &31.83  &43.31  &77.31  &82.29  &\color{blue}{17.25} & 15.33 &45.00  \\
     & & (0.00) & (0.00) & (0.00) & (0.00) & (0.00) & (0.00) &(0.00) & (0.00) & (0.00) & (0.00) &(0.00) & (0.00) & (0.00)& (0.00) \\\midrule[0.05pt] \midrule[0.05pt]
         
     \multirow{26}{*}{LR}    &  \multirow{2}{*}{Gold}  & 81.56 &89.16 &94.24 &93.79 &86.51 &68.56 & -- &63.09 &93.23 &77.96 &91.01 &82.73 &62.82 &67.12 &80.91  \\
     & &(0.20) &(0.27) &(0.41) &(0.61) &(0.28) &(1.15) & -- &(0.36) &(0.31) &(0.25) &(0.12) &(0.65) &(1.57) &(0.52)   \\\cmidrule(lr){2-17} 
    &  \multirow{2}{*}{MV} & 76.93 &86.21 &90.72 &\color{red}{90.77} &82.69 &57.56 &23.99 &54.44 &82.83 &\color{blue}{55.84} &\color{red}{90.62} &83.59 &26.31 &47.96&67.89   \\
     & &(0.45) &(0.27) &(1.42) &(1.02) &(0.05) &(4.99) &(0.98) &(0.54) &(1.91) &(0.65) &(0.08) &(0.07) &(4.60) &(4.23)  \\\cmidrule(lr){2-17}
     &  \multirow{2}{*}{\underline{MV}} & \color{red}{77.26} &86.33 &\colorbox{lightgray!60}{\red{93.36}} &90.07 &82.69 &62.68 &22.45 &\color{blue}{56.69} &\colorbox{lightgray!60}{\red{85.73}} &\colorbox{lightgray!60}{\red{56.73}} &90.25 &82.15 &\color{blue}{30.67} &51.56& \red{69.19}  \\
     & &(0.14) &(0.19) &(0.93) &(2.62) &(0.14) &(4.56) &(2.79) &(0.65) &(1.08) &(0.33) &(0.28) &(0.23) &(8.52) &(2.59)  \\\cmidrule(lr){2-17} 
     & \multirow{2}{*}{WMV} & 76.63 &85.23 &88.80 &90.25 &82.88 &52.88 &20.24 &53.62 &72.70 &54.91 &89.94 &83.57 &23.48 &23.94 &64.22 \\
     & &(0.22) &(0.21) &(0.25) &(0.48) &(0.26) &(4.50) &(2.81) &(0.98) &(4.31) &(0.36) &(0.15) &(0.00) &(16.23) &(14.25)  \\\cmidrule(lr){2-17} 
     & \multirow{2}{*}{\underline{WMV}} & 77.03 &86.11 &\color{blue}{92.64} &90.08 &82.84 &\color{blue}{63.84} &23.23 &55.58 &83.87 &56.63 &\color{blue}{90.34} &82.38 &26.65 &30.12&67.24 \\
     & &(0.38) &(0.20) &(0.41) &(1.20) &(0.05) &(7.60) &(2.08) &(1.70) &(2.13) &(0.49) &(0.17) &(0.22) &(8.40) &(12.97) \\\cmidrule(lr){2-17}
     
     & \multirow{2}{*}{DS}   & 76.54 & 85.43 & 88.32 & 90.32 & 82.95 & 47.16 & 19.01 & 51.84 & 72.80 & 49.25 & 89.77 & 83.57 & 24.37 & 49.70&65.07 \\
     &  & (0.30) & (0.20) & (0.82) & (1.66) & (0.07) & (1.30) & (2.33) & (0.50) & (2.20) & (1.51) & (0.18) & (0.00) & (11.77) & (0.24) \\\cmidrule(lr){2-17}
     & \multirow{2}{*}{\underline{DS}}  & 77.15 & 85.91 & 88.88 & 89.88 & 82.92 & 50.00 & 17.07 & 49.88 & 72.97 & 48.31 & 89.88 & 83.59 & 20.45 & 50.10&64.79 \\
     &  & (0.36) & (0.14) & (0.93) & (0.87) & (0.21) & (2.54) & (2.22) & (1.44) & (1.10) & (2.19) & (0.22) & (0.04) & (11.09) & (0.39) \\\cmidrule(lr){2-17}

      & \multirow{2}{*}{DP} & 76.90 &85.38 &90.00 &25.42 &82.04 &\color{red}{64.08} &24.75 &56.24 &72.80 &52.94 &87.44 &83.57 &24.69 &15.71  &60.14\\
     & & (0.61) &(0.33) &(0.80) &(0.65) &(0.14) &(4.41) &(1.11) &(0.57) &(3.23) &(0.65) &(0.17) &(0.00) &(1.70) &(15.36)  \\\cmidrule(lr){2-17}
     & \multirow{2}{*}{\underline{DP}} & 76.86 &84.98 &89.92 &43.33 &\color{blue}{83.21} &52.96 &22.80 &55.90 &\color{blue}{84.00} &55.15 &87.51 &\color{red}{83.68} &24.94 &21.02 &61.88 \\
     & & (0.27) &(0.36) &(0.93) &(7.04) &(0.27) &(3.11) &(3.68) &(0.74) &(2.35) &(0.54) &(0.18) &(0.00) &(2.70) &(13.55)   \\\cmidrule(lr){2-17} 
     & \multirow{2}{*}{MeTaL} &76.30 &86.32 &89.84 &89.13 &83.16 &59.52 &21.77 &56.52 &75.90 &54.60 &90.00 &\color{red}{83.68} &4.66 &\color{blue}{57.39} &66.34 \\
     & & (0.28) &(0.22) &(0.78) &(0.88) &(0.05) &(1.82) &(0.76) &(0.57) &(3.99) &(0.41) &(0.06) &(0.00) &(4.96) &(0.78)  \\\cmidrule(lr){2-17} 
     & \multirow{2}{*}{\underline{MeTaL}} &\color{blue}{77.18} &86.41 &88.00 &\color{blue}{90.76} &\color{red}{83.36} &54.64 &22.17 &\color{red}{57.80} &79.73 &55.68 &90.15 &83.57 &25.62 &\colorbox{lightgray!60}{\red{58.16}} & \blue{68.09} \\
     & & (0.20) &(0.22) &(2.01) &(0.83) &(0.30) &(3.98) &(1.43) &(0.42) &(2.67) &(0.59) &(0.12) &(0.00) &(17.00) &(0.72)  \\\cmidrule(lr){2-17}
    & \multirow{2}{*}{FS} & 76.74 &\colorbox{lightgray!60}{\red{86.63}} &87.68 &66.04 &82.43 &34.24 &\color{blue}{28.69} &48.68 &31.83 &47.26 &87.18 &\color{blue}{83.64} &\color{red}{31.13} &26.53 &58.48   \\
     & & (0.79) &(0.17) &(0.78) &(5.54) &(0.21) &(1.99) &(1.96) &(0.60) &(0.00) &(0.22) &(0.19) &(0.05) &(2.25) &(15.68) \\ \cmidrule(lr){2-17}
     & \multirow{2}{*}{\underline{FS}} & 76.84 &\color{blue}{86.48} &88.72 &63.75 &82.86 &35.56 &\color{red}{31.69} &55.53 &40.13 &48.21 &89.21 &\color{red}{83.68} &25.41 &21.37&59.25   \\
     & & (0.34) &(0.29) &(0.53) &(5.16) &(0.19) &(4.93) &(2.14) &(0.77) &(3.48) &(1.35) &(0.34) &(0.00) &(7.07) &(15.09)    \\\midrule[0.05pt] \midrule[0.05pt]

    \multirow{26}{*}{MLP}    &  \multirow{2}{*}{Gold}  &81.79 &89.19 &94.00 &94.45 &87.69 &66.04 & -- &63.02 &93.33 &80.15 &91.69 &81.48 &64.97 &67.13 &81.15  \\
     &  &(0.32) &(0.31) &(0.44) &(0.59) &(0.18) &(4.05) & -- &(0.48) &(0.24) &(0.55) &(0.07) &(0.50) &(13.65) & (0.16) \\\cmidrule(lr){2-17} 
      &  \multirow{2}{*}{MV}  &77.14 &84.24 &89.44 &89.03 &83.37 &61.40 &21.52 &56.42 &83.13 &\color{blue}{56.04} &\color{blue}{90.42} &81.85 &39.40  &54.62 & \blue{69.14}\\
     &  &(0.13) &(1.19) &(0.74) &(0.82) &(0.27) &(3.10) &(0.99) &(0.86) &(1.50) &(0.59) &(0.27) &(0.16) &(4.82) &(3.78) \\\cmidrule(lr){2-17}
     &  \multirow{2}{*}{\underline{MV}}  &77.10 &84.91 &\color{blue}{90.16} &\colorbox{lightgray!60}{\red{91.91}} &83.41 &\color{blue}{63.88} &22.59 &\color{blue}{57.66} &\color{red}{85.53} &55.83 &\color{red}{90.55} &82.23 &39.84 &56.73 &70.17  \\
     & &(0.37) &(1.28) &(0.60) &(0.73) &(0.20) &(4.49) &(0.66) &(1.09) &(1.07) &(0.63) &(0.27) &(0.14) &(21.02) &(3.77)  \\\cmidrule(lr){2-17} 
     & \multirow{2}{*}{WMV}  &76.66 &79.17 &88.16 &90.73 &83.62 &59.76 &18.71 &53.77 &72.37 &54.64 &88.59 &83.56 &38.75 &39.04  &66.25\\
     &  &(0.40) &(5.31) &(0.86) &(1.00) &(0.16) &(2.14) &(2.10) &(1.17) &(0.74) &(0.58) &(0.58) &(0.03) &(17.47) & (4.10) \\\cmidrule(lr){2-17}
     & \multirow{2}{*}{\underline{WMV}}  &76.90 &85.45 &\color{red}{92.48} &\color{blue}{91.20} &83.54 &63.48 &19.70 &57.21 &\color{blue}{83.77} &\color{red}{56.52} &90.07 &82.64 &\color{blue}{40.73} &50.86 & \red{69.61} \\
     &  &(0.24) &(1.21) &(0.16) &(1.46) &(0.18) &(5.37) &(0.88) &(0.52) &(2.93) &(0.79) &(0.38) &(0.14) &(11.98) & (8.55) \\\cmidrule(lr){2-17}
     
     & \multirow{2}{*}{DS}   & 76.64 & \color{blue}{86.00} & 88.00 & 88.63 & 83.45 & 47.28 & 17.13 & 51.96 & 73.60 & 48.10 & 88.73& 83.59& 22.79 & 51.19&64.79 \\
     &  & (0.37) & (0.29) & (0.98) & (0.48) & (0.22) & (1.65) & (0.35) & (0.41) & (1.01) & (0.64) & (0.60)& (0.04)& (12.00)& (1.30) \\\cmidrule(lr){2-17}
     
     & \multirow{2}{*}{\underline{DS}}& 77.18 & \color{red}{86.06} & 87.44 & 88.82 & \colorbox{lightgray!60}{\red{83.86}} & 49.92 & 16.42 & 51.14 & 72.93 & 44.64 & 89.86& 83.59 & 34.81& 50.06&65.48 \\
     &  & (0.38) & (0.23) & (0.82) & (0.61) & (0.23) & (1.03) & (0.59) & (0.45) & (2.46) & (0.43) & (0.16)& (0.04)& (19.01)& (0.53) \\\cmidrule(lr){2-17}
     
      & \multirow{2}{*}{DP}  &76.42 &83.98 &90.00 &26.16 &83.05 &\colorbox{lightgray!60}{\red{68.40}} &21.65 &56.69 &72.83 &52.88 &88.40 &83.57 &37.50 &47.54&63.51 \\
     &  &(0.51) &(1.88) &(0.25) &(3.35) &(0.22) &(1.41) &(0.49) &(1.31) &(2.26) &(1.59) &(0.37) &(0.00) &(3.76)  &(6.59) \\\cmidrule(lr){2-17}
     & \multirow{2}{*}{\underline{DP}}  &76.77 &80.91 &90.08 &27.15 &83.71 &55.52 &23.77 &45.32 &79.23 &55.52 &88.68 &83.66 &40.70 &54.57 &63.26  \\
     &  &(0.38) &(1.56) &(1.11) &(2.98) &(0.17) &(2.83) &(0.94) &(22.67) &(0.31) &(0.77) &(0.24) &(0.04) &(7.20) &(4.21) \\\cmidrule(lr){2-17} 
     & \multirow{2}{*}{MeTaL}  &76.35 & 85.61 &88.88 &88.07 &\color{blue}{83.78} &56.32 &20.84 &56.58 &73.00 &55.02 &89.73 &\color{blue}{83.70} &36.74  &\color{blue}{57.66} &68.02 \\
     &  &(0.37) &(0.54) &(1.30) &(0.29) &(0.19) &(4.41) &(0.64) &(0.46) &(1.04) &(0.75) &(0.11) &(0.04) &(18.93) &(0.32) \\\cmidrule(lr){2-17} 
     & \multirow{2}{*}{\underline{MeTaL}}  &\colorbox{lightgray!60}{\red{77.61}} &85.19 &87.44 &91.10 &83.77 &63.80 &21.17 &\color{red}{58.17} &74.27 &55.52 &89.86 &83.56 &36.35  &\color{red}{57.84}&68.98 \\
     &  &(0.36) &(0.16) &(0.90) &(0.97) &(0.21) &(1.25) &(0.49) &(0.21) &(2.87) &(0.82) &(0.08) &(0.03) &(14.00) &(0.83)  \\\cmidrule(lr){2-17}
     & \multirow{2}{*}{FS}  &76.78 &84.50 &86.32 &71.81 &83.43 &28.48 &\color{red}{30.55} &49.20 &31.83 &46.46 &88.20 &83.57 &38.53  & 21.93&58.69 \\
     & &(15.99)  &(1.33) &(1.35) &(4.99) &(0.22) &(1.00) &(2.06) &(0.40) &(0.00) &(0.46) &(0.37) &(0.12) &(9.83)  &(0.21)\\ \cmidrule(lr){2-17} 
     & \multirow{2}{*}{\underline{FS}}  &\color{blue}{77.35} &83.95 &85.20 &37.54 &82.65 &25.60 &\color{blue}{30.37} &49.33 &32.50 &48.23 &89.59 &\colorbox{lightgray!60}{\red{83.77}} &\colorbox{lightgray!60}{\red{43.18}} & 39.03&57.74 \\
     &  &(0.42) &(0.81) &(0.91) &(16.98) &(0.22) &(3.72) &(2.72) &(1.30) &(1.17) &(1.22) &(0.09) &(0.09) &(7.79) & (1.76) \\\midrule[0.05pt]\midrule[0.05pt]
     
     \multicolumn{2}{c}{\multirow{2}{*}{Denoise}}  & 76.22 &71.56 &76.56 &91.69 &83.45 &56.20 &22.47 &56.54 &80.83 &53.96 &\colorbox{lightgray!60}{\red{91.34}} &82.34 &33.73 &43.71   &65.76 \\
     &  & (0.37) &(15.80) &(19.24) &(1.42) &(0.11) &(6.73) &(7.50) &(0.37) &(1.31) &(0.38) &(0.16) &(2.46) &(3.43) &(3.51) \\
    \bottomrule
    \end{tabular}
    }

    \label{tab:classification}
\end{table*}

\begin{table*}[h]
    \centering
    \caption{Comparisons on textual datasets among pre-trained language model-based methods.}
    \scalebox{0.46}{
    \begin{tabular}{ c c c c c c c c c c c c c}
        \toprule
          \textbf{End Model ($\downarrow$)} & \textbf{Label Model ($\downarrow$)} & \textbf{IMDb} & \textbf{Yelp} & \textbf{Youtube}& \textbf{SMS} & \textbf{AGNews} & \textbf{TREC} & \textbf{Spouse} & \textbf{CDR} & \textbf{SemEval} & \textbf{ChemProt} & \multirow{2}{*}{\textbf{\underline{Average}}}   \\
           &  & (Acc.)  & (Acc.)  & (Acc.) & (F1) & (Acc.) & (Acc.)  & (F1) &(F1)& (Acc.) & (Acc.) \\
         \midrule

     \multirow{26}{*}{B}    &  \multirow{2}{*}{Gold}  &91.58 &95.48 &97.52 &96.96 &90.78 &96.24 & -- &65.39 &95.43 &89.76&91.02 \\
     & &(0.31) &(0.53) &(0.64) &(0.66) &(0.49) &(0.61) & -- &(1.18) &(0.65) &(0.88)  \\\cmidrule(lr){2-13} 
    &  \multirow{2}{*}{MV} &79.73 &82.26 &\color{red}{95.36} &94.56 &86.27 &66.56 &19.56 &57.16 &83.93 &56.09& \blue{72.15} \\
     & &(2.60) &(3.50) &(1.71) &(1.88) &(0.53) &(2.31) &(1.22) &(0.83) &(1.74) &(1.08)  \\\cmidrule(lr){2-13}
     &  \multirow{2}{*}{\underline{MV}} &79.91 &85.64 &93.68 &\color{blue}{94.85} &86.62 &66.56 &19.43 &\color{red}{58.89} &\color{red}{85.03} &\color{red}{57.32} & \red{72.79} \\
     & &(2.23) &(2.52) &(0.47) &(1.16) &(0.28) &(1.20) &(0.95) &(0.50) &(0.83) &(0.98)   \\\cmidrule(lr){2-13}
      & \multirow{2}{*}{WMV} & \color{blue}{81.32} &81.40 &89.92 &91.79 &85.49 &54.64 &19.74 &53.60 &70.97 &55.40 &68.43  \\
     &  &(1.35) &(5.06) &(1.51) &(2.67) &(0.63) &(4.85) &(5.48) &(3.25) &(0.24) &(1.02)  \\\cmidrule(lr){2-13}
     & \multirow{2}{*}{\underline{WMV}}  &80.70 &81.19 &93.76 &\color{red}{95.02} &86.66 &66.00 &19.34 &57.53 &82.47 &55.66 &71.83    \\
     & &(1.39) &(3.74) &(2.10) &(1.26) &(0.44) &(2.33) &(2.87) &(0.46) &(1.29) &(1.36)  \\\cmidrule(lr){2-13}
     
       & \multirow{2}{*}{DS} & 80.25 & \color{blue}{88.59} & 92.88 & 91.98 & 86.69 & 46.36 & 16.42 & 50.01 & 71.67 & 44.37&66.92   \\
     &   & (2.23) & (1.25) & (0.78) & (1.00) & (0.35) & (3.39) & (0.60) & (0.30) & (0.66) & (0.53)   \\\cmidrule(lr){2-13}
     & \multirow{2}{*}{\underline{DS}}  & 78.79 & 88.57 & 89.36 & 93.06 & 86.59 & 48.40 & 16.23 & 50.49 & 71.70 & 45.71 &66.89    \\
     & & (1.59) & (2.01) & (2.56) & (1.30) & (0.38) & (0.95) & (0.04) & (0.48) & (0.81) & (1.46)  \\\cmidrule(lr){2-13}

     & \multirow{2}{*}{DP}  &80.35 &81.17 &\color{blue}{93.84} &29.97 &85.36 &\color{red}{68.64} &18.66 &\color{blue}{58.48} &71.07 &54.00 &64.15 \\
     &  &(2.16) &(4.36) &(1.61) &(2.33) &(0.92) &(3.57) &(1.55) &(0.73) &(0.33) &(1.41)  \\\cmidrule(lr){2-13}
     & \multirow{2}{*}{\underline{DP}}  &80.82 &82.90 &93.60 &31.96 &86.55 &\color{blue}{68.40} &\color{blue}{28.74} &57.94 &\color{blue}{83.93} &\color{blue}{57.00} &67.18 \\
     & &(1.29) &(3.69) &(0.98) &(2.87) &(0.08) &(2.41) &(7.63) &(0.29) &(0.83) &(1.20)   \\\cmidrule(lr){2-13}
      & \multirow{2}{*}{MeTaL}  &80.02 &86.92 &92.32 &92.28 &\color{blue}{86.77} &58.28 &17.26 &58.48 &71.47 &55.48 &69.93 \\
     & &(2.46) &(3.52) &(1.44) &(2.01) &(0.29) &(1.95) &(0.73) &(0.90) &(0.57) &(1.33)  \\\cmidrule(lr){2-13} 
     & \multirow{2}{*}{\underline{MeTaL}} &81.23 &88.29 &92.48 &90.43 &\color{red}{86.82} &62.44 &17.18 &56.72 &70.80 &56.17 &70.26 \\
     &  &(1.23) &(1.57) &(0.99) &(2.64) &(0.23) &(2.96) &(0.23) &(3.26) &(0.87) &(0.66) \\\cmidrule(lr){2-13}
     & \multirow{2}{*}{FS}  &\color{red}{82.26} &87.76 &91.84 &11.62 &86.29 &27.60 &\color{red}{33.63} &4.29 &31.83 &45.66 &50.28  \\
     &  &(1.41) &(1.30) &(2.10) &(11.39) &(0.49) &(0.00) &(18.57) &(8.59) &(0.00) &(0.45)     \\ \cmidrule(lr){2-13}
     & \multirow{2}{*}{\underline{FS}}  &81.20 &\color{red}{88.86} &91.60 &7.32 &85.51 &30.96 &9.14 &35.25 &31.83 &49.53 &51.12 \\
     & &(1.01) &(0.92) &(2.18) &(5.35) &(0.62) &(4.04) &(18.29) &(5.75) &(0.00) &(1.14)    \\\midrule[0.05pt] \midrule[0.05pt]
     
    \multirow{24}{*}{BC}    &  \multirow{2}{*}{MV}  &82.98 &89.22 &\colorbox{lightgray!60}{\red{98.00}} &\color{red}{97.01} &87.03 &\color{blue}{76.56} &32.39 &58.99 &\color{blue}{86.80} &\color{blue}{58.47} & \red{76.75}\\
    & & (0.05) &(0.05) &(0.00) &(0.00) &(0.00) &(0.08) &(3.41) &(0.09) &(0.46) &(0.08)\\\cmidrule(lr){2-13} 
     &  \multirow{2}{*}{\underline{MV}} & 83.14 &89.64 &95.44 &\color{blue}{96.85} &87.14 &68.56 &42.71 &59.26 &86.13 &58.01& \blue{76.69} \\
     & & (0.42) &(0.03) &(0.20) &(0.31) &(0.00) &(1.13) &(5.47) &(0.17) &(0.19) &(0.02)\\\cmidrule(lr){2-13} 
     & \multirow{2}{*}{WMV}  &83.69 &90.40 &93.44 &95.95 &86.25 &60.48 &36.27 &58.29 &82.90 &56.10&74.38 \\
     & & (0.04) &(0.65) &(0.20) &(0.35) &(0.01) &(0.10) &(3.01) &(0.18) &(0.08) &(0.42)\\\cmidrule(lr){2-13}
     &  \multirow{2}{*}{\underline{WMV}}  &83.28 &87.87 &\color{blue}{97.20} &96.34 &87.22 &70.88 &32.49 &\color{blue}{59.55} &86.70 &57.93&75.95 \\
     & & (0.12) &(0.00) &(0.00) &(0.31) &(0.00) &(1.14) &(1.54) &(0.09) &(0.22) &(0.00)\\\cmidrule(lr){2-13}

     & \multirow{2}{*}{DS} & \color{red}{91.54} & 90.84 & 94.16 & 93.90 & 87.19 & 53.36 & 23.33 & 52.09 & 72.50 & 49.65 &70.86\\
     &   & (0.54) & (0.30) & (0.20) & (0.05) & (0.0) & (0.29) & (0.70) & (0.03) & (0.00) & (0.68)    \\\cmidrule(lr){2-13}
     & \multirow{2}{*}{\underline{DS}}    & 80.48 & \color{blue}{91.12} & 93.04 & 95.37 & 87.06 & 51.72 & 24.76 & 51.73 & 72.83 & 49.43 &69.75\\
     & & (0.0) & (0.11) & (0.20) & (0.08) & (0.01) & (1.17) & (0.57) & (0.04) & (0.00) & (1.15)  \\\cmidrule(lr){2-13}
     
     & \multirow{2}{*}{DP}  &\color{blue}{84.58} &88.44 &96.32 &33.70 &86.98 &\color{red}{78.72} &30.71 &\color{red}{60.46} &75.77 &57.51 &69.32\\
     & & (0.08) &(0.03) &(0.16) &(0.00) &(0.39) &(0.43) &(9.78) &(0.11) &(1.33) &(0.02)\\\cmidrule(lr){2-13} 
     &  \multirow{2}{*}{\underline{DP}}  &82.73 &91.02 &94.80 &36.44 &86.67 &72.40 &33.83 &58.47 &\colorbox{lightgray!60}{\red{88.77}} &\colorbox{lightgray!60}{\red{61.56}} &70.67 \\
     & &(0.03) &(0.13) &(0.00) &(0.00) &(0.00) &(0.00) &(0.00) &(0.16) &(0.13) &(0.06) \\\cmidrule(lr){2-13} 
     & \multirow{2}{*}{MeTaL}  &83.47 &89.76 &94.88 &95.62 &\color{blue}{87.26} &61.80 &35.84 &59.33 &79.20 &55.46 &74.26\\
     & & (0.12) &(0.00) &(0.53) &(0.31) &(0.02) &(0.00) &(6.73) &(0.04) &(2.33) &(0.12)\\\cmidrule(lr){2-13} 
     &  \multirow{2}{*}{\underline{MeTaL}}  &83.83 &90.68 &94.72 &93.75 &\color{red}{87.41} &71.20 &27.23 &59.14 &81.20 &57.85&74.70 \\
     & &(0.14) &(0.05) &(0.16) &(0.00) &(0.01) &(0.36) &(2.80) &(0.04) &(0.64) &(0.26)\\\cmidrule(lr){2-13} 
     & \multirow{2}{*}{FS}  &84.40 &89.05 &94.80 &62.27 &87.16 &27.60 &\colorbox{lightgray!60}{\red{56.52}} &48.89 &31.83 &48.10 &63.06  \\
     & & (0.00) &(0.07) &(0.00) &(0.17) &(0.16) &(0.00) &(0.32) &(0.08) &(0.00) &(0.60)\\\cmidrule(lr){2-13} 
     &  \multirow{2}{*}{\underline{FS}}  &82.64 &\color{red}{91.18} &96.16 &63.54 &86.57 &36.20 &\color{blue}{53.46} &55.69 &31.83 &49.35 &64.66\\
     & &(0.19) &(0.03) &(0.20) &(4.71) &(0.00) &(0.00) &(0.13) &(0.03) &(0.00) &(0.00)\\\midrule[0.05pt] \midrule[0.05pt]
     
     \multirow{26}{*}{R}    &  \multirow{2}{*}{Gold}  &93.25 &97.13 &95.68 &96.31 &91.39 &96.68 & -- &65.86 &93.23 &86.98  &90.72\\
     & &(0.30) &(0.26) &(1.42) &(0.58) &(0.38) &(0.82) & -- &(0.60) &(1.83) &(1.49)  \\\cmidrule(lr){2-13} 
    &  \multirow{2}{*}{MV} &85.76 &89.91 &\color{red}{96.56} &\color{blue}{94.17} &86.88 &66.28 &17.99 &55.07 &\color{blue}{84.00} &\color{blue}{56.85}& \blue{73.35} \\
     & &(0.70) &(1.76) &(0.86) &(2.88) &(0.98) &(1.21) &(1.99) &(3.47) &(0.84) &(1.91)  \\\cmidrule(lr){2-13}
     &  \multirow{2}{*}{\underline{MV}} &86.17 &87.87 &\color{blue}{95.60} &\color{red}{95.06} &87.14 &66.16 &\color{red}{21.68} &54.96 &\color{red}{84.13} &\color{red}{57.31} & \red{73.61} \\
     & &(1.31) &(1.18) &(0.80) &(1.66) &(0.18) &(1.25) &(8.32) &(5.42) &(0.59) &(1.07)   \\\cmidrule(lr){2-13}
      & \multirow{2}{*}{WMV} &86.06 &82.27 &92.96 &92.96 &86.70 &58.88 &16.14 &42.37 &67.47 &46.56 &67.24   \\
     &  &(0.88) &(4.11) &(1.73) &(1.71) &(0.51) &(0.92) &(1.40) &(21.19) &(6.93) &(11.71)   \\\cmidrule(lr){2-13}
     & \multirow{2}{*}{\underline{WMV}}  &86.03 &86.06 &95.52 &93.96 &86.99 &63.64 &17.43 &54.88 &82.87 &55.57 &72.30   \\
     & &(1.03) &(3.97) &(0.99) &(1.11) &(0.37) &(1.94) &(1.21) &(3.82) &(2.49) &(0.78)   \\\cmidrule(lr){2-13}
     
     & \multirow{2}{*}{DS} & 84.74 & \color{blue}{92.30} & 93.52 & 94.10 & 87.16 & 48.32 & 16.57 & 50.77 & 69.67 & 45.69  &68.28 \\
     &   & (1.41) & (1.75) & (1.39) & (1.72) & (0.58) & (1.50) & (0.25) & (0.12) & (1.18) & (0.86)   \\\cmidrule(lr){2-13}
     & \multirow{2}{*}{\underline{DS}}    & 86.85 & 92.06 & 92.96 & 93.17 & 86.82 & 50.12 & 16.93 & 50.85 & 70.80 & 46.96 &68.75 \\
     & & (0.72) & (1.20) & (1.53) & (0.89) & (0.29) & (1.99) & (0.52) & (0.37) & (0.61) & (0.38)  \\\cmidrule(lr){2-13}

     & \multirow{2}{*}{DP}  &86.26 &89.59 &\color{blue}{95.60} &28.25 &86.81 &\color{red}{72.12} &17.62 &54.42 &70.57 &39.91 &64.12  \\
     &  &(1.02) &(2.87) &(0.80) &(2.83) &(0.42) &(4.58) &(4.24) &(5.32) &(0.83) &(9.33)  \\\cmidrule(lr){2-13}
     & \multirow{2}{*}{\underline{DP}}  &84.86 &85.73 &94.48 &46.66 &\color{red}{87.65} &\color{blue}{66.80} &17.71 &\color{blue}{57.78} &72.60 &56.18  &67.05\\
     & &(0.58) &(3.49) &(1.17) &(11.89) &(0.37) &(0.85) &(2.27) &(0.79) &(20.40) &(1.12)  \\\cmidrule(lr){2-13}
      & \multirow{2}{*}{MeTaL}  &84.98 &89.08 &94.56 &93.28 &\color{blue}{87.18} &60.04 &16.42 &53.68 &70.73 &54.59 &70.45  \\
     & &(1.07) &(3.71) &(0.65) &(1.57) &(0.45) &(1.18) &(2.79) &(4.00) &(0.68) &(0.77) \\\cmidrule(lr){2-13} 
     & \multirow{2}{*}{\underline{MeTaL}} &\color{red}{87.23} &92.22 &94.08 &93.00 &86.87 &65.60 &\color{blue}{20.80} &\color{red}{59.19} &70.27 &42.02&71.13  \\
     &  &(0.97) &(1.14) &(1.70) &(1.42) &(0.37) &(1.67) &(7.13) &(0.35) &(0.88) &(11.91)  \\\cmidrule(lr){2-13}
     & \multirow{2}{*}{FS}  &86.95 &92.08 &93.84 &10.72 &86.69 &30.44 &0.00 &0.00 &31.83 &39.95&47.25    \\
     &  &(0.58) &(2.63) &(1.57) &(10.15) &(0.29) &(3.48) &(0.00) &(0.00) &(0.00) &(6.50)    \\ \cmidrule(lr){2-13}
     & \multirow{2}{*}{\underline{FS}}  &\color{blue}{87.10} &\color{red}{94.34} &93.20 &18.20 &86.17 &28.84 &0.00 &0.00 &31.83 &39.43 &47.91  \\
     & &(1.06) &(0.89) &(3.19) &(3.93) &(0.78) &(2.48) &(0.00) &(0.00) &(0.00) &(8.74)    \\\midrule[0.05pt] \midrule[0.05pt]
     
     \multirow{24}{*}{RC}    &  \multirow{2}{*}{MV}  &88.22 &94.23 &\color{red}{97.60} &96.67 &\color{blue}{88.15} &77.96 &\color{blue}{40.50} &60.38 &\color{blue}{86.20} &\color{red}{59.43}& \red{78.93} \\
     &  &(0.22) &(0.20) &(0.00) &(0.37) &(0.30) &(0.34) &(1.23) &(0.05) &(0.07) &(0.00) \\\cmidrule(lr){2-13} 
     &  \multirow{2}{*}{\underline{MV}}  &\color{blue}{88.48} &91.06 &\color{red}{97.60} &96.82 &88.04 &74.28 &\color{red}{46.28} &\color{blue}{61.13} &83.93 &\color{blue}{59.32}& \blue{78.69}  \\
     &  &(0.00) &(0.39) &(0.00) &(0.29) &(0.06) &(0.75) &(1.59) &(0.12) &(0.20) &(0.06) \\\cmidrule(lr){2-13} 
     & \multirow{2}{*}{WMV}  &87.46 &92.53 &95.60 &\colorbox{lightgray!60}{\red{98.02}} &87.83 &70.28 &20.76 &56.27 &72.77 &55.58 &73.71  \\
     & &(0.05) &(0.06) &(0.00) &(0.38) &(0.13) &(1.09) &(1.44) &(0.10) &(0.48) &(0.34)  \\\cmidrule(lr){2-13} 
     &  \multirow{2}{*}{\underline{WMV}}  &88.00 &93.16 &\color{blue}{97.20} &97.27 &88.11 &72.08 &30.07 &58.66 &84.67 &58.31&76.75 \\
     &  &(0.00) &(0.03) &(0.00) &(0.36) &(0.05) &(1.01) &(2.35) &(0.46) &(0.00) &(0.09) \\\cmidrule(lr){2-13} 
     
     & \multirow{2}{*}{DS} & 88.01 & 94.19 & 96.24 & 96.79 & \colorbox{lightgray!60}{\red{88.20}} & 59.40 & 21.34 & 51.37 & 71.70 & 46.75&71.40 \\
     &  & (0.56) & (0.18) & (0.41) & (0.27) & (0.11) & (0.42) & (1.19) & (0.60) & (0.07) & (0.27)     \\\cmidrule(lr){2-13}
     & \multirow{2}{*}{\underline{DS}}    & 87.77 & 95.01 & 95.52 & 97.10 & 87.21 & 57.96 & 28.75 & 52.25 & 77.03 & 49.23 &72.78\\
     &  & (0.05) & (0.25) & (0.30) & (0.31) & (0.0) & (0.15) & (1.03) & (0.25) & (0.52) & (0.11)  \\\cmidrule(lr){2-13}

     & \multirow{2}{*}{DP}  & 87.91 &94.09 &96.80 &31.71 &87.53 &\colorbox{lightgray!60}{\red{82.36}} &28.86 &\colorbox{lightgray!60}{\red{61.40}} &75.17 &52.86 &69.87 \\
     & &(0.15) &(0.06) &(0.00) &(0.29) &(0.03) &(0.08) &(10.02) &(0.07) &(0.95) &(0.06)  \\\cmidrule(lr){2-13}
      &  \multirow{2}{*}{\underline{DP}}  &87.30 &94.40 &95.60 &64.22 &88.04 &74.00 &21.74 &59.86 &\color{red}{86.73} &55.96 &72.79 \\
     &  &(0.66) &(0.37) &(0.00) &(0.00) &(0.00) &(0.77) &(0.00) &(0.17) &(0.08) &(0.06) \\\cmidrule(lr){2-13} 
     & \multirow{2}{*}{MeTaL}  &86.46 &93.11 &97.04 &\color{blue}{97.71} &87.85 &71.64 &23.99 &58.29 &70.90 &53.32 &74.03 \\
     &  &(0.11) &(0.01) &(0.20) &(0.00) &(0.02) &(0.59) &(8.47) &(0.39) &(0.08) &(0.19) \\\cmidrule(lr){2-13}
      &  \multirow{2}{*}{\underline{MeTaL}}  &\colorbox{lightgray!60}{\red{88.86}} &93.95 &96.00 &96.18 &87.43 &\color{blue}{79.84} &21.89 &60.16 &84.20 &56.89&76.54 \\
     &  &(0.14) &(0.00) &(0.00) &(0.00) &(0.01) &(0.23) &(4.72) &(0.16) &(0.16) &(0.11)  \\\cmidrule(lr){2-13} 
     & \multirow{2}{*}{FS} &87.65 &\colorbox{lightgray!60}{\red{95.45}} &95.20 &82.24 &87.73 &38.80 &16.06 &38.14 &31.83 &48.60 &62.17   \\
     & &(0.06) &(0.10) &(0.00) &(0.93) &(0.12) &(0.33) &(0.15) &(6.62) &(0.00) &(0.11) \\\cmidrule(lr){2-13} 
      &  \multirow{2}{*}{\underline{FS}}  &\color{blue}{88.48} &\color{blue}{95.33} &96.80 &65.65 &87.23 &33.80 &0.00 &0.00 &31.83 &39.89&53.90 \\
     &  &(0.00) &(0.06) &(0.00) &(0.00) &(0.00) &(0.00) &(0.00) &(0.00) &(0.00) &(0.00)   \\
    \bottomrule
    \end{tabular}
    }

    \label{tab:classification_continue}
\end{table*}

\subsection{Sequence Tagging}
\label{sec:seq_app}

The detailed comparisons over the collected sequence tagging datasets are in Table~\ref{tab:seq_full}.

\begin{table*}[t]
    \centering
    \caption{\textbf{Sequence Tagging.} The detailed results for different methods. The number stands for the F1 score (Precision, Recall) with standard deviation in the bracket under each value. Each metric value is averaged over 5 runs. {\textcolor{red}{red}} and {\color{blue}{blue}} indicate the best and second best result for each end model respectively, and \colorbox{lightgray!60}{gray} is the best weak supervision method.}
    \scalebox{0.43}{
    \begin{tabular}{ l c c c c c c c c c c}
        \toprule
           \textbf{End Model ($\downarrow$)} &\textbf{Label Model ($\downarrow$)} & \textbf{CoNLL-03} & \textbf{WikiGold} & \textbf{BC5CDR} & \textbf{NCBI-Disease}& \textbf{Laptop-Review} & \textbf{MIT-Restaurant} & \textbf{MIT-Movies} & \textbf{Ontonotes 5.0} & \textbf{\underline{Average}}     \\
         \midrule
    
    \multirow{16}{*}{--}    
     & \multirow{2}{*}{MV}     & 60.36(59.06/61.72) & 52.24(48.95/56.00) & 83.49(91.69/76.64) & \blue{78.44(93.04/67.79) } & \blue{73.27(88.86/62.33) } & \colorbox{lightgray!60}{\red{48.71(74.25/36.24) }} & 59.68(69.92/52.05) & 58.85(54.17/64.40)   & 64.38(72.50/59.65)     \\
     & & (0.00) & (0.00) & (0.00) & (0.00) & (0.00) & (0.00) & (0.00) & (0.00) \\
     \cmidrule(lr){2-11} 
     & \multirow{2}{*}{WMV}    & 60.26(59.03/61.54) & 52.87(50.74/55.20) & 83.49(91.66/76.66) & \blue{78.44(93.04/67.79) } & \blue{73.27(88.86/62.33) } & \blue{48.19(73.73/35.80) } & 60.37(70.98/52.52) & 57.58(53.15/62.81) & 64.31(72.65/59.33)        \\ 
     & & (0.00) & (0.00) & (0.00) & (0.00) & (0.00) & (0.00) & (0.00) & (0.00) \\
     \cmidrule(lr){2-11} 
     & \multirow{2}{*}{DS}     & 46.76(45.29/48.32) & 42.17(40.05/44.53) & 83.49(91.66/76.66) & \blue{78.44(93.04/67.79) } & \blue{73.27(88.86/62.33) } & 46.81(71.71/34.75) & 54.06(63.64/46.99) & 37.70(34.33/41.82)  & 57.84(66.07/52.90)    \\
     & & (0.00) & (0.00) & (0.00) & (0.00) & (0.00) & (0.00) & (0.00) & (0.00) \\
     \cmidrule(lr){2-11} 
     & \multirow{2}{*}{DP}      & 62.43(61.62/63.26) & 54.81(53.10/56.64) & \blue{83.50(91.69/76.65) } & \blue{78.44(93.04/67.79) } & \blue{73.27(88.86/62.33) } & 47.92(73.24/35.61) & 59.92(70.65/52.01) & \blue{61.85(57.44/66.99) }  & \blue{65.27(73.71/60.16)}      \\
     & & (0.22) & (0.13) & (0.00) & (0.00) & (0.00) & (0.00) & (0.43) & (0.19) \\
     \cmidrule(lr){2-11} 
     & \multirow{2}{*}{MeTaL}   & 60.32(59.07/61.63) & 52.09(50.31/54.03) & 83.50(91.66/76.67) & \blue{78.44(93.04/67.79) } & 64.36(83.21/53.63) & 47.66(73.40/35.29) & 56.60(72.28/47.70) & 58.27(54.10/63.14)   & 62.66(72.14/57.48)      \\
     & & (0.08) & (0.23) & (0.00) & (0.00) & (17.81) & (0.00) & (7.71) & (0.48) \\
     \cmidrule(lr){2-11} 
     & \multirow{2}{*}{FS}     & \blue{62.49(63.25/61.76) } & \blue{58.29(62.77/54.40) } & 56.71(88.03/41.83) & 40.67(72.24/28.30) & 28.74(60.59/18.84) & 13.86(84.10/7.55) & 43.04(77.73/29.75) & 5.31(2.87/35.74)   & 38.64(63.95/34.77)     \\
     & & (0.00) & (0.00) & (0.00) & (0.00) & (0.00) & (0.00) & (0.00) & (0.00) \\
     \cmidrule(lr){2-11} 
     & \multirow{2}{*}{HMM}   & 62.18(66.42/58.45) & 56.36(61.51/52.00) & 71.57(93.48/57.98) & 66.80(96.79/51.00) & \colorbox{lightgray!60}{\red{73.63(89.30/62.63) }} & 42.65(71.44/30.40) & \blue{60.56(75.04/50.76) } & 55.67(57.95/53.57) & 61.18(76.49/52.10) \\
     & & (0.00) & (0.00) & (0.00) & (0.00) & (0.00) & (0.00) & (0.00) & (0.00)\\
     \cmidrule(lr){2-11} 
     & \multirow{2}{*}{CHMM}    & \red{63.22(61.93/64.56) } & \red{58.89(55.71/62.45) } & \colorbox{lightgray!60}{\red{83.66(91.76/76.87) }}  & \colorbox{lightgray!60}{\red{78.74(93.21/68.15) }} & 73.26(88.79/62.36) & 47.34(73.05/35.02) & \red{61.38(73.00/52.96) } & \red{64.06(59.70/69.09) } & \red{66.32(74.65/61.43)} \\ 
     & & (0.26) & (0.97) & (0.04) & (0.10) & (0.13) & (0.57) & (0.10) & (0.07) \\
     \midrule[0.05pt] \midrule[0.05pt]
     
    \multirow{20}{*}{LSTM-CNN-MLP}  
    & \multirow{2}{*}{Gold}         & 87.46(87.72/87.19) & 80.45(80.80/80.11) & 78.02(79.80/76.34) & 79.41(80.94/77.97) & 69.83(74.51/65.73) & 77.80(78.42/77.19) & 86.18(87.05/85.33) & 79.52(79.91/79.14)  & 79.83(81.14/78.63)  \\
     & & (0.35) & (1.46) & (0.20) & (0.54) & (0.51) & (0.28) & (0.40) & (0.40) \\
     \cmidrule(lr){2-11} 
     & \multirow{2}{*}{MV}            & 66.33(67.54/65.19) & 58.27(56.65/60.00) & \red{74.56(79.70/70.05) } & 70.54(80.95/62.51) & 62.32(74.03/53.81) & \red{41.30(63.57/30.59) } & 61.50(72.78/53.25) & 60.04(56.89/63.57) & 61.86(69.01/57.37) \\
     & & (0.52) & (0.67) & (0.21) & (0.79) & (1.24) & (0.37) & (0.22) & (0.53) \\
     \cmidrule(lr){2-11} 
     & \multirow{2}{*}{WMV}      & 64.60(65.31/63.91) & 53.86(54.01/53.76) & 74.29(79.95/69.39) & 70.94(81.81/62.64) & 61.73(72.92/53.54) & 40.60(62.48/30.07) & 61.31(72.62/53.05) & 58.47(55.46/61.83) & 60.72(68.07/56.02)     \\ 
     & & (0.73) & (0.48) & (0.16) & (0.81) & (0.66) & (0.26) & (0.25) & (0.09) \\
     \cmidrule(lr){2-11} 
     & \multirow{2}{*}{DS}       & 50.60(49.35/51.90) & 40.01(38.00/42.24) & 74.21(80.15/69.11) & 70.71(81.38/62.53) & \red{63.12(74.62/54.70) } & 39.86(61.98/29.37) & 54.59(67.94/45.63) & 41.13(40.55/41.75)       & 54.28(61.75/49.66)\\
     & & (0.72) & (3.13) & (0.34) & (0.67) & (0.59) & (0.23) & (0.45) & (0.98) \\
     \cmidrule(lr){2-11} 
     & \multirow{2}{*}{DP}        & \red{66.98(68.95/65.13) } & 57.87(58.34/57.44) & 74.26(79.09/70.00) & 70.75(80.69/63.06) & 61.02(72.90/52.47) & \blue{41.06(63.10/30.43) } & 61.05(72.66/52.64) & \blue{62.68(59.51/66.21) }  & \blue{61.96(69.41/57.17)}     \\
     & & (0.50) & (1.27) & (0.33) & (1.35) & (0.76) & (0.17) & (0.45) & (0.06) \\
     \cmidrule(lr){2-11} 
     & \multirow{2}{*}{MeTaL}    & 65.05(66.20/63.94) & 56.31(55.94/56.69) & \blue{74.37(79.86/69.62) } & \blue{71.23(80.59/63.82) } & 62.43(74.18/53.91) & 40.97(62.58/30.46) & 61.02(72.44/52.71) & 59.15(55.99/62.68)   & 61.32(68.47/56.73)     \\
     & & (0.73) & (1.64) & (0.20) & (0.78) & (0.47) & (0.60) & (0.32) & (0.45)\\
     \cmidrule(lr){2-11} 
     & \multirow{2}{*}{FS}      & 66.49(69.13/64.05) & 59.80(65.71/54.88) & 54.37(77.11/42.00) & 42.70(71.11/30.52) & 27.08(54.14/18.13) & 13.09(74.74/7.17) & 45.77(75.50/32.84) & 41.54(44.13/39.48) & 43.85(66.45/36.13)
      \\
     & & (0.25) & (1.65) & (0.42) & (1.03) & (3.20) & (0.30) & (0.61) & (2.06)  \\
     \cmidrule(lr){2-11} 
     & \multirow{2}{*}{HMM}      & 64.85(69.48/60.81) & \blue{60.87(67.45/55.47) } & 64.07(79.54/53.67) & 57.80(81.75/44.71) & \blue{62.53(74.62/53.81) } & 37.47(60.36/27.17) & \blue{61.55(76.16/51.65) } & 58.17(61.13/55.49)   & 58.41(71.31/50.35)     \\
     & & (1.41) & (0.95) & (0.38) & (0.60) & (0.48) & (0.50) & (0.24) & (0.23) \\
     \cmidrule(lr){2-11} 
     & \multirow{2}{*}{CHMM}     & \blue{66.67(67.75/65.64) } & \red{61.34(60.49/62.24) } & 74.29(79.55/69.70) & \red{71.45(80.90/64.01) } & 62.18(73.86/53.72) & 40.97(63.11/30.33) & \red{62.05(73.57/53.65) } & \red{63.70(60.58/67.17) }  & \red{62.83(69.98/58.31)}    \\ 
     & & (0.25) & (1.79) & (0.11) & (0.72) & (0.84) & (0.17) & (0.23) & (0.45) \\
     \midrule[0.05pt] \midrule[0.05pt]
    
     \multirow{22}{*}{LSTM-CNN-CRF}  
    & \multirow{2}{*}{Gold}         & 86.80(86.80/86.80) & 79.79(79.79/79.79) & 78.59(80.90/76.41) & 79.39(80.34/78.48) & 71.25(76.37/66.80) & 79.18(79.68/78.69) & 87.07(87.49/86.64) & 79.42(80.14/78.72)   & 80.19(81.44/79.04) \\
    & & (0.74) & (0.49) & (0.70) & (0.59) & (1.80) & (0.20) & (0.19) & (2.76)  \\
     \cmidrule(lr){2-11} 
     & \multirow{2}{*}{MV}    & 65.97(67.14/64.85) & 57.04(54.91/59.36) & 74.75(79.90/70.22) & 72.44(82.56/64.60) & 63.52(75.14/55.01) & \red{41.70(63.92/30.95) } & \blue{62.41(74.45/53.72) } & 61.92(59.47/64.59)  & 62.47(69.69/57.91)     \\
     & & (0.81) & (1.33) & (0.60) & (1.44) & (0.96) & (0.21) & (0.28) & (0.55)  \\
     \cmidrule(lr){2-11} 
     & \multirow{2}{*}{WMV}    & 63.76(63.76/63.76) & 55.39(54.30/56.53) & 74.31(79.59/69.69) & 72.21(83.05/63.89) & 63.02(75.36/54.15) & 41.27(63.20/30.64) & 61.79(73.72/53.19) & 59.22(56.76/61.91)& 61.37(68.72/56.72)  \\ 
     & & (1.06) & (0.95) & (0.63) & (1.33) & (0.98) & (0.31) & (0.42) & (0.31) \\
     \cmidrule(lr){2-11} 
     & \multirow{2}{*}{DS}  & 49.74(48.66/50.91) & 40.61(38.31/43.20) & \red{75.37(80.88/70.56) } & \red{72.86(82.69/65.15) } & \red{63.96(75.27/55.62) } & 41.21(63.02/30.61) & 55.99(68.63/47.29) & 44.92(42.91/47.16) & 55.58(62.55/51.31)  \\
     & & (1.41) & (1.89) & (0.28) & (1.01) & (0.79) & (0.34) & (0.50) & (1.96) \\
     \cmidrule(lr){2-11} 
     & \multirow{2}{*}{DP}     & \red{67.15(67.47/66.83) } & 57.89(57.56/58.24) & \blue{74.79(80.48/69.91) } & \blue{72.50(82.83/64.48) } & 62.59(74.16/54.15) & \blue{41.62(63.43/30.97) } & 62.29(74.31/53.63) & \red{63.82(61.11/66.80) }  & \blue{62.83(70.17/58.13)}     \\
     & & (0.69) & (1.99) & (0.68) & (0.86) & (0.83) & (0.47) & (0.22) & (0.29) \\
     \cmidrule(lr){2-11} 
     & \multirow{2}{*}{MeTaL}     & 64.48(65.77/63.29) & 55.37(54.26/56.53) & 74.66(79.95/70.03) & 72.42(83.41/64.01) & \blue{63.87(75.34/55.44) } & 41.48(63.09/30.90) & 62.10(73.97/53.52) & 60.43(57.99/63.08) & 61.85(69.22/57.10)    \\
     & & (0.85) & (1.69) & (0.88) & (1.44) & (1.53) & (0.45) & (0.48) & (0.31) \\
     \cmidrule(lr){2-11} 
     & \multirow{2}{*}{FS}  & 66.21(68.71/63.90) & \blue{60.49(65.46/56.27) } & 54.49(77.53/42.02) & 44.90(74.39/32.19) & 28.35(54.68/19.20) & 12.74(71.00/7.00) & 45.62(77.19/32.38) & 43.25(46.46/40.56)  & 44.51(66.93/36.69) \\
     & & (0.79) & (3.30) & (0.47) & (1.15) & (1.61) & (0.13) & (0.44) & (0.53)  \\
     \cmidrule(lr){2-11} 
     & \multirow{2}{*}{HMM}  & 66.18(70.66/62.24) & \red{62.51(71.09/55.79) } & 63.68(77.69/53.98) & 59.12(84.26/45.55) & 62.57(74.21/54.09) & 37.90(62.48/27.20) & 61.94(76.85/51.89) & 59.43(61.53/57.47) & 59.17(72.35/51.03)  \\
     & & (1.27) & (1.21) & (0.71) & (1.15) & (0.30) & (0.56) & (0.44) & (0.62) \\
     \cmidrule(lr){2-11} 
     & \multirow{2}{*}{CHMM}   & \blue{66.58(67.05/66.17) } & 59.90(57.38/62.67) & 74.54(80.49/69.44) & 72.15(82.51/64.12) & 62.28(74.15/53.69) & 41.59(63.67/30.89) & \red{62.97(74.82/54.36) } & \blue{63.71(61.40/66.20) } & \red{62.97(70.19/58.44)} \\ 
     & & (0.75) & (2.86) & (0.50) & (1.42) & (0.69) & (0.40) & (0.30) & (0.36) \\
     \midrule[0.05pt] \midrule[0.05pt]

     \multicolumn{2}{c}{\multirow{2}{*}{LSTM-ConNet}}    & 66.02(67.98/64.19) & 58.04(61.10/55.36) & 72.04(77.71/67.18) & 63.04(74.55/55.16) & 50.36(63.04/42.73) & 39.26(61.74/28.78) & 60.46(75.61/50.38) & 60.58(59.43/61.83)   & 58.73(67.65/53.20)  \\ 
     & & (0.95) & (1.60) & (0.54) & (12.69) & (7.74) & (0.46) & (0.90) & (0.46) \\
     \midrule[0.05pt] \midrule[0.05pt]
     
     \multirow{20}{*}{BERT-MLP}  
    & \multirow{2}{*}{Gold}      & 89.41(90.06/88.76) & 87.21(87.88/86.56) & 82.49(86.67/78.70) & 84.05(84.08/84.03) & 81.22(83.67/78.96) & 78.85(78.74/78.96) & 87.56(87.57/87.54) & 84.11(83.11/85.14)       & 84.36(85.22/83.58) \\
     & & (0.21) & (0.65) & (0.28) & (0.29) & (1.20) & (0.33) & (0.21) & (0.55)\\
     \cmidrule(lr){2-11} 
     & \multirow{2}{*}{MV}      & 67.08(68.35/65.86) & 63.17(64.15/62.29) & \blue{77.93(84.26/72.50) } & 77.93(85.84/71.38) & 69.88(75.99/64.84) & 41.89(62.80/31.43) & 63.20(73.74/55.35) & 63.86(61.26/66.71)   & 65.62(72.05/61.29)     \\
     & & (0.71) & (2.15) & (0.43) & (0.73) & (1.17) & (0.59) & (0.70) & (0.60) \\
     \cmidrule(lr){2-11} 
     & \multirow{2}{*}{WMV}    & 65.96(66.88/65.07) & 61.28(63.88/59.25) & 77.76(84.63/71.94) & 78.53(86.44/71.97) & \blue{71.60(76.96/66.95) } & \red{42.40(62.88/31.98) }& 65.26(71.65/60.89) & 62.90(72.32/55.66) & 61.63(59.19/64.29)         \\ 
     & & (0.52) & (2.18) & (0.47) & (0.85) & (0.68) & (0.57) & (0.47) & (0.48) \\
     \cmidrule(lr){2-11} 
     & \multirow{2}{*}{DS}    & 59.48(65.12/55.91)   & 54.04(54.24/53.91) & 49.09(46.69/51.79) & 77.57(84.62/71.62) & \blue{78.69(86.37/72.27) } & 71.41(76.00/67.38) & 41.14(61.70/30.86) & 58.62(68.60/51.19) & 45.32(42.74/48.24)       \\
     & & (0.90) & (1.66) & (0.20) & (0.55) & (0.88) & (0.72) & (0.56) & (2.14)\\
     \cmidrule(lr){2-11} 
     & \multirow{2}{*}{DP}      & 67.66(68.82/66.55) & 62.91(64.44/61.49) & 77.67(83.87/72.33) & 78.18(85.91/71.74) & 70.86(77.70/65.15) & 42.06(62.49/31.70) & \blue{63.28(73.36/55.64) } & \blue{65.16(63.15/67.33) }  & \blue{65.97(72.47/61.49)}       \\
     & & (0.73) & (1.23) & (0.40) & (0.71) & (1.10) & (0.31) & (0.35) & (0.74) \\
     \cmidrule(lr){2-11} 
     & \multirow{2}{*}{MeTaL}    & 66.34(67.46/65.28) & 61.74(61.55/61.97) & 77.80(83.77/72.63) & \red{79.02(85.98/73.12) } & \red{71.80(76.17/67.93) } & 42.07(62.69/31.66) & 63.00(73.02/55.40) & 63.08(60.94/65.36)    & 65.61(71.45/61.67)   \\
    & & (1.29) & (1.84) & (0.21) & (0.59) & (0.81) & (0.74) & (0.28) & (0.46)  \\
     \cmidrule(lr){2-11} 
     & \multirow{2}{*}{FS}     & 67.54(69.81/65.43) & \colorbox{lightgray!60}{\red{66.58(72.23/61.76) }} & 62.89(79.81/52.02) & 46.50(72.13/34.34) & 36.87(63.98/26.09) & 13.52(71.73/7.47) & 49.37(75.67/36.64) & 49.63(57.13/43.91)    & 49.11(70.31/40.96)     \\
     & & (1.32) & (1.40) & (1.39) & (1.32) & (1.96) & (0.74) & (0.34) & (2.49) \\
     \cmidrule(lr){2-11} 
     & \multirow{2}{*}{HMM}     & \colorbox{lightgray!60}{\red{68.48(71.04/66.17) }} & 64.25(68.96/60.16) & 68.70(81.86/59.20) & 65.52(87.25/52.52) & 71.51(75.86/67.66) & 38.10(61.40/27.62) & 63.07(76.72/53.55) & 61.13(63.55/58.93)        & 62.59(73.33/55.72) \\
     & & (0.16) & (1.65) & (0.82) & (1.44) & (0.58) & (0.57) & (0.53) & (0.47) \\
     \cmidrule(lr){2-11} 
     & \multirow{2}{*}{CHMM}    & \blue{68.30(69.10/67.54) } & \blue{65.16(63.45/66.99) } & \red{77.98(83.74/72.98) } & 78.20(85.04/72.40) & 70.58(77.41/64.87) & \blue{42.10(62.88/31.64) } & \red{63.68(73.49/56.18) } & \colorbox{lightgray!60}{\red{66.03(63.42/68.87) }}   & \colorbox{lightgray!60}{\red{66.50(72.32/62.68)}}     \\ 
     & & (0.44) & (0.67) & (0.13) & (0.71) & (0.48) & (0.27) & (0.29) & (0.29)\\
     \midrule[0.05pt] \midrule[0.05pt]
     
     \multirow{22}{*}{BERT-CRF}  
    & \multirow{2}{*}{Gold}        & 87.38(87.70/87.06) & 86.78(87.27/86.29) & 79.65(79.48/79.83) & 80.64(81.50/79.83) & 79.15(81.26/77.18) & 78.83(79.14/78.53) & 87.03(87.12/86.94) & 83.86(82.32/85.45)      & 82.91(83.22/82.64)\\
    & & (0.34) & (0.84) & (0.29) & (0.27) & (0.77) & (0.44) & (0.91) & (0.18)  \\
     \cmidrule(lr){2-11} 
     & \multirow{2}{*}{MV}      & 66.63(67.68/65.62) & 62.09(61.89/62.29) & 74.93(75.84/74.04) & 72.87(83.57/64.63) & 71.12(76.74/66.34) & \red{42.95(63.18/32.54) } & 63.71(73.46/56.25) & 63.97(61.15/67.08)    & \blue{64.78(70.44/61.10)}    \\
     & & (0.85) & (1.06) & (0.32) & (0.62) & (1.83) & (0.43) & (0.23) & (0.58) \\
     \cmidrule(lr){2-11} 
     & \multirow{2}{*}{WMV}      & 64.38(66.55/62.35) & 59.96(60.33/59.73) & 75.32(77.59/73.18) & \blue{73.23(83.77/65.07) } & 71.09(76.16/66.68) & 42.62(63.56/32.06) & 63.44(72.85/56.19) & 61.29(58.70/64.14)  & 63.92(69.94/59.93)     \\ 
     & & (1.09) & (1.08) & (0.39) & (0.71) & (0.73) & (0.23) & (0.29) & (0.32) \\
     \cmidrule(lr){2-11} 
     & \multirow{2}{*}{DS}      & 53.89(54.10/53.68) & 48.89(46.80/51.20) & \red{75.42(76.91/74.00) } & 72.91(82.60/65.30) & 70.19(76.49/64.87) & 42.26(62.65/31.89) & 58.89(69.67/51.01) & 48.55(46.97/50.26)   & 58.87(64.52/55.28)     \\
     & & (1.42) & (1.59) & (0.32) & (0.52) & (1.46) & (0.78) & (0.34) & (1.23) \\
     \cmidrule(lr){2-11} 
     & \multirow{2}{*}{DP}       & 65.48(66.76/64.28) & 61.09(61.07/61.12) & 75.08(76.78/73.47) & 72.86(81.54/65.85) & \red{71.46(76.42/67.14) } & 42.27(62.81/31.86) & 63.92(73.05/56.84) & \blue{65.09(63.27/67.04) }   & 64.66(70.21/60.95)     \\
     & & (0.37) & (1.53) & (0.55) & (0.92) & (0.86) & (0.53) & (0.36) & (0.31)\\
     \cmidrule(lr){2-11} 
     & \multirow{2}{*}{MeTaL}     & 65.11(66.87/63.45) & 58.94(61.53/56.75) & \blue{75.32(76.71/73.99) } & \red{74.16(82.66/67.31) } & \blue{71.24(76.66/66.62) } & 42.26(62.82/31.84) & \blue{64.19(73.30/57.10) } & 62.13(60.26/64.13)  & 64.17(70.10/60.15)     \\
     & & (0.69) & (3.22) & (0.20) & (1.02) & (1.14) & (0.49) & (0.52) & (0.40) \\
     \cmidrule(lr){2-11} 
     & \multirow{2}{*}{FS}       & \blue{67.34(70.05/64.83) } & \red{66.44(72.86/61.17) } & 59.38(71.35/50.94) & 44.12(73.05/31.62) & 38.57(61.91/28.09) & 13.80(72.63/7.62) & 49.79(75.45/37.17) & 42.45(44.03/41.51)      & 47.73(67.67/40.37)\\
    & & (0.75) & (1.40) & (1.30) & (1.15) & (2.55) & (0.23) & (1.03) & (5.05) \\
     \cmidrule(lr){2-11} 
     & \multirow{2}{*}{HMM}      & \red{67.49(71.26/64.14) } & \blue{63.31(70.95/57.33) } & 67.37(75.17/61.12) & 61.43(84.29/48.36) & 70.28(76.41/65.08) & 39.51(62.49/28.90) & 63.38(76.46/54.15) & 61.29(63.86/58.93)       & 61.76(72.61/54.75) \\
     & & (0.89) & (1.02) & (0.70) & (1.60) & (0.71) & (0.72) & (0.81) & (0.78) \\
     \cmidrule(lr){2-11} 
     & \multirow{2}{*}{CHMM}     & 66.72(67.17/66.27) & 63.06(62.12/64.11) & 75.21(76.61/73.88) & 72.96(81.31/66.25) & 71.17(76.66/66.43) & \blue{42.79(63.19/32.35) } & \colorbox{lightgray!60}{\red{64.58(74.77/56.84) }} & \red{65.26(61.99/68.91) }       & \red{65.22(70.48/61.88)}\\ 
     & & (0.41) & (1.91) & (0.41) & (0.93) & (0.95) & (0.22) & (0.74) & (0.20)      \\
     \midrule[0.05pt] \midrule[0.05pt]
     \multicolumn{2}{c}{\multirow{2}{*}{BERT-ConNet}}  & 67.83(69.37/66.40) & 64.18(72.17/57.92) & 72.87(73.25/72.60) & 71.40(80.30/64.56) & 67.32(73.60/62.14) & 42.37(62.88/31.95) & 64.12(74.03/56.56) & 60.36(57.81/63.21)  & 63.81(70.43/59.42)  \\ 
     & & (0.62) & (1.71) & (0.91) & (1.81) & (1.24) & (0.72) & (0.51) & (0.61)\\
    \bottomrule
    \end{tabular}
    }
    \label{tab:seq_full}
\end{table*}

\end{document}